\documentclass[10pt,twocolumn]{article}
\usepackage{amsmath,amssymb}
\usepackage{algorithm}
\usepackage{algorithmic}
\usepackage{graphicx}
\usepackage{subfig}
\usepackage{tabularx}
\usepackage{color}
\usepackage{verbatim}
\usepackage{gensymb}
\usepackage{url}
\usepackage{xcolor,colortbl}
\usepackage{array}
\usepackage{float}
\usepackage{authblk}

\input{./Definitions}
\graphicspath{{./Figures/}}

\makeatletter
\newsavebox\myboxA
\newsavebox\myboxB
\newlength\mylenA

\newcommand*\xoverline[2][0.75]{%
    \sbox{\myboxA}{$\m@th#2$}%
    \setbox\myboxB\null
    \ht\myboxB=\ht\myboxA%
    \dp\myboxB=\dp\myboxA%
    \wd\myboxB=#1\wd\myboxA
    \sbox\myboxB{$\m@th\overline{\copy\myboxB}$}
    \setlength\mylenA{\the\wd\myboxA}
    \addtolength\mylenA{-\the\wd\myboxB}%
    \ifdim\wd\myboxB<\wd\myboxA%
       \rlap{\hskip 0.5\mylenA\usebox\myboxB}{\usebox\myboxA}%
    \else
        \hskip -0.5\mylenA\rlap{\usebox\myboxA}{\hskip 0.5\mylenA\usebox\myboxB}%
    \fi}
\makeatother

\begin{document}

\title{\emph{Lie-X}: Depth Image Based Articulated Object Pose Estimation, Tracking, and Action Recognition on Lie Groups
}

\author[1]{Chi~Xu}
\author[1]{Lakshmi~Narasimhan~Govindarajan}
\author[1]{Yu~Zhang}
\author[1]{Li~Cheng\thanks{Corresponding author with email address: chengli@bii.a-star.edu.sg}}
\affil[1]{Bioinformatics Institute, A*STAR, Singapore}

\date{}

%
%
%

\maketitle

\begin{abstract}
Pose estimation, tracking, and action recognition of articulated objects from depth images are important and challenging problems, which are normally considered separately.
In this paper, a unified paradigm based on Lie group theory is proposed, which enables us to collectively address these related problems.
Our approach is also applicable to a wide range of articulated objects. Empirically it is evaluated on lab animals including mouse and fish, as well as on human hand.
On these applications, it is shown to deliver competitive results compared to the state-of-the-arts, and non-trivial baselines including convolutional neural networks and regression forest methods.
Moreover, new sets of annotated depth data of articulated objects are created which, together with our code, are made publicly available.
\end{abstract}

\section{Introduction}

With 3D cameras becoming increasingly ubiquitous in the recent years, there has been growing interest in utilizing depth images for key problems involving articulated objects
(e.g. human full-body and hand) such as pose estimation~\cite{ShoEtAl:pami13,TomEtAl:siggraph14,XuEtAl:IJCV15,ObeWohLep:iccv15,TanEtAl:iccv15,TanEtAl:cvpr16,zhoumodel:ijcai16}, tracking~\cite{BraBel:cvpr05,OikArg:bmvc11,BalEtAl:eccv12,QiaEtAl:cvpr14,HuaEtAl:cvpr16},
and action recognition~\cite{DolEtAl:PETS05,MahTod:cvpr16,VemChe:cvpr16,RahMia:cvpr16}.
%
On the other hand, although they are closely related,
most existing research efforts targeting these problems in literature are based on diverse and possibly disconnected principles.
Moreover, existing algorithms typically focus on a unique type of articulated objects, such as                                                                                                                            human full-body, or human hand.
This leads us to consider in this paper a principled approach to address these related problems across object categories, in a consistent and sensible manner. 

\begin{figure}[!t]
    \centering
	    \includegraphics[width=0.45\textwidth]{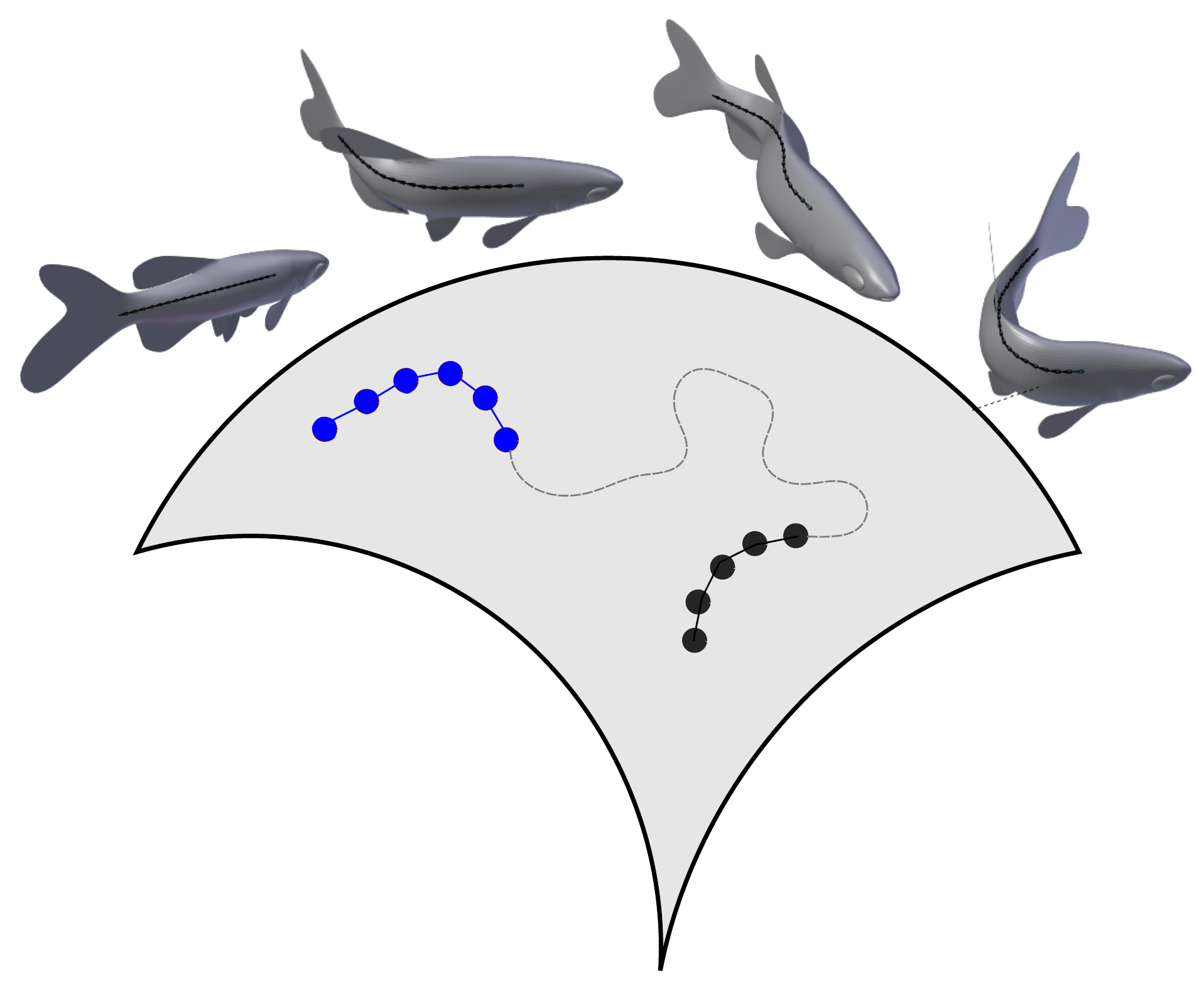}
    \caption{A cartoon illustration of our main idea: An articulated object can be considered as a point in certain manifold. Its 3D pose, a long track of its 3D motion, and its action sequences (color-coded herein) correspond to a point, a curve, and curve segments in the underlying manifold respectively.}
    \label{fig:concept1}
\end{figure}

Our approach possesses the following contributions:
(1) A unified Lie group-based paradigm is proposed to address the problems of pose estimation, tracking, and action recognition of articulated objects from depth images.
As illustrated in Fig.~\ref{fig:concept1},
a 3D pose of an articulate object corresponds to a point in the underlying pose manifold, a long-time track of its 3D poses amounts to a long curve in the same manifold,
whilst an action is represented as a certain curve segment.
Therefore, given a depth image input, pose estimation corresponds to inferring the optimal point in the manifold;
Action recognition amounts to classifying a curve segment in the same manifold as a particular action type; 
Meanwhile for the tracking problem, Brownian motion on Lie groups is employed as the generator to produce pose candidates as particles.
%
This paradigm is applicable to a diverse range of articulated objects, and for this reason it is referred to as \emph{Lie-X}.
(2) Learning based techniques are incorporated instead of the traditional Jacobian matrices for solving the incurred inverse kinematics problem, namely, presented with visual discrepancies of current results, how to improve on skeletal estimation results.
More specifically, an iterative sequential learning pipeline is proposed: multiple initial poses are engaged simultaneously to account for possible location, orientation, and size variations, with each producing its corresponding estimated pose. They then pass through a learned scoring metric to deliver the final estimated pose. Note the purpose of the learned metric in our approach is to mimic the behavior of the practical evaluation metric.
%
(3) Empirically our approach has demonstrated competitive performance on fish, mouse, and human hand from different imaging modalities, where it is also specifically referred to as e.g. Lie-fish, Lie-mouse, Lie-hand, respectively. The runtime speed of our pose estimation system is more than realtime --- it executes at around 83-267 FPS (frame per second) on a desktop computer without resorting to GPUs.
Moreover, new sets of annotated depth images and videos of articulated objects are created.
It is worth noting that the depth imaging devices considered in our empirical context are also diverse, including structured illumination and light field technologies, among others.
These datasets and our code are to be made publicly available in support of the open-source research
activities.~\footnote{Our datasets, code, and detailed information pertaining to the project can be found at a dedicated project webpage \url{http://web.bii.a-star.edu.sg/~xuchi/Lie-X.html}.}.

\section{Related Work} 


The recent introduction of commodity depth cameras has led to significant progress in analyzing articulated objects, especially human full-body and hand.
In terms of pose estimation, Microsoft Kinect is already widely used in practice at the scale of human full-body,
while it is still a research topic at human hand scale~\cite{TomEtAl:siggraph14,ObeWohLep:cvww15,XuEtAl:IJCV15,ObeWohLep:iccv15,TanEtAl:iccv15,TanEtAl:cvpr16,zhoumodel:ijcai16}, partly due to the dexterous nature of hand articulations.
\cite{TomEtAl:siggraph14} is among the first to develop a dedicated convolutional neural net (CNN) method for hand pose estimation, which is followed by~\cite{ObeWohLep:cvww15}. \cite{ObeWohLep:iccv15} also utilizes deep learning in a synthesizing-estimation feedback loop.
\cite{zhoumodel:ijcai16} further considers to incorporate geometry information in hand modelling by embedding a non-linear generative process within a deep learning framework.
\cite{XuEtAl:IJCV15} studies and evaluates a theoretically motivated random forest method for hand pose estimation.
A hierarchical sampling optimization procedure is adopted by~\cite{TanEtAl:iccv15} to minimize the error-induced energy functions, where a similar energy function is optimized via efficient gradient-based techniques in~\cite{TanEtAl:cvpr16} for personalizing hand shapes to individual users.
\cite{SinChoRam:cvpr16} instead casts hand pose estimation as a matrix completion problem with deeply learned features.
%
Meanwhile, various tracking methods have been developed for full-body~\cite{HuaEtAl:cvpr16} and hand~\cite{OikArg:bmvc11,BalEtAl:eccv12,QiaEtAl:cvpr14}.
A particle swarm optimization (PSO) scheme is utilized in~\cite{OikArg:bmvc11} to recover temporal hand poses by stochastically seeking solution to the induced minimization problem.
A hybrid method is adopted in~\cite{QiaEtAl:cvpr14} that combines PSO further with the widely-used iterated closest point technique.
\cite{BalEtAl:eccv12} considers learning salient points on fingers, for which an objective function is introduced to jointly take into account of edges, flow and collision cues.
\cite{HuaEtAl:cvpr16} describes a tracking-by-detection~\cite{AndRotSch:CVPR08} type method based on 3D volumetric representation.
%
3D action recognition has also drawn great amount of attentions lately~\cite{NieXioZhu:cvpr15,MahTod:cvpr16,VemChe:cvpr16,RahMia:cvpr16}.
For example, \cite{MahTod:cvpr16} tackles action recognition using variants of recurrent neural nets.
\cite{RahMia:cvpr16} considers a mapping to a view-invariant high-level space by CNNs and Fourier temporal pyramid.
Moreover, the work of~\cite{NieXioZhu:cvpr15} discusses a method to jointly train models for human full-body pose estimation and action recognition using spatial temporal and-or graphs.
%
%
On the other hand, it is a much harder problem when a color camera is used instead of a depth camera,
such as~\cite{AgaTri:pami06}, where pose estimation is formulated as a regression problem that is subsequently addressed by relevance vector machine and support vector machine.
%
Now, let us look at the other two articulated objects to be described in this paper, i.e. fish and mouse. They are relatively simple in nature but are less studied.
Existing literature~\cite{BraBel:cvpr05,DolEtAl:PETS05,DolWelPer:cvpr10} are mostly 2D-based, and the focus is mainly on pose estimation.
\cite{WiltAl:neruon15} is a very recent work in analyzing group-level behavior of lab mice that relies on a simplified straight-line representation of a mouse skeleton.
We also would like to point out that there are research efforts across object categories: \cite{DolWelPer:cvpr10} estimates poses of zebrafish, lab mouse, and human face;
Meanwhile there are also works that deal with more than one problem, such as~\cite{NieXioZhu:cvpr15}. They have achieved very promising results as discussed previously. Our work may be considered as a renewed attempt to address related problems and work with a broad range of articulated objects under one unified principle.
For more detailed overview of related works, interested readers may consult to the recent surveys~\cite{Pop:cviu07,CheWeiFer:prl13,PerEtAl:sensors14,Bar:arxiv16}.

Lie groups~\cite{Pro:book07} have been previously used in \cite{TuzPorMee:cvpr08} for detection and tracking of relatively rigid objects in 2D,
however this requires the expensive image warping operations.
\cite{SriTurKur:ivc12} reviews in particular the recent development of applying shape manifold based approaches in tracking and action recognition.
Its application in articulated objects is relatively sparse.
Lie algebraic representation is considered in~\cite{MikEtAl:ijcv03} for human full-body pose estimation based on multiple cameras or motion captured data.
Rather than resorting to the traditional Jacobian matrices as in~\cite{MikEtAl:ijcv03},
learning based modules are employed in our approach to tackle the the incurred inverse kinematics problem.
\cite{VemArrChe:cvpr14,VemChe:cvpr16} also extract Lie algebra based features for action recognition.
Instead of focusing on specific problem and object, here we attempt to provide a unified approach.

Part-based models have long been considered in the vision community, such as the pictorial structures~\cite{FelHut:ijcv05}, the flexible mixtures-of-parts~\cite{YanRam:cvpr11}, the poslet model~\cite{BouMal:iccv09}, the deep learning model~\cite{TomEtAl:nips14}, among others.
The most related works are probably~\cite{DolWelPer:cvpr10,SunEtal:cvpr15}, where the idea of group action has been utilized.
Moreover, multiple types of objects are also evaluated in~\cite{DolWelPer:cvpr10} that focuses on the 2D pose estimation problem, while \cite{SunEtal:cvpr15} is dedicated to 3D hand pose estimation.
On the other hand, our approach aims to address these three related problems altogether in 3D, and we explicitly advocate the usage of Lie group theory.
Note that the concept of pose indexed feature has been coined and employed in~\cite{FleGem:jmlr08,AliEtAl:iccv09}.
In addition, learning based optimization has been considered in e.g.~\cite{XioTor:cvpr13}, although in very different contexts.
Finally, the idea of learning the internal evaluation metric is conceptually related to the recent learning to rank approaches~\cite{BurEtAl:icml05}
in the information retrieval community for constructing ranking models.

\section{Notations and Mathematical Background}
The skeletal representation is in essence based on the group of rigid transformations in 3D Euclidean $\mathbb{R}^{3}$, a Lie group that is usually referred to as the special Euclidean group SE$(3)$.
In what follows, we provide an account of the related mathematical concepts that will be utilized in our paper.

\begin{figure*}[!t]
    \centering
	\includegraphics[width=0.9\textwidth]{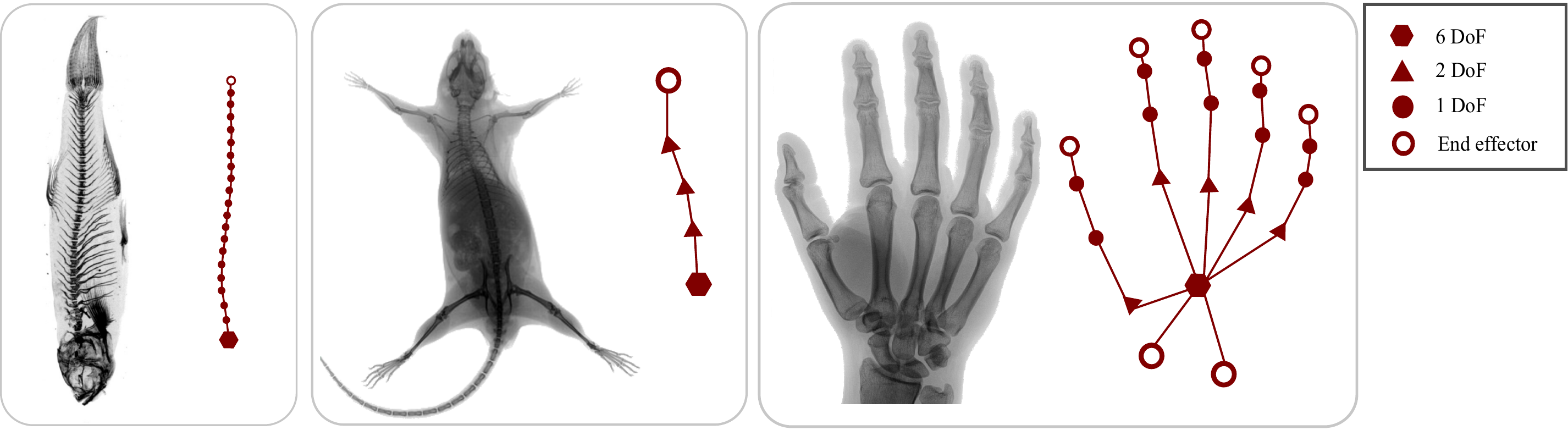}
    \caption
    {A display of three different articulated objects considered in our paper, which are for (from left to right) fish, mouse, and human hand, respectively. Each of our skeletal models is an approximation of the underlying anatomy, presented as a side-by-side pair. Note end-effectors have zero degree of freedom (DoF). See text for details.}
    \label{fig:concept}
\end{figure*}

An articulated object, such as a human hand, a mouse or a fish,
is characterized in our paper by a skeletal model in the form of a kinematic tree that contains one or multiple kinematic chains.
As illustrated in Fig.~\ref{fig:concept},
a fish or a mouse skeleton both possess one kinematic chain, while a human hand contains a kinematic tree structure of multiple chains.
Note that only the main spine is considered herein for the mouse model.
The skeletal model is represented in the form of
$J_o$ joints interconnected by a set of bones or segments of fixed lengths.
Empirical evidence has suggested that it is usually sufficient to use such fixed skeletal models with proper scaling,
when working with pose estimation of articulated objects in depth images~\cite{SunEtal:cvpr15}.
The pose of this object can thus be defined as a set of skeletal joint locations.
Furthermore, we define the home position of an articulated object as a set of default joint locations.
Taking a mouse model as depicted in the middle panel of Fig.~\ref{fig:concept} for example,
its home position could be a top-view upward-facing mouse with the full body straight-up, and the bottom joint at the coordinate origin.
Note this bottom joint contains 6 degrees of freedom (DoF) of the entire object, and is also referred to as the base joint.
Then the pose could also be interchangeably referred to as the sequence of SE$(3)$ transformations or group actions applied to the home position,
$\Theta = \{ \theta_1, \ldots, \theta_{J_o} \}$.
The estimated pose is denoted as $\tilde{\Theta} = \{ \tilde{\theta}_1, \ldots, \tilde{\theta}_{J_o} \}$ to better differentiate from the ground-truth pose.
Here $\theta$ could be either $\xi$ or $\hat{\xi}$ (to be discussed later) when without confusion in the context.
To simplify the notation, we assume a kinematic chain contains $J$ joints. Clearly $J=J_o$ for fish and mouse models, while $J<J_o$ for human hand or human full-body, by focusing on one of the chains.
A depth image is not only a 2D image but also a set of 3D points (i.e. a 3D point cloud) describing the surface profile of such object under a particular view.
Ideally the estimated pose (the set of predicted joint locations) is expected to align nicely with the 3D point cloud of the object in the observed depth image.
%


Before proceeding further with the proposed approach,
let us pause for a moment to have a concise review of the involved mathematical background. Motivated readers may refer to~\cite{MurSasLi:book94,Hsu:book02,Lee:book03} for further details.

\paragraph{Lie Groups}

A Lie group $G$ is a group as well as a smooth manifold such that the group operations of 
$(g, h) \mapsto gh$ and $g \mapsto g^{-1}$ are smooth for all $g, h \in G$.
For example, the rotational group SO$(3)$ is identified as the set of 3 $\times$ 3 orthonormal matrices $\big\{ R \in \mathbb{R}^{3 \times 3}: RR^T =\mathrm{I}_3, \mathrm{det}(R)=1 \big\}$,
with $R^T$ denoting the transpose, det$(\cdot)$ being the determinant, and $\mathrm{I}_3$ being a 3 $\times$ 3 identity matrix.
Another example is SE$(3)$, which is defined as the set of rotational and translational transformations 
of the form $g(\mathbf{x}) = R \mathbf{x} + \mathbf{t}$, with $R \in$ SO$(3)$ and $\mathbf{t} \in \mathbb{R}^3$. In other words, $g$ is the 4 $\times$ 4 matrix of the form
\begin{center}
\begin{align}
g = \left(
    \begin{array}{cc}
        R & \mathbf{t} \\
        \mathbf{0}^T & 1
    \end{array}
    \right),
\end{align}
\end{center}
where $\mathbf{0} = (0,0,0)^T$.
Note the identity element of $\mathrm{SE}(3)$ is the 4 $\times$ 4 identity matrix $\mathrm{I}_4$.
Both $\mathrm{I}_3$ and $\mathrm{I}_4$ will be simply denoted as $\mathrm{I}$ if there is no confusion in the context.
Now given a reference frame, a rigid-body transformation of two consecutive joints $\mathbf{x}$ and $\mathbf{x}'$ in a kinematic chain can be represented as
$
\begin{pmatrix} \mathbf{x}' \\ 1 \end{pmatrix} \leftarrow g \begin{pmatrix} \mathbf{x} \\ 1 \end{pmatrix}.
$
Moreover the product of multiple SE$(3)$ groups (i.e. a kinematic chain) is still a Lie group.
In other words, as tree-structured skeletal models are considered in general for articulated objects, each of the induced kinematic chains forms a Lie group.

\paragraph{Lie Algebras and Exponential Map}

The tangent plane of Lie group SO$(3)$ or SE$(3)$ at identity $\mathrm{I}$ is known as its Lie algebra, $\mathrm{so}(3) := T_I \mathrm{SO}(3)$ or $\mathrm{se}(3) := T_I \mathrm{SE}(3)$,
respectively. An arbitrary element of so$(3)$ admits a skew-symmetric matrix representation parameterized by a three dimensional vector $\omega=(\omega_1, \omega_2, \omega_3)^T \in \mathbb{R}^3$ as
\begin{align}
\hat{\omega} = \left(
    \begin{array}{ccc}
        0 & -\omega_3 & \omega_2\\
        \omega_3 & 0 & -\omega_1\\
        -\omega_2 & \omega_1 & 0
    \end{array}
    \right).
\end{align}
In other words, the DoF of a full SO$(3)$ is three.
Note that a rotational matrix can alternatively be represented by the Euler angles decomposition~\cite{MurSasLi:book94}.
A bijective mapping $\vee: so(3) \rightarrow \mathbb{R}^3$ and its reverse mapping $\wedge : \mathbb{R}^3 \rightarrow so(3)$ are defined
as $\hat{\omega}^{\vee} = \omega$, and $\omega^{\wedge} = \hat{\omega}$, respectively. Let $\nu \in \mathbb{R}^3$, an element of se$(3)$ can then be identified as
\begin{align}
\hat{\xi} = \left(
    \begin{array}{cc}
        \hat{\omega} & \nu \\
        \mathbf{0}^T & 0
    \end{array}
    \right).
\end{align}
With a slight abuse of notation, similarly there exist the bijective maps $\hat{\xi}^{\vee} = \xi$, and $\xi^{\wedge} = \hat{\xi}$, with $\xi = (\omega^T, \nu^T)^T$.
Now a tangent vector $\xi \in \mathbb{R}^6$ (or its matrix form $\hat{\xi} \in \mathbb{R}^{4 \times 4}$) is represented as $\hat{\xi} = \sum_{i=1}^6 \xi^{i} \partial_i$,
with $\xi^{i}$ indexing the $i$-th component of $\xi$.
Here $\partial_1 = (1, 0, \ldots, 0)^T$, $\ldots$, $\partial_6 = (0, \ldots, 0, 1)^T$, or in their respective matrix forms,
\\
$
\partial_1 = \begin{pmatrix} 0 & 0 & 0 & 0 \\ 0 & 0 & -1 & 0 \\ 0 & 1 & 0 & 0 \\ 0 & 0 & 0 & 0  \end{pmatrix},
$
$
\partial_2 = \begin{pmatrix} 0 & 0 & 1 & 0 \\ 0 & 0 & 0 & 0 \\ -1 & 0 & 0 & 0 \\ 0 & 0 & 0 & 0  \end{pmatrix},
$
$
\partial_3 = \begin{pmatrix} 0 & -1 & 0 & 0 \\ 1 & 0 & 0 & 0 \\ 0 & 0 & 0 & 0 \\ 0 & 0 & 0 & 0  \end{pmatrix},
$
\\
$
\partial_4 = \begin{pmatrix} 0 & 0 & 0 & 1 \\ 0 & 0 & 0 & 0 \\ 0 & 0 & 0 & 0 \\ 0 & 0 & 0 & 0  \end{pmatrix},
$
$
\partial_5 = \begin{pmatrix} 0 & 0 & 0 & 0 \\ 0 & 0 & 0 & 1 \\ 0 & 0 & 0 & 0 \\ 0 & 0 & 0 & 0  \end{pmatrix},
$
$
\partial_6 = \begin{pmatrix} 0 & 0 & 0 & 0 \\ 0 & 0 & 0 & 0 \\ 0 & 0 & 0 & 1 \\ 0 & 0 & 0 & 0  \end{pmatrix}.
$

The exponential map $\mathrm{Exp}: \mathrm{se}(3) \rightarrow \mathrm{SE}(3)$ in our context is simply
the familiar matrix exponential $\mathrm{Exp}_{I} (\hat{\xi}) = e^{\hat{\xi}} = I + \hat{\xi} + \frac{1}{2} \hat{\xi}^2 + \ldots$ for any $\hat{\xi} \in \mathrm{se}(3)$.
From the Rodrigues's formula, it can be further simplified as
\begin{align}
e^{\hat{\xi}}= \begin{pmatrix} e^{\hat{\omega}} & A \nu \\ \mathbf{0}^T & 1 \end{pmatrix},
\end{align}
with
\begin{align}
A = I + \frac{\hat{\omega}}{\|\omega\|^2} \big( 1- \cos \|\omega\| \big) + \frac{\hat{\omega}^2}{\|\omega\|^3} \big( \|\omega\| - \sin \|\omega\|  \big),
\end{align}
where $\|\cdot\|$ is the vector norm.

It has been known in the screw theory of robotics~\cite{MurSasLi:book94} that every rigid motion is a screw motion that can be realized as
the exponential of a \emph{twist} (i.e. a infinitesimal generator) $\hat{\xi}$,
with its components $\omega$ and $\nu$ corresponding to the angular velocity and translation velocity of the segment (i.e. bone) around its joint, respectively.

\paragraph{Product of Exponentials and Adjoint Representation}
Consider a partial kinematic chain involving $j$ joints with $j \in \{1, 2, \cdots, J\}$, which becomes the full chain when $j=J$.
With a slight abuse of notation, let $g_{\Theta_{1:j}}$ be the Lie group action on the partial kinematic chain, and $g_{\theta_j}$ or simply $g_j$ be the group action on the $j$-th joint.
Its forward kinematics can be naturally represented as the product of exponentials formula, $g_{\Theta_{1:j}} = e^{\hat{\xi}_1 } e^{\hat{\xi}_2 } \cdots e^{\hat{\xi}_{j} }$.
Therefore, for an end-effector from the home configuration (or home pose) $\begin{pmatrix} \mathbf{x} \\ 1 \end{pmatrix}$, its new configuration is described by
$
\begin{pmatrix} \mathbf{x}' \\ 1 \end{pmatrix} = g_{\Theta_{1:j}} \begin{pmatrix} \mathbf{x} \\ 1 \end{pmatrix}
= e^{\hat{\xi}_1 } e^{\hat{\xi}_2 } \cdots e^{\hat{\xi}_{j} }
\begin{pmatrix} \mathbf{x} \\ 1 \end{pmatrix}.
$
As discussed in~\cite{MurSasLi:book94},
this formula can be regarded as a series of transformations from the body coordinate $b$ (for local joint) of each joint to the spatial coordinate $s$ (for global kinematic chain).
Let us focus on a joint $j$ and denote $\hat{\xi}^{(b)}$ and $\hat{\xi}^{(s)}$ the twists of this joint in the body and spatial coordinates, respectively.
Assume the transformation of this joint to the spatial coordinate is $g_{\Theta_{1:j-1}} = e^{\hat{\xi}_1 } e^{\hat{\xi}_2 } \cdots e^{\hat{\xi}_{j-1} }$.
The two twists can be related by the adjoint representation
\begin{align*}
\hat{\xi}^{(s)} = \mathrm{Ad}_{g_{\Theta_{1:j-1}}} \left( \hat{\xi}^{(b)} \right) := g_{\Theta_{1:j-1}} \hat{\xi}^{(b)} g_{\Theta_{1:j-1}}^{-1},
\end{align*}
which is obtained by
\begin{align*}
e^{\hat{\xi}^{(s)}} = g_{\Theta_{1:j-1}} e^{\hat{\xi}^{(b)}} g_{\Theta_{1:j-1}}^{-1} = e^{g_{\Theta_{1:j-1}} \hat{\xi}^{(b)} g_{\Theta_{1:j-1}}^{-1}},
\end{align*}
and repeatedly applying the identity $g e^{\xi} g^{-1} = e^{g \xi g^{-1}}$ for $g \in \mathrm{SE}(3)$ and $\xi \in \mathrm{se}(3)$.

\paragraph{Geodesics}
Given two configurations $g_1$ and $g_2$, the geodesic curve between them is
$
g(\tilde{t}) = \begin{pmatrix} R(\tilde{t}) & A \mathbf{t}(\tilde{t}) \\ \mathbf{0}^T & 1 \end{pmatrix},
$
with $R(\tilde{t}) = R_1 e^{(\Omega_0 \tilde{t})}$, $\mathbf{t}(\tilde{t}) = (\mathbf{t}_2 - \mathbf{t}_1) \tilde{t} + \mathbf{t}_1$, $\tilde{t} \in [0,1]$, and $\Omega_0 = \mathrm{Log}_I (R_1^{-1} R_2 )$.
Here the logarithm map $\mathrm{Log}_I$ or its simplified notion $\log$ can be regarded as the inverse of the exponential map.

\paragraph{Brownian Motion on Manifolds}
We refer interested readers to~\cite{Hsu:book02} for a more rigorous account of Brownian motion and stochastic differential geometry as they are quite involved.
Here it is sufficient to know that Brownian motion can be regarded as a generalization of Gaussian random variables on manifolds,
where the increments are independent and Gaussian distributed, and the generator of Brownian motion is the Laplace-Beltrami operator.
In what follows we will focus more on the computational aspect~\cite{Man:jstsp13}.
Let $\tilde{t} \in \mathbb{R}$ denote a continuous variable, and $\delta >0$ be a small step size.
Let $\xi_{\tilde{t}}=(\xi_{\tilde{t}}^{1}, \cdots, \xi_{\tilde{t}}^{6})^T$ denote a random vector sampled from normal distribution $\mathcal{N}(0, C)$,
for $k=0, 1, \cdots$ with $C \in \mathbb{R}^{6 \times 6}$ being a covariance matrix.
Then a left-invariant Brownian motion with starting point $g(0) \in \mathrm{SE}(3)$ can be approximated by
\begin{align}
g \big((k+1) \delta \big) = g \big(k \delta \big) e^{ \big\{ \sqrt{\delta} \sum_{i=1}^6 \xi_{k}^{i} \partial_i \big\} }.
\label{eq:brownianMotion}
\end{align}
In addition, these sampled points can be interpolated by geodesics to form a continuous sample path. In other words, for $\tilde{t} \in \big( k \delta, (k+1)\delta \big)$ we have
\begin{align}
g (\tilde{t}) = g \big(k \delta \big) e^{ \big\{ \frac{\tilde{t} - k \delta}{\sqrt{\delta}} \sum_{i=1}^6 \xi_{k}^{i} \partial_i \big\} }.
\label{eq:brownianMotionGeodesic}
\end{align}

\section{Our Approach}

In what follows we describe the proposed \emph{Lie-X} approach for pose estimation, tracking, and action recognition of various articulated objects.

\paragraph{Preprocessing \& Initial Poses}
For simplicity we assume that there exists one and only one articulated object in an input depth image or patch.
A simple preprocessing step is employed in our approach to extract individual foreground objects of interest.
This corresponds to the point cloud of the object of interest extracted from the image.
We then estimate the initial 3D location of base joint as follows: The 2D location of the base joint is set as the center of the point cloud,
while its depth value is the average depth of the point cloud.
Initial poses are obtained by first setting these poses as the home pose of the underlying articulate object, i.e. bones of the object are straight-up for the three empirical applications.
For each of the initial poses, the initial orientation of the object is generated by perturbing the in-plane orientation of the above-mentioned base joint from a uniform distribution over $(-\pi, \pi)$.
To account for the size variations, the bone lengths of each initial pose estimate are also scaled by a scalar that is uniformly distributed in the range of [0.9, 1.1].


\paragraph{Skeletal Models}
After preprocessing, an initial estimated pose is provided for an input depth image.
The objects of interest are represented here in terms of kinematic chains.
Without loss of generality, in this paper we focus on fish and mouse that both have one chain, as well as human hand that possesses multiple chains, as depicted in the respective panels of Fig.~\ref{fig:concept}.
Our fish and mouse models contain 21 and 5 joints (including the end-effectors) along the main spine, respectively, while our hand model has 23 joints.
Their corresponding DoFs are 25, 12, and 26, respectively.
Overall our models are designed as proper approximations following the respective articulate objects' anatomies.
The base joint is fixed at coordinate origin which always has six DoFs describing 3D locations and orientations of the entire object;
One DoF joints are applied to the rest fish joints characterizing the yaw of fish bones; Two DoFs are used for the rest mouse joints to account for both yaw and pitch;
Similarly in our human hand model, two DoFs are used for each root joint of finger chain, while one DoFs are used for the rest joints. In all three models, zero DoFs are associated with the end-effectors, as each of them can entirely be determined by the preceding joints of the chain.
Note that although simplified, the mouse model includes the most essential components (joints of the spine) at a reasonable resolution in our study.
Our human hand model follows that of the existing literature (e.g.~\cite{ObeWohLep:iccv15,XuEtAl:IJCV15}) that works with the widely-used NYU hand depth image benchmark~\cite{TomEtAl:siggraph14}.

\begin{figure*}[!t]
	\centering
	\subfloat
    {
		\includegraphics[width=0.9\textwidth]{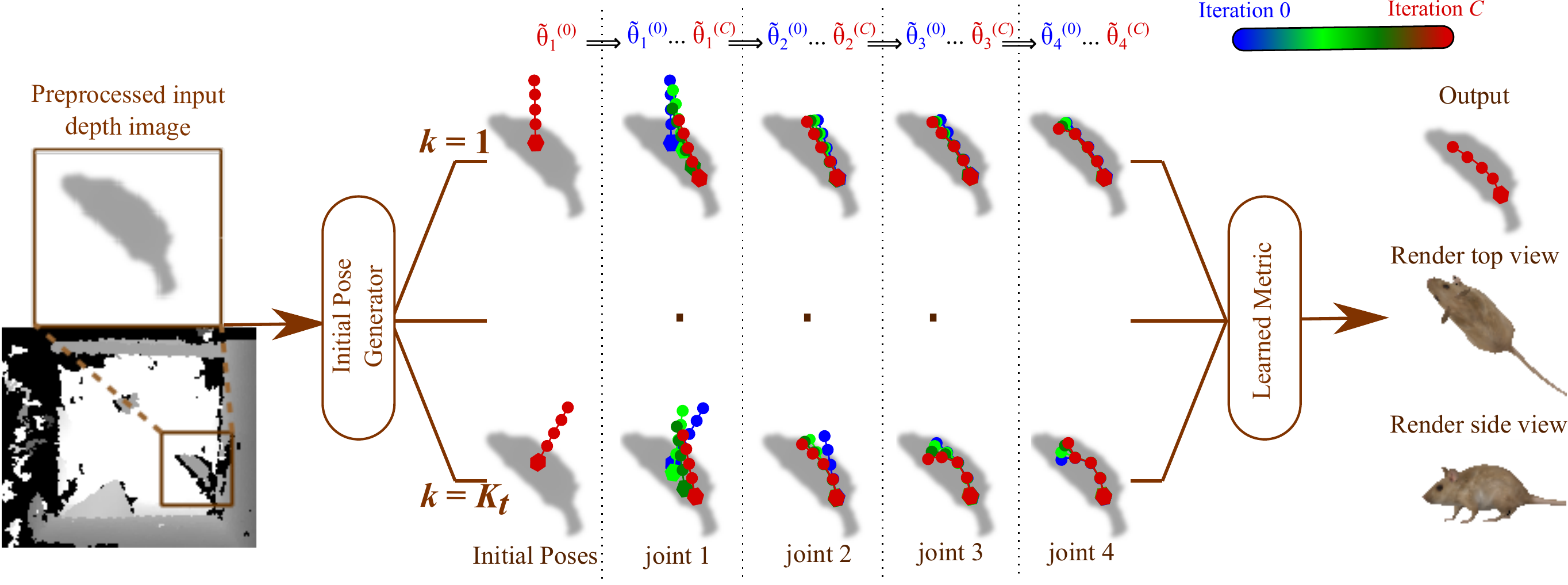} 
    }
	\caption
    {An illustrative example of our pose estimation pipeline in Algorithm 1.
In this example, a top-view depth image is used as the sole input. After a brief preprocessing and obtaining an initial pose,
an iterative process is executed over each joint $j$ and every round $c$, to produce its output estimate.
For demonstration purpose, we also present a 3D virtual mouse fitted with the predicted skeletal model and with triangle meshing and skin texture mapping, which is then rendered in its top-view as well as side-view. Note the limbs of this virtual mouse are pre-fixed to default configuration.}
    \label{fig:framework}
\end{figure*}

\subsection{Pose Estimation}

Fig.~\ref{fig:framework} provides a visual mouse example that illustrates the execution pipeline of our pose estimation procedure at test run. This is also presented more formally in  Algorithm~\ref{alg:PE_test}. Meanwhile the corresponding training process is explained in Algorithm~\ref{alg:PE_train}. Note that inside both the training and testing processes, an internal evaluation metric or scoring function is used, which is also learned from data. In what follows we are to explain each of the components in detail.

Given a set of $n_t$ training images, define
\begin{align}
\overline{\Delta \theta}_j := \frac{1}{n_t} \sum_{i \in \{1, \cdots, n_t\}} \Delta \theta_j
\end{align}
the mean deviation over training images for the $j$-th joint of the set of $J$ joints. The deviation $\Delta \theta_j$ characterizes the amount of changes between the estimated pose and the ground-truth pose, with its concrete form to be described later.
%
A global error function can be defined over a set of examples that evaluates the sum of differences from the mean deviation, as for example the following form,
\begin{align}
\sum_{j \in \{1, \cdots, J\}} \| \overline{\Delta \theta}_j \|_2^2,
\label{eq:entropy}
\end{align}
with $\| \cdot \|_2$ being the standard vector norm in Euclidean space. Presented with the above visual discrepancies of current results, our aim here is to improve skeletal estimation results.
Traditionally Jacobian matrices are employed for solving the incurred inverse kinematics problem.
Here we instead advocate the usage of an iterative learning pipeline as displayed in Fig.~\ref{fig:framework} with a mouse example.
At test run, it behaves as follows:
Assume for each of the $J$ joints there are $C$ rounds or iterations.
As presented in Algorithm~\ref{alg:PE_test}, given a test image and an initial pose estimation,
for each joint $j \in \{1, \cdots, J\}$ following the kinematic chain of length $J$ from the base joint, at current round $c \in \{1, \cdots, C\}$,
the current pose of the joint will be corrected by the Lie group action $e^{r_j^{(c)}}$, with the twist $r_j^{(c)}$ being the output of a local regressor,
$\mathcal{R}_{\mathrm{j}}^{(c)}$.
%
In other words, denote the short-hand notations $g^{(C)}_{\tilde{\Theta}_{1:j-1}} := g^{(1:C)}_{\tilde{\theta}_{1}} g^{(1:C)}_{\tilde{\theta}_{2}} \cdots g^{(1:C)}_{\tilde{\theta}_{j-1}}$, $e^{\hat{\xi}_j^{(1:c-1)}} := e^{\hat{\xi}_j^{(1)}} e^{\hat{\xi}_j^{(2)}} \cdots e^{\hat{\xi}_j^{(c-1)}}$, and $g_{\tilde{\Theta}_{1:j}}^{(c-1)} := g^{(1:C)}_{\tilde{\Theta}_{1:j-1}} e^{\hat{\xi}_j^{(1:c-1)}}$,
The $j$-th joint spatial coordinate can be updated by the following left group action
\begin{align}
g^{(c)}_{\tilde{\Theta}_{1:j}} = g_{\tilde{\Theta}_{1:j}}^{(c-1)} e^{r_j^{(c)}}, 
\label{eq:OneStepUpdate}
\end{align}
with $e^{r_j^{(c)}}$ being the latest group element used to further correct the spatial location of $j$-th joint at round $c$.
It is worth emphasizing that this process requires learning the set of local regressors $\left\{ \mathcal{R}_{\mathrm{j}}^{(c)} \right\}$, where the output of each regressor, $r_j^{(c)}$, is dedicated to a particular round $c$ and joint $j$. Each of these local models is learned based on local features, i.e. the pose-indexed depth features to be described later.
In a sense, it endows our system with the ability to memorize local gradient updating rules from similar training patterns.
This essentially forms the key ingredient that allows for removal of the commonly used Jacobian matrices for error-prorogation in our approach.
Moreover, at test run, multiple initial poses are generated for each input image.
They will then pass through our learned inverse kinematic regressors and produce corresponding candidate poses.
These output poses will nevertheless be screened by our learned metric to be discussed later, where the optimal one is to be picked as the final estimated pose.

At training stage, a set of $K$ initial poses of the input image is obtained in the same manner as those of the testing stage.
The aforementioned local regressors are then learned as follows.
Each example of the training dataset consists of an \emph{instance}:
a pair of poses including the estimated pose and the ground-truth pose, $(\tilde{\theta}_j^{(c-1)}, \theta_j)$,
as well as its \emph{label}: the deviation of estimation $\tilde{\theta}_j$ from ground-truth
$\theta_j$, as
\begin{align}
\Delta \theta_j = \log \big( {g^{(1:c-1)}_{\tilde{\Theta}_{1:j}}}^{-1} g_{\Theta_{1:j}} \big).
\end{align}
%
For the first joint $j=1$ (the base joint in the kinematic chain) and at the first round $c=1$,
the label of an example will be the amount of changes from the first joint of the initial pose to that of the ground-truth.
Then at any round $c$, its corresponding initial pose is obtained by executing the current partial kinematic model until the immediate previous round $c-1$.
Similarly for the second joint and at round $c$,
the initial pose in each of the training examples is attained by executing the current partial model from base joint up to round $c-1$ of the current joint,
and its label is then the amount of changes from the current joint of the aforementioned initial pose to the second joint of the ground-truth.
In this way, the training examples are prepared separately over joints and then across rounds until the very last joint $J$ $\&$ round $C$.
Algorithm~\ref{alg:PE_train} presents the procedure of learning the set of regressors $\left\{ \mathcal{R}_{\mathrm{j}}^{(c)} \right\}$, with each regressor
$\mathcal{R}_{\mathrm{j}}^{(c)}$ of round $c$ and joint $j$ being learned from its local context to make its prediction, $r_j^{(c)}$.
Without loss of generality the random forest method~\cite{Bre:ml01} is engaged here as the learning engine.

Note that our hand skeletal model contains five kinematic chains, all of which share the hand base joint as root of the tree.
For each chain, the sub-chain resulting from the exclusion of the base joint is independent of the other sub-chains given that the root is set.
This motivates us to consider the following procedure:
At test run, the base joint is first worked out, following the process described above. After this is done, Algorithm~\ref{alg:PE_test} is executed for each of the five sub-chains separately.

\begin{algorithm}[!t]
\caption{Pose Estimation: Testing Stage}
\label{alg:PE_test}
\begin{algorithmic}
	\STATE {\bfseries Input:} An unseen depth image
    \STATE {\bfseries Output:} Estimated skeletal joint locations and a prediction of its evaluation score 
    \STATE {Preprocessing to obtain $K_{\mathrm{t}}$ initial poses by random perturbation of the home pose centered at the object point cloud.}
    \FOR{ k=1:$K_{\mathrm{t}}$}
        \FOR{ j=1:J}
            \FOR{ c=1:C}
            \STATE{ Twist prediction by applying a learned local regressor $\mathcal{R}_{\mathrm{j}}^{(c)}: \big( \tilde{\theta}_j^{(c-1)}, \theta_j \big) \mapsto r_j^{(c)}$.}
            \STATE{ Update prediction of current joint spatial coordinate by applying the corresponding left group action of Eq.\eqref{eq:OneStepUpdate}. }
            \ENDFOR
        \ENDFOR
    \ENDFOR
    \STATE{ Pick the best out of $K_{t}$ candidates by applying the learned metric.}
\end{algorithmic}
\end{algorithm}

\begin{algorithm}[!t]
\caption{Pose Estimation: Training Stage}
\label{alg:PE_train}
\begin{algorithmic}
	\STATE {\bfseries Input:} The set of training examples. For each example $i$, obtain $K$ initial poses by random perturbations from the base system estimate. 
    \STATE {\bfseries Output:} a series of learned regressors $\{ \mathcal{R}_{\mathrm{j}}^{(c)}: j=1,\cdots, J; c=1,\cdots, C \}$.  
    \FOR{ j=1:J}
        \FOR{ c=1:C}
            \STATE{ Given the context, learn a local regressor $\mathcal{R}_{\mathrm{j}}^{(c)}$.} 
            \STATE{ Update prediction of current joint spatial coordinate by $\mathcal{R}_{\mathrm{j}}^{(c)}$ using Eq.\eqref{eq:OneStepUpdate}. }
        \ENDFOR
        \STATE{ Prepare the training set of $j+1$-th joint spatial coordinate by applying $\left\{ \mathcal{R}_{\mathrm{j'}}^{(c)} \right\}_{j'=1, c=1}^{j, C} $, the partial set of local regressors learned so far. }
    \ENDFOR
\end{algorithmic}
\end{algorithm}



\paragraph{Learning the Internal Evaluation Metric}
Since multiple pose hypotheses are presented in our approach, it remains to decide on which one from the candidate pool we should choose as the final pose estimate.
Traditionally this can be dealt with by either mode seeking or taking their empirical average as
in e.g. Hough voting methods~\cite{Hou:ICHEAI59,LeiLeoSch:eccv04whp,GalEtAl:pami11} or random forests~\cite{Bre:ml01}, respectively;
It could also be carried out by simply matching with a small set of carefully crafted templates such as distance transform or DOT~\cite{HinEtAl:cvpr10}.
Instead we propose to learn a surrogate scoring function that is to be consistent with the \emph{real} evaluation metric employed during practical quantitative analysis.
This learned scoring function then becomes a built-in module in our approach to select the pose hypothesis with the least error.

More concretely, the widely used criteria of average joint error~\cite{XuChe:iccv13} is adopted as the evaluation metric for our scoring function to mimic.
During training stage, a set of $n_{m}$ training examples are generated,
where a training example consists of an instance and a label: A training instance contains an input depth image,
its ground-truth pose (i.e. skeletal joint locations) and an estimated pose as a set of corresponding joint locations after random perturbations from the ground-truth.
Its label is the average joint error between the estimated and the ground-truth poses.
Therefore a second type of regressor, $\mathcal{R}_{\mathrm{m}}$, is utilized here to learn to predict the error at test run. Namely, given an unseen depth image and an estimated pose, our regressor would produce a real-valued score mimicking the average joint error as where the ground-truth is known.

\subsection{Tracking}

\begin{figure}[!t]
\includegraphics[width=0.5\textwidth]{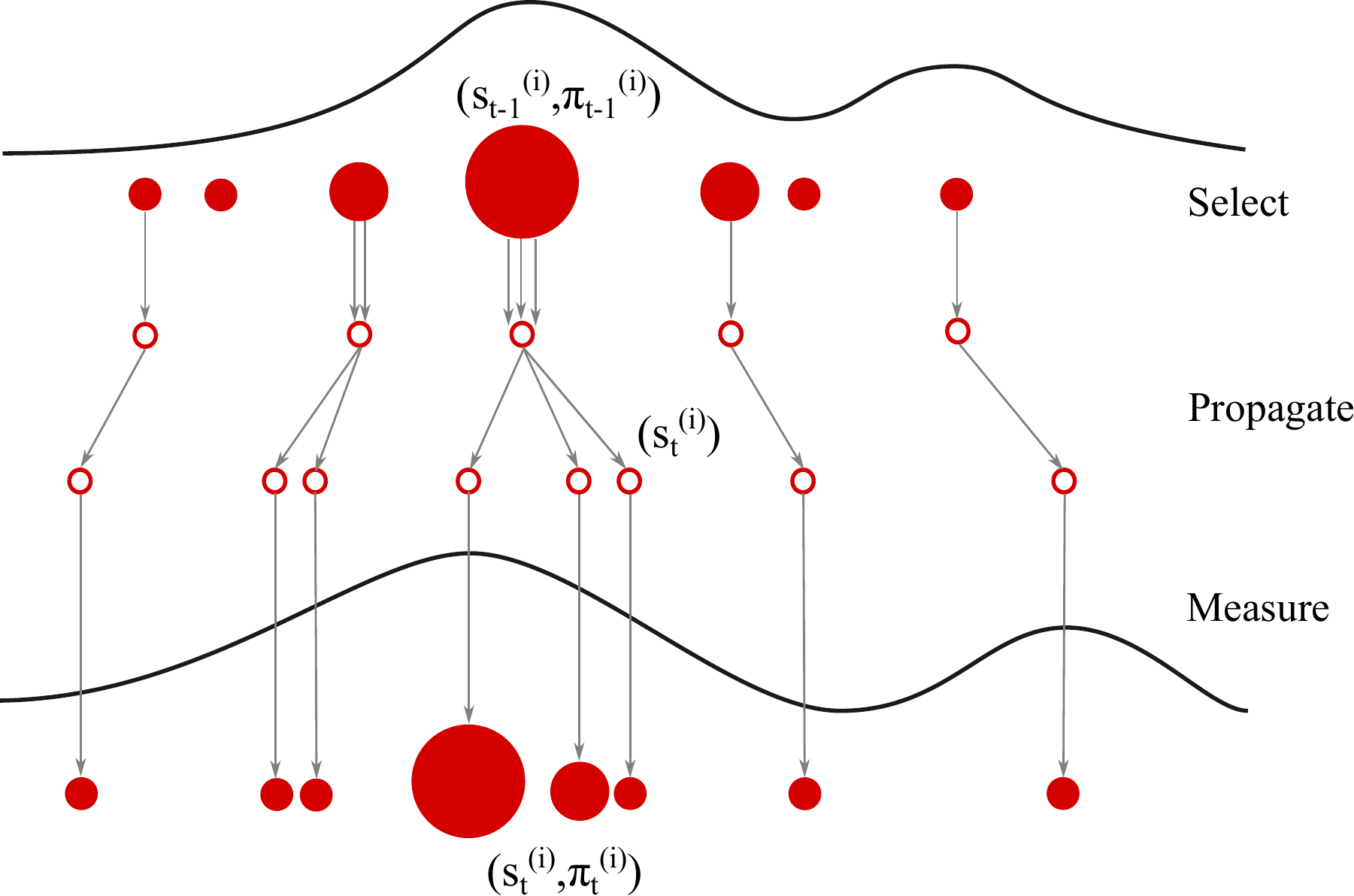}
\caption{A visual illustration of one time-step update process of the particle filter paradigm considered in our tracking task.}
\label{fig:track}
\end{figure}

Particle filters such as~\cite{isard1998condensation} have long been regarded as a powerful mean for tracking, and is also considered in our context to address the tracking problem. To facilitate a favorable balance between efficiency and effectiveness, we consider a probabilistic particle filter based approach only for the base joint, where particles are formed by Brownian motion based sampling in the pose manifold; Meanwhile the parameters of the remaining joints are obtained by invoking the same inference machinery as in our deterministic pose estimation algorithm. This design is also motivated from our empirical observation that often the object poses are also well-estimated when the prediction of the base joint is in close vicinity of the true values. That is, according to our observation the first joint 
is crucial in pose estimation:
If $\xi_{1}$ is wrongly predicted, estimation results of the remaining joints could be seriously damaged;
When our prediction of $\xi_{1}$ is accurate, the follow-up joints estimates would also be accurate.
Algorithm~\ref{alg:3Dtracking} further presents the main procedure for our tracking task, which is also visually illustrated in Fig.~\ref{fig:track}, with a detailed description in the following paragraphs. 

Following the particle filter paradigm~\cite{isard1998condensation} we consider a discretized time step $t$, and use
$x$ to denote a latent random variable as well as $y$ for its observation. Here the state of a tracked object (i.e. the estimated pose $\tilde{\Theta}$ at time $t$) is denoted as $x_{t}$ and its history is $x_{1:t}=\left( x_1,\cdots,x_t \right)$.
Similarly, current observation is denoted as $y_{t}$, and its history as $y_{1:t}=\left( y_1,\cdots,y_t \right)$.
The underlying first-order temporal Markov chain induces conditional independence property, which by definition gives $p(x_t|x_{1:t-1})=p(x_t|x_{t-1})$.
Following the typical factorization of this state-space dynamic model, we have
\begin{align*}
p(y_{1:t-1},x_t|x_{1:t-1}) &= p(x_t|x_{1:t-1}) p(y_{1:t-1}|x_{1:t-1}) \\
                           &= p(x_t|x_{t-1}) \prod_{i=1}^{t-1} p(y_i | x_i),
\end{align*}
with $p(y_{1:t-1}|x_{1:t-1}) = \prod_{i=1}^{t-1} p(y_i | x_i)$.
We also need the posterior probability $p(x_t|y_{1:t})$ for filtering purpose, which in our context is defined as
$p(x_t|y_{1:t}) \propto p(y_t|x_t)p(x_t|y_{1:t-1})$,
with
$p(x_t|y_{1:t-1}) = \int{p(x_t|x_{t-1})p(x_{t-1}|y_{1:t-1}) dx_{t-1}}$. In other words, it is evaluated by considering the posterior $p(x_{t-1}|y_{1:t-1})$ from the previous time step in a recursively manner. 

The realization of the particle filter paradigm in our context involves the three-step probabilistic inference process of selection-prorogation-measurement, which serves as the one time-step update rule in particle filter, and is also described in Algorithm~\ref{alg:3Dtracking}.
Specifically, the process at current time-step $t$ corresponds to an execution of the selection-prorogation-measurement triplet steps:
The output of previous time-step contains a set of $K_r$ weighted particles
\begin{align*}
S_{t-1} := \left\{ (s_{t-1}^{(i)}, \pi_{t-1}^{(i)} ) \right\}_ {i=1}^{K_r}.
\end{align*}
Here each particle $i$, $s_{t-1}^{(i)}$, corresponds to a particular realization of the set of tangent vector parameters $\tilde{\Theta}^{(i)}$ that uniquely determines a pose, where each of the vectors is attached to a joint following the underlying kinematic chain. The particle $s_{t-1}^{(i)}$ is also associated with its weight $\pi_{t-1}^{(i)} \in [0,1]$. Collectively this set of weighted particles is thus regarded as an approximation to the posterior distribution $p(x_{t-1}|y_{1:t-1})$.
The \emph{selection} step operates by uniform sampling from the cumulative distribution function (CDF) of $p(x_{t-1}|y_{1:t-1})$ to produce a new set of $K_r$ particles with equal weights. It is followed by the \emph{propagation} step where the manifold-based Brownian motion sampling of Eq.\eqref{eq:brownianMotionGeodesic} is employed to realize $p(x_t|x_{t-1})$, i.e. to obtain the new state based on discretized Brownian motion deviation from the previous time-step.
Note that this Brownian motion sampling is carried out only on the base joint, while the remaining joints are obtained by directly executing the same inference machinery of Eq.\eqref{eq:OneStepUpdate} as in our pose estimation algorithm.
Now, the sample set constitutes an approximation to the predictive distribution function of $p(x_t|y_{1:t-1})$. The \emph{measurement} step finally provides an updated weight $\pi_{t}^{(i)}$ for each particle $s_{t}^{(i)}$ as follows:
Let $m_t^{(i)}$ be the predicted error value of the $i$-th particle $s_t^{(i)}$, obtained by applying our learned metric.
The weight is thus evaluated as
\begin{align}
\pi_t^{(i)} = \frac{1}{\sqrt{2\pi}\sigma}e^{-\frac{{m_t^{(i)}}^2}{2\sigma^2}}.
\label{eq:pi_evaluate}
\end{align}
After obtaining all the $K_r$ weights, each of the weights, $\pi_t^{(i)}$, is further normalized as
\begin{align}
\pi_t^{(i)} \leftarrow \frac{\pi_t^{(i)}} {\sum_{i'=1}^{K_r} \pi_t^{(i')}}.
\label{eq:pi_normalize}
\end{align}
The updated sample set now collectively approximates the corresponding posterior distribution $p(x_{t}|y_{1:t})$ at time $t$.

The set of weighted particles allows us to represent the entire distribution instead of a point estimate as what we have done during the pose estimation task.
The final pose estimate, $x^*_t$ (i.e. $\tilde{\Theta}$ at time $t$), is produced by weighted averaging over this set of particles,
\begin{align}
x^*_t \leftarrow \sum_{i=1}^{K_r} \pi_{t}^{(i)} s_t^{(i)}.
\label{eq:tracking_finalPose}
\end{align}


\begin{algorithm}[!t]
\caption{Tracking at time-step $t$}
\label{alg:3Dtracking}
\begin{algorithmic}
	\STATE {\bfseries Input:} $S_{t-1}$
    \STATE {\bfseries Output:} $x^*_t, S_t$

    \STATE {\textbf{(1) Select:}}
    \STATE {calculate the normalized cumulative probabilities:}
    \FOR {$i = 1 \cdots K_r$}
	    \STATE {Sample a particle $s'^{(i)}_t$ uniformly from the CDF of $p(x_{t-1}|y_{1:t-1})$.}
	\ENDFOR

	
	\STATE {\textbf{(2) Propagate:}}
	\FOR {$i = 1 \cdots K_r$}
	    \STATE {Obtain $s^{(i)}_t$ by sampling from $\left\{s'^{(i)}_t \right\}$ using the transition probability $p(x_t|x_{t-1})$, which is realized by manifold-based Brownian motion sampling of Eq.\eqref{eq:brownianMotionGeodesic} of the tangent vector for the base joint, $\xi_{1}$, followed by directly executing Eq.\eqref{eq:OneStepUpdate} for each of the remaining joints following the kinematic chain.}
	\ENDFOR
	
	\STATE {\textbf{(3) Measure:}}
	\FOR {$i = 1 \cdots K_r$}
		\STATE {Evaluate $\pi_t^{(i)}$ by Eq.\eqref{eq:pi_evaluate}.}
	\ENDFOR
	\STATE {normalize $\pi_t^{(i)}$ by Eq.\eqref{eq:pi_normalize}. Now $S_{t} = \left\{ (s_{t}^{(i)}, \pi_{t}^{(i)} ) \right\}_ {i=1}^{K_r}$ is ready.}
	
	\STATE {\textbf{(4) Estimate} the pose}
	\STATE {The estimated pose $x^*_t$ is finally obtained by Eq.\eqref{eq:tracking_finalPose}.}
\end{algorithmic}
\end{algorithm}


\subsection{Action Recognition}
Our approach can be further employed to work with the problem of action recognition. 
The key insight is that an action instance (i.e. a pose sequence) corresponds to a curve segment in the manifold, whereas the set of all instances of a particular action type corresponds to a group of curves.
Therefore, the task of action recognition can be cast as separating different groups of action curves.
It motivates us to consider a third type of learned predictor, $\mathcal{R}_{\mathrm{a}}$.
Here dedicated features are extracted as to be described next, and the output concerns that of predicting its action categories.

\paragraph{Action Recognition Features}
As the length of action sequences may vary, they are firstly normalized to the same length (in practice 32 frames) using linear interpolation.
Local features from each frame of a sequence can be obtained based on the tangent vectors (Lie algebras) of the estimated joints in the manifold.
Each temporal sequence is further split into 4 equal-length segments, where the frames in a segment collectively contain the set of tangent vectors as local features.
Moreover, a temporal pyramid structural representation is utilized in a sense similar to that of the spatial pyramid matching~\cite{LazSchPon:cvpr06},
where features are extracted using hierarchical scales of \{4,2,1\}, where $4$ corresponds to the 4 segments introduced previously, and the rest correspond to those coarser scales built over it layer by layer. In total it leads to 7 temporal segments (or sets of variable sizes) over these 3 scale spaces.
For each such segment, the mean and standard deviation of its underlying pose representation (in terms of Lie algebras to be described below) are then used as features.
Now let us investigate the details of these pose representations defined on single frames, which can be decomposed into joints following the aforementioned kinematic chain structure.
For the base joint, we use the Lie algebras of the transformation from the current frame to the next frame.
For the rest of the joints, we use the Lie algebras of the transformation from previous joint to the current joint and that of the transformation from current frame to the next frame.
Besides, we also use the 3D location and orientation 
of the first joint (i.e. base joint) 
as features.
In particular for fish-related actions, rather than using the Lie algebras of all 20 joints,
we emphasize on robust estimation by considering a compact feature representation:
The first component or sub-vector of the feature vector corresponds to the Lie algebra of the base joint;
The second and the third components are the sub-vectors of the same length obtained by averaging over the set of Lie algebras from second to tenth joints, and from eleventh to the last joint, respectively.
Overall a 
252-dim feature vector is thus constructed to fully characterize an action sequence.

\begin{figure*}[!t]
    \centering
    \subfloat[]{
        \includegraphics[width=0.45\textwidth]{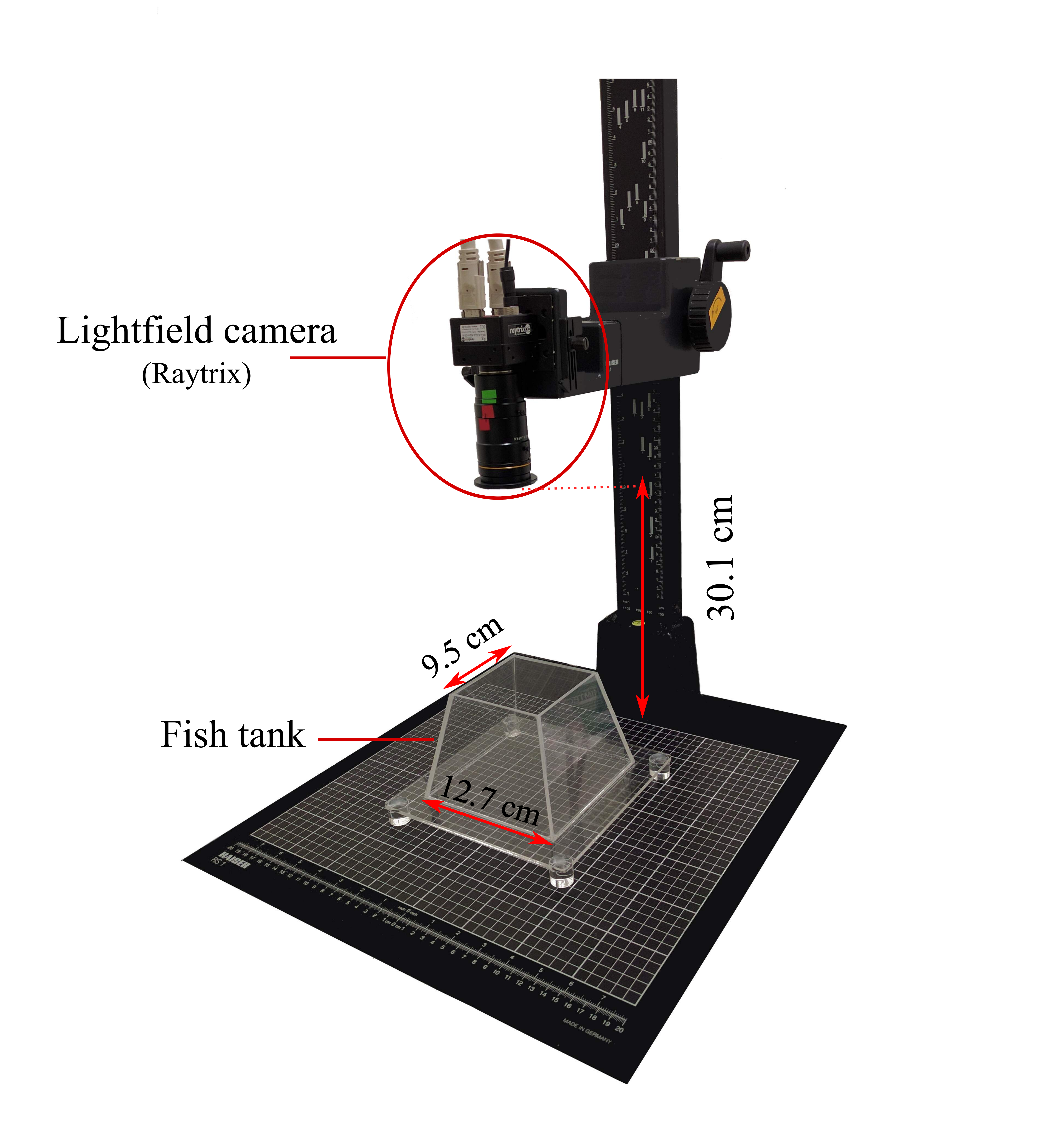}
        }
    \subfloat[]{
	    \includegraphics[width=0.45\textwidth]{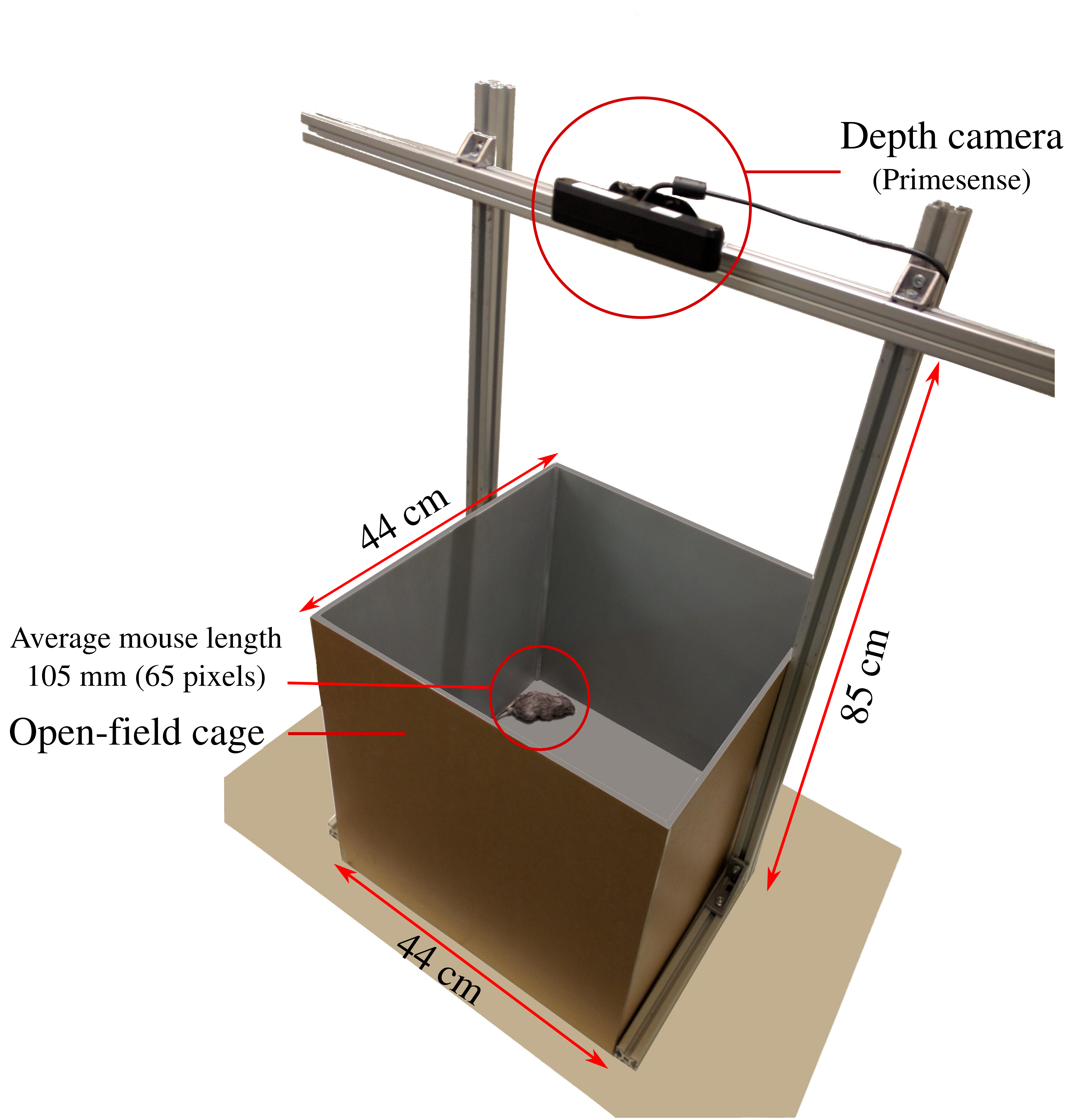}
    		}
    \caption{The capture setups used in constructing our (a) fish and (b) mouse depth image datasets, respectively.}
    \label{fig:capture_setup}
\end{figure*}


\subsection{Random Forests, Pose-indexed Depth features, and Binary Tests}
There are three types of learned predictors
(namely the set of local regressors $\left\{ \mathcal{R}_{\mathrm{j}}^{(c)} \right\}$, the learned internal metric $\mathcal{R}_{\mathrm{m}}$,
and the action recognition predictor $\mathcal{R}_{\mathrm{a}}$) considered in our approach.
In general any reasonable learning method can be used to realize these three types of predictors.
In practice the random forest method~\cite{Bre:ml01,GalEtAl:pami11} is engaged for these learning tasks, so it is worthwhile to describe its details here.

For action recognition, a unique set of action features are used as stated previously.
In what follows, we thus focus on the description of our pose-indexed depth features, which are used in the first two types of regressors,
$\left\{ \mathcal{R}_{\mathrm{j}}^{(c)} \right\}$, and $\mathcal{R}_{\mathrm{m}}$.
Our feature representation can be regarded as an extension of the popular depth features as discussed in~\cite{ShoEtAl:pami13,XuChe:iccv13}
by incorporating the idea of pose-indexed features~\cite{FleGem:jmlr08,AliEtAl:iccv09} to model 3D objects.
We start by focusing on a joint $j$ with its current 3D location $\mathbf{x} \in \mathbb{R}^3$, where a 3D offset $\mathbf{u}$ can be obtained
by random sampling from the home pose.
Let $g_{\tilde{\Theta}_{1:j}} (\mathbf{u})$ denote the Lie group left action of current object pose $\tilde{\Theta}_{1:j}$ applied onto $\mathbf{u}$. 
Now the 3D location of the offset is naturally $\mathbf{x}+g_{\tilde{\Theta}_{1:j}} (\mathbf{u})$,
and its projection onto 2D image plane under current camera view is denoted as $\bar{\mathbf{u}} = \mathrm{Proj} \big( \mathbf{x}+g_{\tilde{\Theta}_{1:j}} (\mathbf{u}) \big)$.
Similarly we can obtain another random offset $\bar{\mathbf{v}}$.
For a 2D pixel location $\bar{\mathbf{x}} = \mathrm{Proj} \big( \mathbf{x} \big) \in \mathbb{R}^2$ of an image patch $I$ containing the object of interest,
its depth value can be denoted as $d_I (\bar{\mathbf{x}})$.
Now we are ready to construct a feature $\phi_{I, (\bar{\mathbf{u}}, \bar{\mathbf{v}})} (\bar{\mathbf{x}})$ or its short-hand notation $\phi$,
by considering two 2D offsets positions $\bar{\mathbf{u}}, \bar{\mathbf{v}}$ from $\bar{\mathbf{x}}$:
\begin{align}
\phi_{I, (\bar{\mathbf{u}}, \bar{\mathbf{v}})} (\bar{\mathbf{x}}) = d_I \Big( \bar{\mathbf{x}} + \bar{\mathbf{u}} \Big) - d_I \Big( \bar{\mathbf{x}} + \bar{\mathbf{v}} \Big).
\label{eq:depthFeat}
\end{align}
Due to the visibility constraint, we are only able to obtain the depth values of the projected 2D locations $\bar{\mathbf{u}}$ and $\bar{\mathbf{v}}$ from the object surface.
Thus $\mathrm{Proj}$ is a surjective map. Nevertheless, this serves our intention of sampling random features well. 
%
Following~\cite{Bre:ml01}, a binary test is defined as a pair of elements, $(\phi, \epsilon)$, with $\phi$ being the feature function,
and $\epsilon$ being a real-valued threshold. When an instance with pixel location $\mathbf{x}$ passes through a split node of our binary trees,
it will be sent to the left branch if $\phi(\mathbf{x})>\epsilon$, and to the right side otherwise.

Our random forest predictors are constructed based on these features and binary tests for split nodes.
Similar to existing regression forests in literature including e.g.~\cite{ShoEtAl:pami13}, at a split node,
we randomly select a relatively small set of $m$ distinct features $\Phi := \{\phi_i\}_{n=1}^m$ as candidate features.
For every candidate feature, a set of candidate thresholds $\Lambda$ is uniformly selected over the range defined by the empirical set of training examples in the node.
The best test $(\phi^*, \epsilon^*)$ is chosen from these features and accompanying thresholds by maximizing the following gain function.
This procedure is then repeated until there are $L$ levels in the tree \textit{or} once the node contains fewer than $l_n$ training examples.
More specifically, the above-mentioned split test is obtained by
\begin{align*}
(\phi^*, \epsilon^*)=\arg\max_{\phi \in \Phi, \epsilon \in \Lambda} \mathcal{I} (\phi, \gamma),
\end{align*}
where the gain $\mathcal{I}(\phi, \epsilon)$ is defined as:
\begin{align}
\mathcal{I}(\phi, \epsilon) = E(S) - \bigg( \frac { \left| S_l \right| }{ \left| S \right| } E(S_l)  + \frac { \left| S_r \right| }{ \left| S \right| } E(S_r) \bigg).
\label{eq:informationGain}
\end{align}
Here $|\cdot|$ denotes the cardinality of the set, $S$ denotes the set of training examples arriving at current node,
which is further split into two subsets $S_l$ and $S_r$ according to the test $(\phi, \epsilon)$.
Define $\overline{\Delta \theta}_j := \frac{1}{\|S\|} \sum_{i \in S} \Delta \theta_j$ the mean deviation of the set to $j$-th joint, and accordingly for $S_l$ and $S_r$.
The function $E$ is defined over a set of examples that evaluates the sum of differences from the mean deviation:
\begin{align}
E(S)= \sum_{i \in S} \| \overline{\Delta \theta}_j \|_2.
\label{eq:entropy_1}
\end{align}

In the final decision stage, for the first two regression modules, the mean-shift mode searching in Hough voting space is used as e.g. in~\cite{GalEtAl:pami11},
while for the third module (action recognition) the classical random forest strategy~\cite{Bre:ml01} is used to pick the category with largest counts from the averaged histogram.

\section{Empirical Evaluations}

Empirically our \emph{Lie-X} approach is examined on three different articulated objects: fish, mouse, and human hand.
%
%

\paragraph{Performance Evaluation Metric}
Our performance evaluation metric is based on the commonly-used \emph{average joint error}, computed as the averaged Euclidean distance in 3D space over all the joints.
Formally, let $\mathbf{v}_g$ and $\mathbf{v}_e$ be the ground-truth and estimated joint locations, respectively. The joint error of the pose estimate $\mathbf{v}_e$ is defined as $e = \frac{1}{m}\sum_{i} \| v_{gi} - v_{ei} \|$, where $\| \cdot \|$ is the Euclidean norm in 3D space.
When dealing with test images, let $k=1,\ldots, n_{\mathrm{tst}}$ index over the test images, and their corresponding joint errors denoted as $\{e_1, \cdots, e_{n_{\mathrm{tst}}}\}$.
The \emph{average joint error} is then defined as $\frac{1}{n_{\mathrm{tst}}}\sum_{j} e_j$.

\paragraph{Internal Parameters}
Throughout experiments, a fixed set of values is always used for the internal parameters of our approach, unless otherwise stated, as follows.
For the the first type of regressors (namely the set of local regressors $\left\{ \mathcal{R}_{\mathrm{j}}^{(c)} \right\}$),
the number of trees is fixed to (3, 10, 10), while the tree depth is (24, 24, 24) for the triplet of articulated objects (fish, mouse, hand), respectively.
For the second type (the learned internal metric $\mathcal{R}_{\mathrm{m}}$),
the number of trees is (20, 20, 20), while the tree depth is (15, 15, 20) for the triplet of articulated objects (fish, mouse, hand), respectively.
For the third type (the action recognition predictor) $\mathcal{R}_{\mathrm{a}}$, the number of trees in the forest is 50, and tree depth is 20.
The number of features is $m$=8,000, and the maximum number of examples in the leaf node is $l_n$=5.
The number of rounds at each joint is set to $C$=7, 3, and 3, for fish, mouse, and hand, respectively.
The local image patch sizes considered in our approach for fish, mouse, and hand are normalized to $25 \times 25$, $100 \times 100$, $100 \times 100$ mm$^2$, respectively. These patches are used as input to the local random forest regressors in our approach to estimate the spatial coordinate of next joint based on current joint following the kinematic chain.
%
One important parameter is $K_{\mathrm{t}}$, the number of initial poses in pose estimation.
In practice, after factoring-in the efficiency consideration, $K_{\mathrm{t}}$ is set to 40, 40, 20 for pose estimation of fish, mouse, and hand, respectively, throughout experiments.
Similarly, the number of initial poses for tracking is set to $K_{\mathrm{r}}$=200.
For learning the internal evaluation metric, the number of candidates is set to 8.
Namely, given a training dataset of size $n_t$, the number of training examples for pose estimation becomes $K_{\mathrm{t}} \times n_t$,
while the number of training examples for learning the internal metric is  $n_{m} = 8 \times n_t$.
%
%
%

\begin{figure}
\centering
\includegraphics[width=0.35\textwidth]{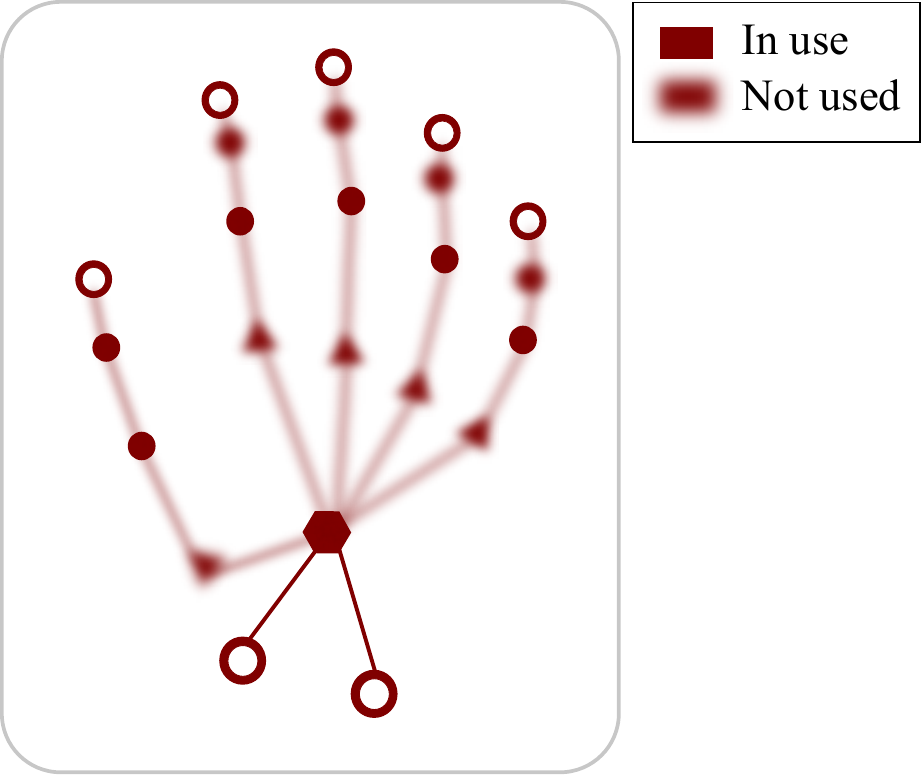}
\caption{Following the evaluation protocol of~\cite{TomEtAl:siggraph14,ObeWohLep:cvww15,ObeWohLep:iccv15,zhoumodel:ijcai16}, for NYU hand depth image dataset, only a subset of 14 joints out of the total 23 hand skeletal joints is considered during performance evaluation for hand pose estimation.}
\label{fig:nyu_hand_model}
\end{figure}

\begin{figure*}[!t]
    \centering
    \begin{tabular} {>{\centering\arraybackslash}m{.15\linewidth}@{}c@{}c@{}c}
    & Fish & Mouse & Hand
    \\
    Number of trees&
    \subfloat {
		\raisebox{-.5\height}{\includegraphics[width=0.22\textwidth]{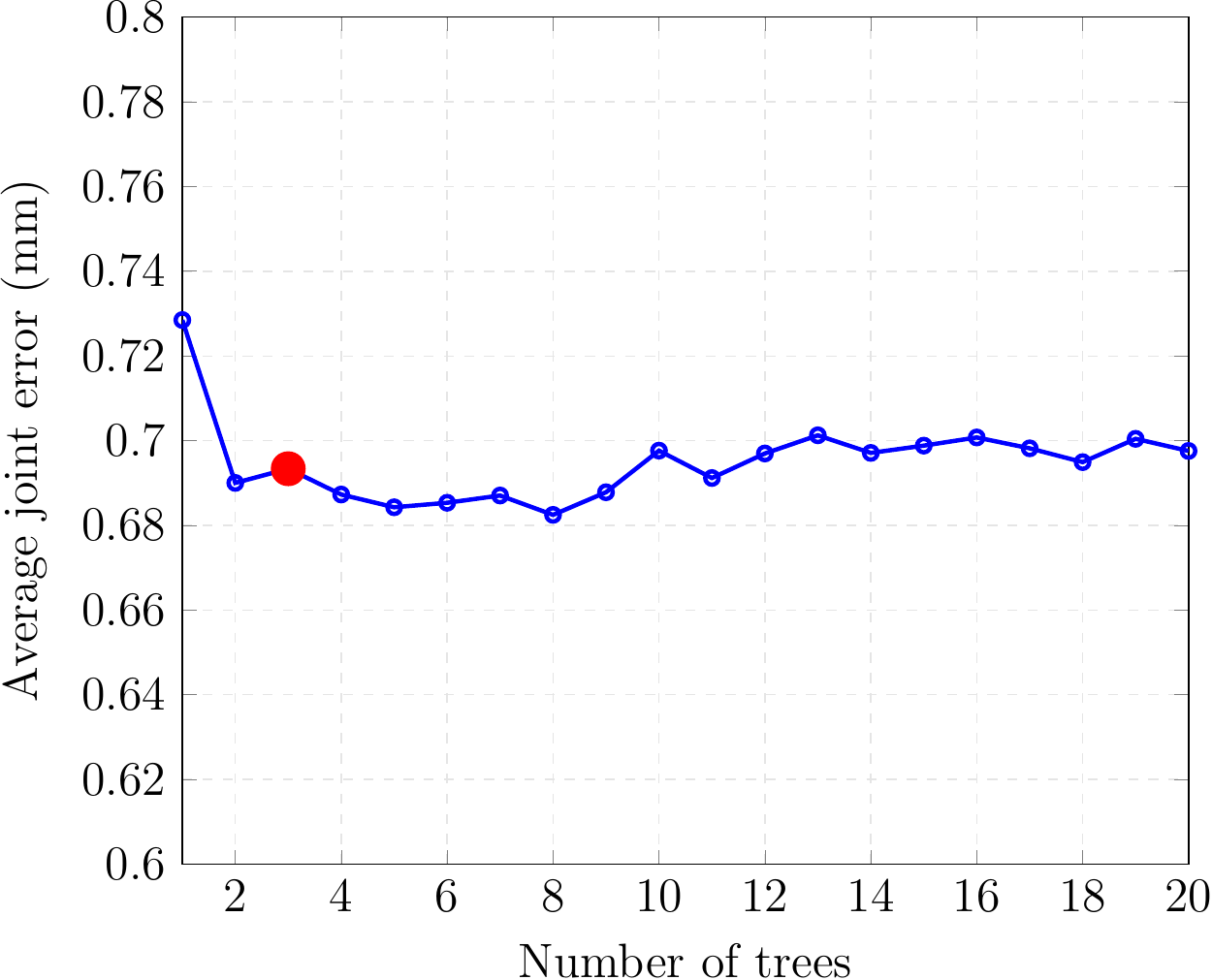}}
    }
	&
	\subfloat {
		\raisebox{-.5\height}{\includegraphics[width=0.22\textwidth]{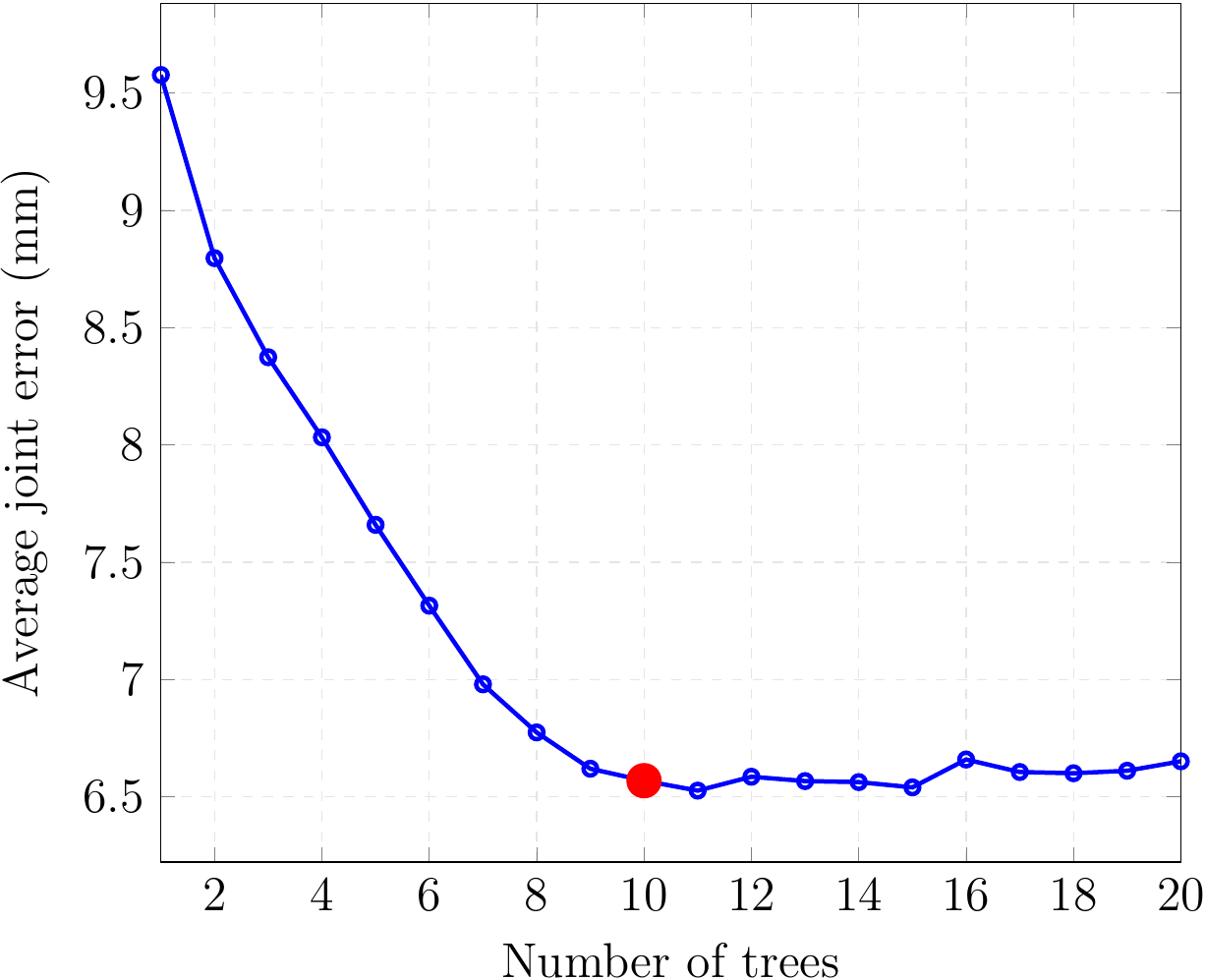}}
    }
	&
	\subfloat {
		\raisebox{-.5\height}{\includegraphics[width=0.22\textwidth]{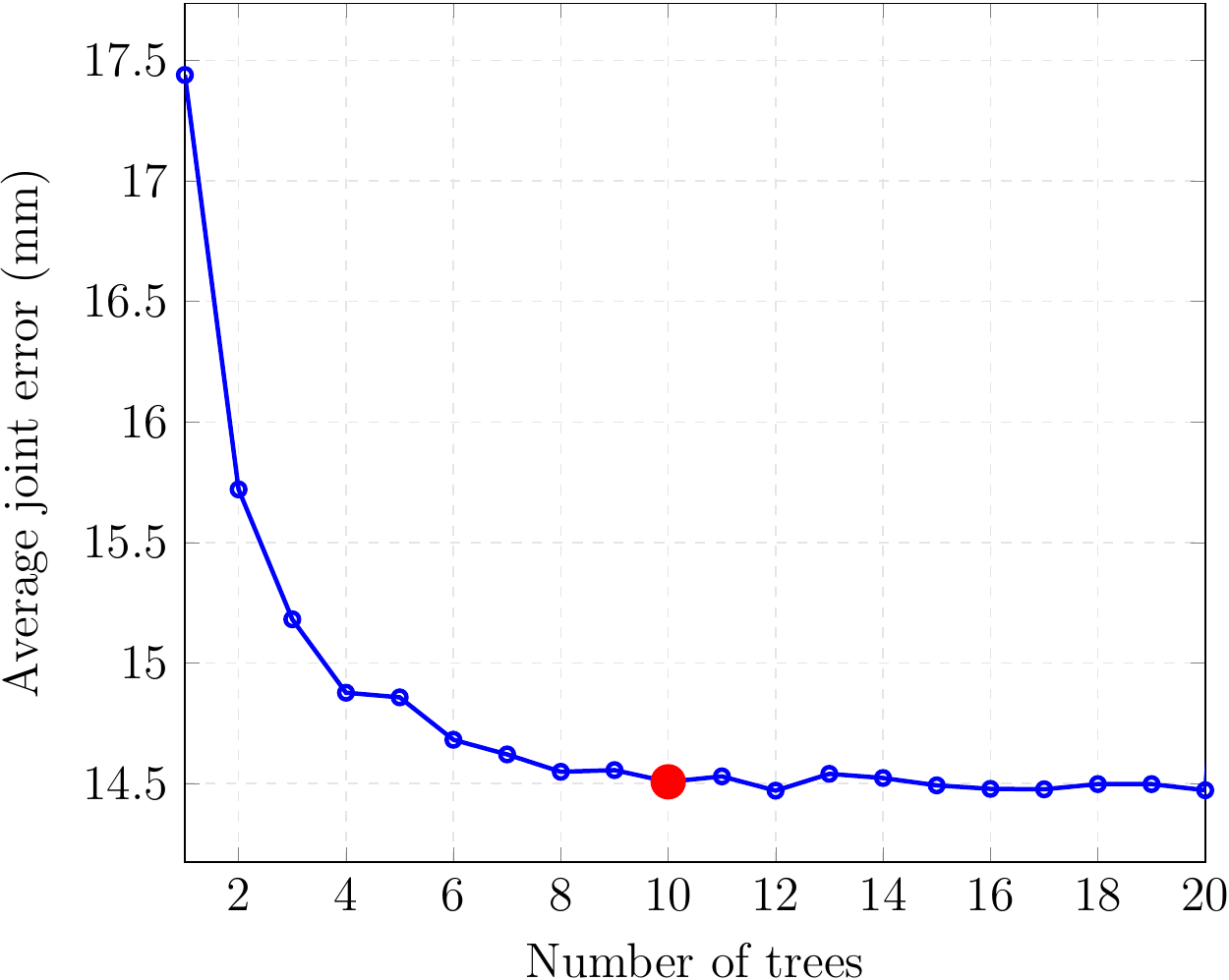}}
    }
	\\
	Tree depth&
    \subfloat {
		\raisebox{-.5\height}{\includegraphics[width=0.22\textwidth]{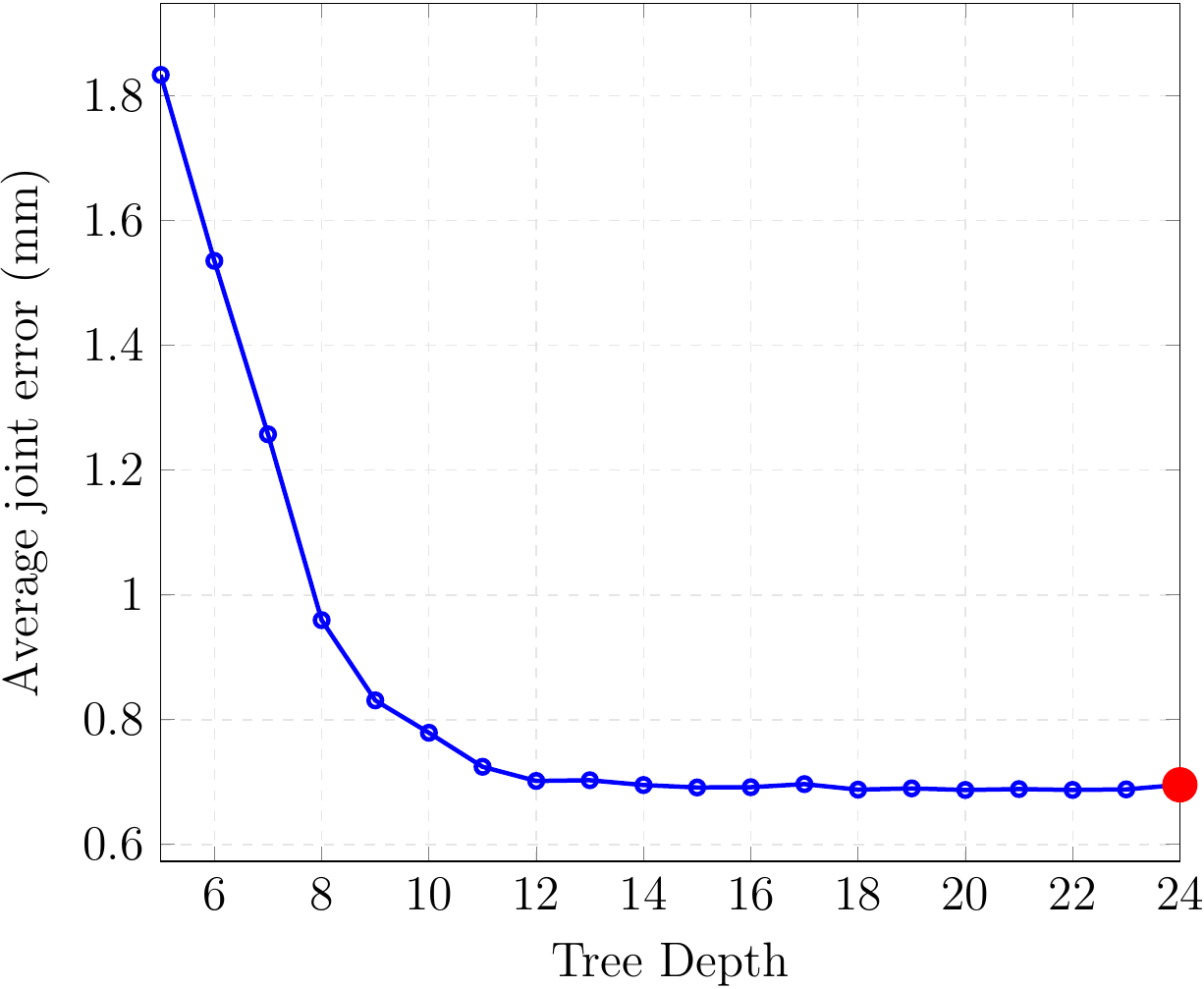}}
    }
	&
	\subfloat {
		\raisebox{-.5\height}{\includegraphics[width=0.22\textwidth]{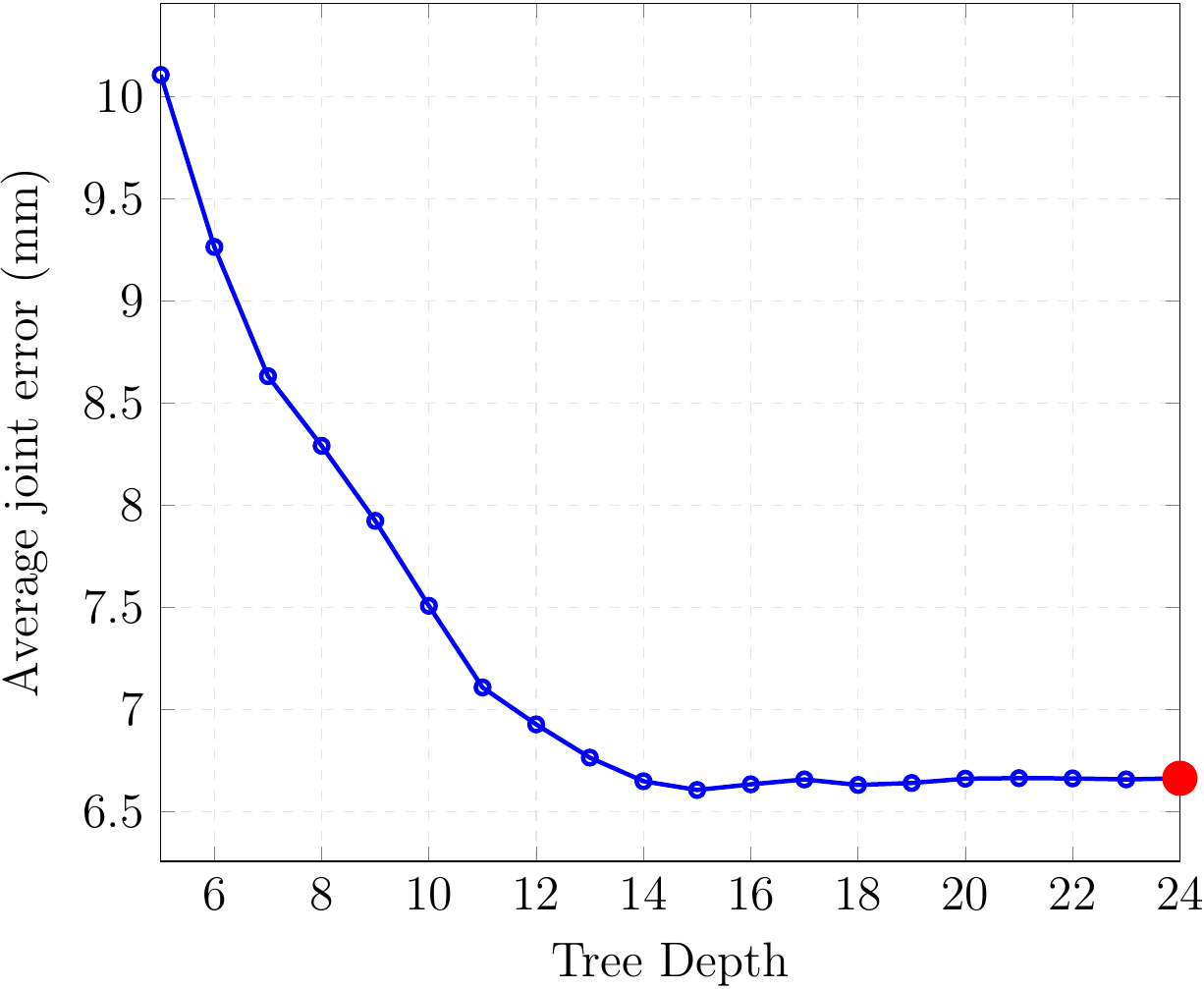}}
    }
	&
	\subfloat {
		\raisebox{-.5\height}{\includegraphics[width=0.22\textwidth]{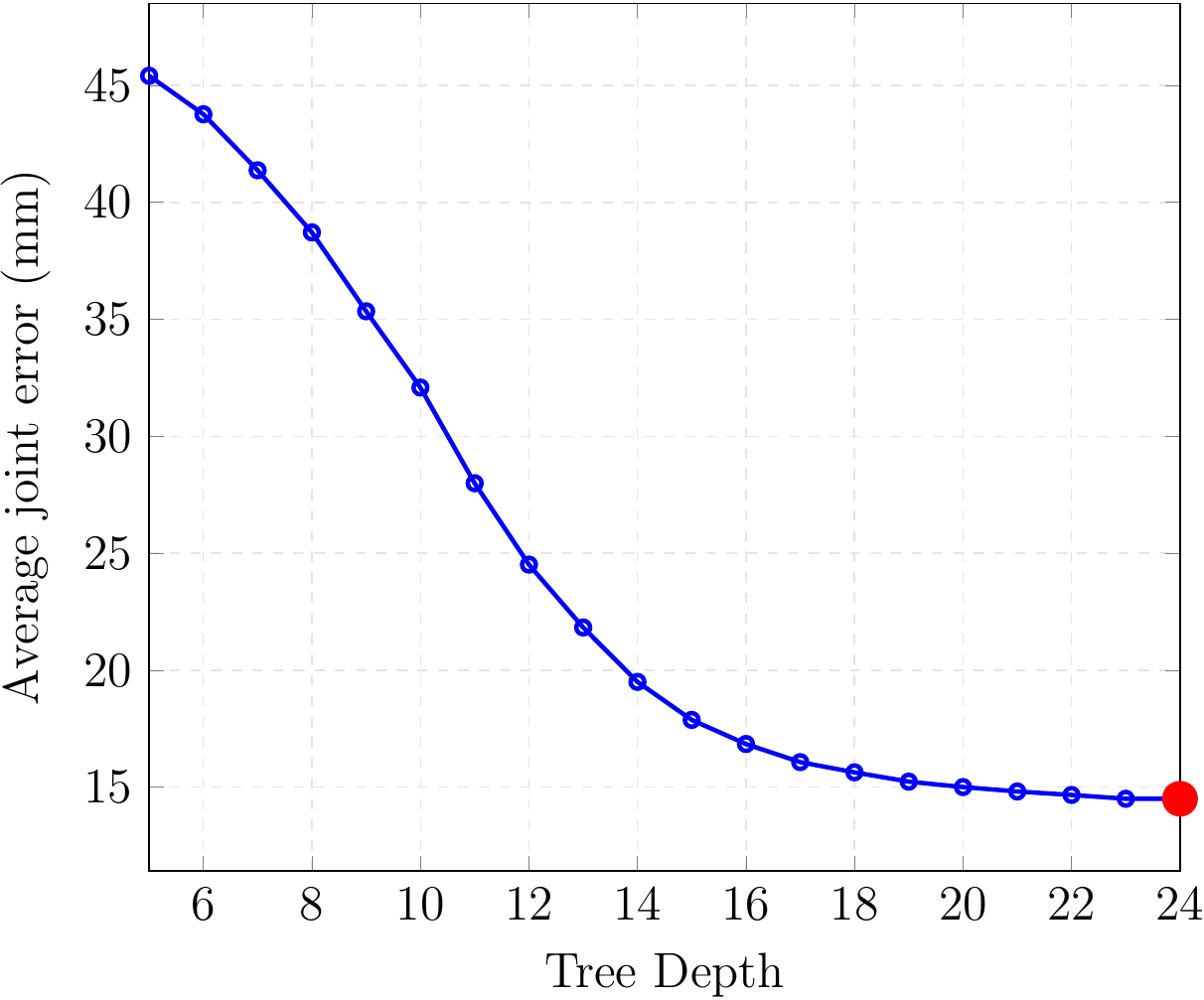}}
    }
	\\
	Number of initial poses $K_t$&
    \subfloat {
		\raisebox{-.5\height}{\includegraphics[width=0.22\textwidth]{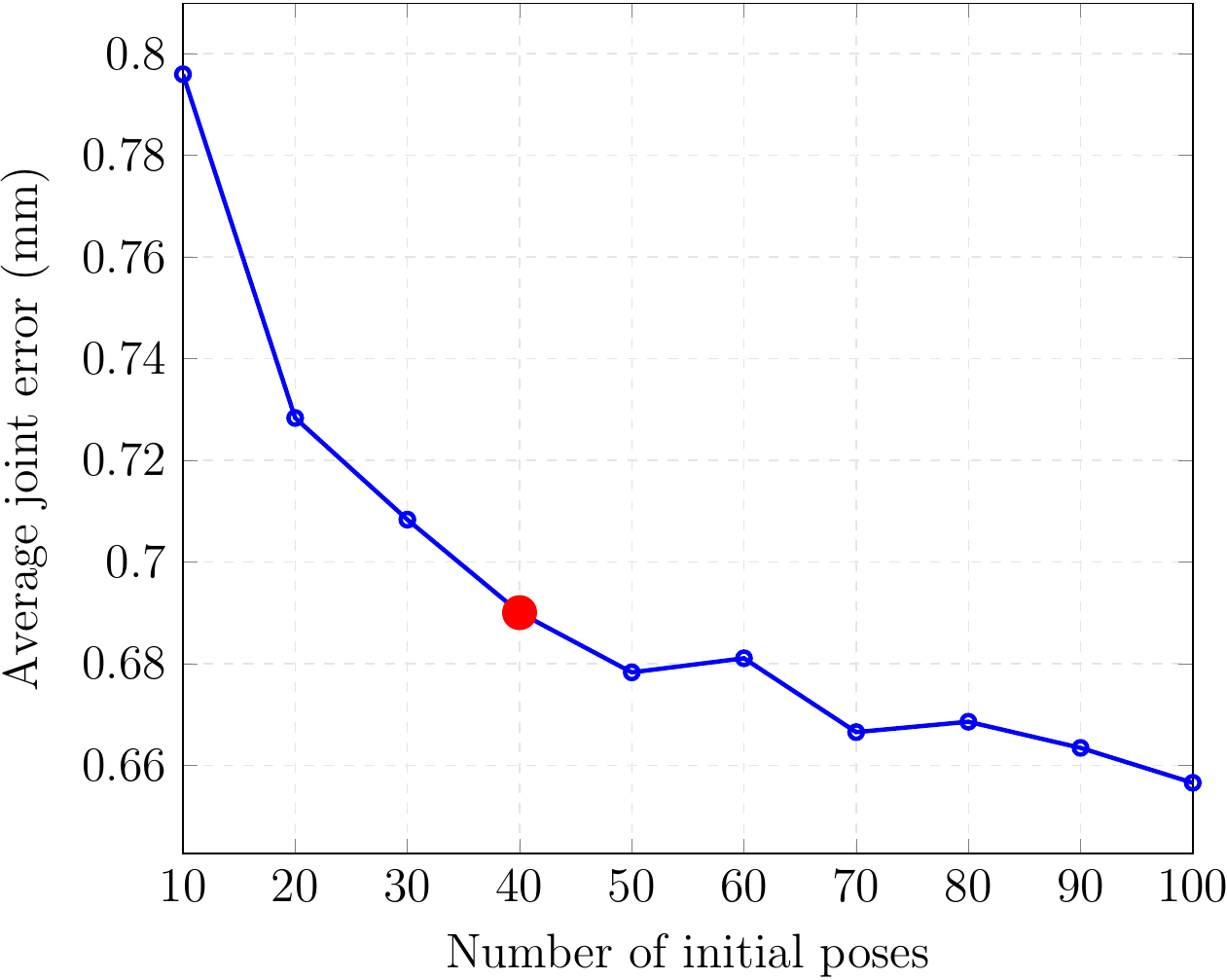}}
    }
	&
	\subfloat {
		\raisebox{-.5\height}{\includegraphics[width=0.22\textwidth]{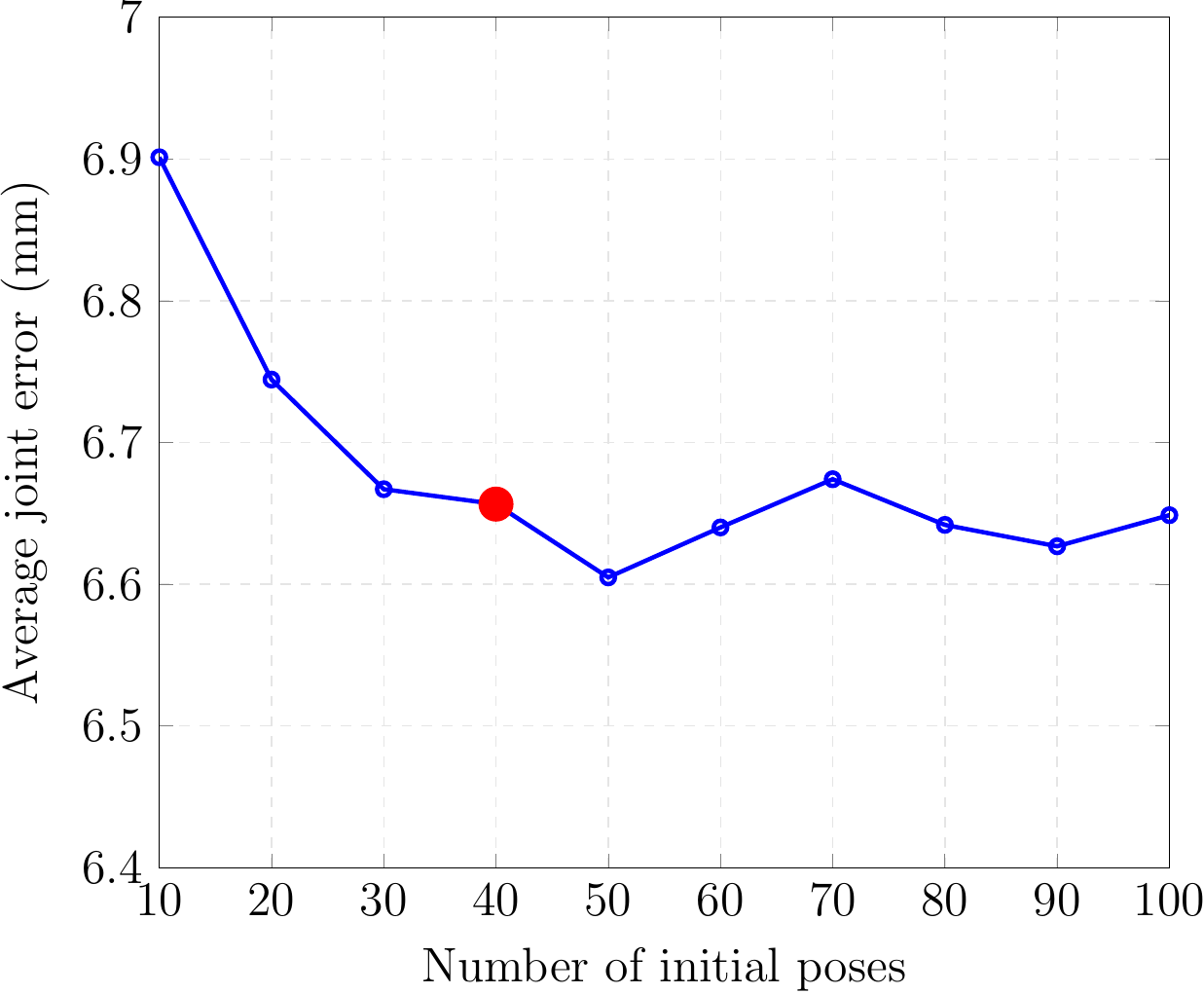}}
    }
	&
	\subfloat {
		\raisebox{-.5\height}{\includegraphics[width=0.22\textwidth]{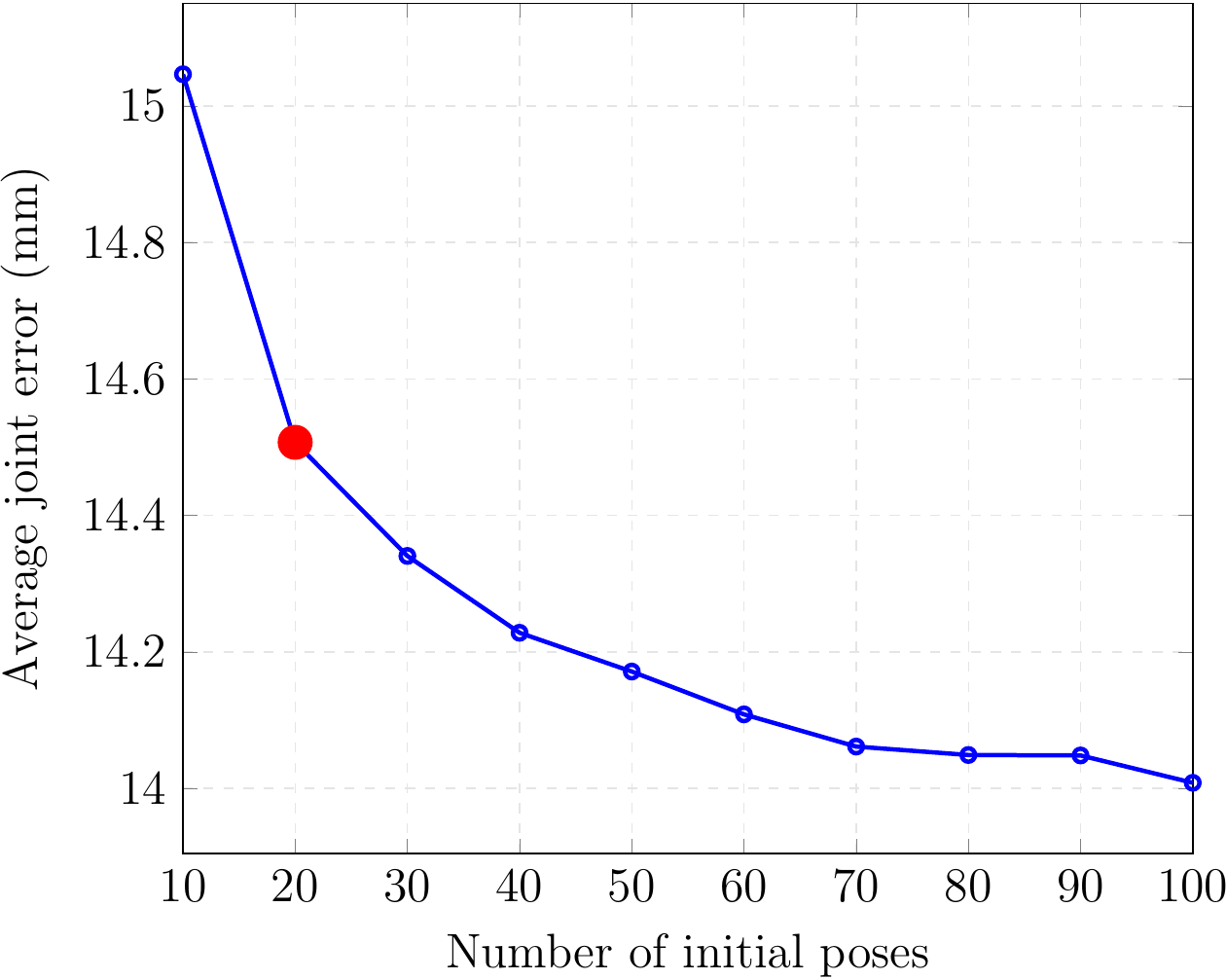}}
    }
	\\
	Number of iterations $C$&
    \subfloat {
		\raisebox{-.5\height}{\includegraphics[width=0.22\textwidth]{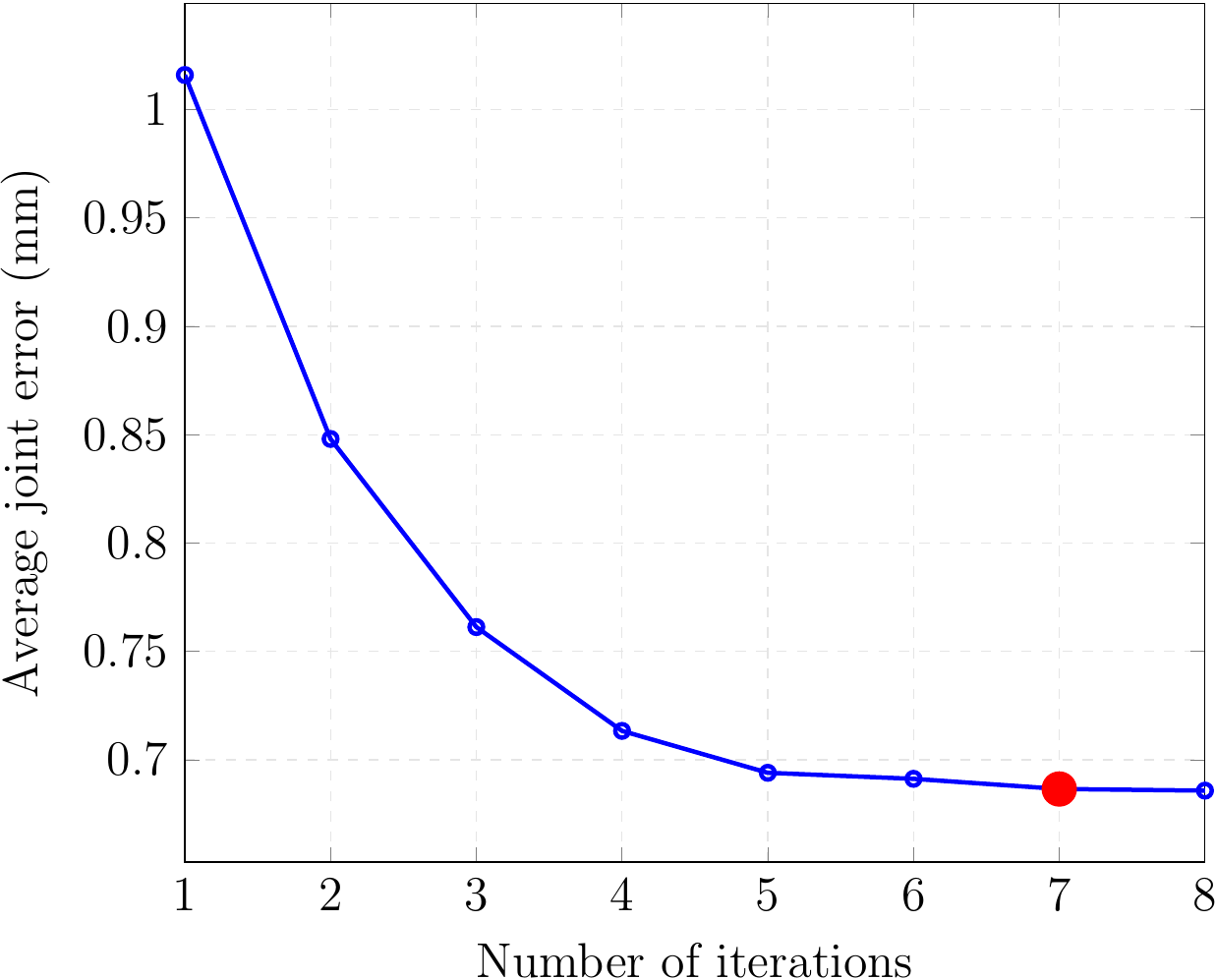}}
    }
	&
	\subfloat {
		\raisebox{-.5\height}{\includegraphics[width=0.22\textwidth]{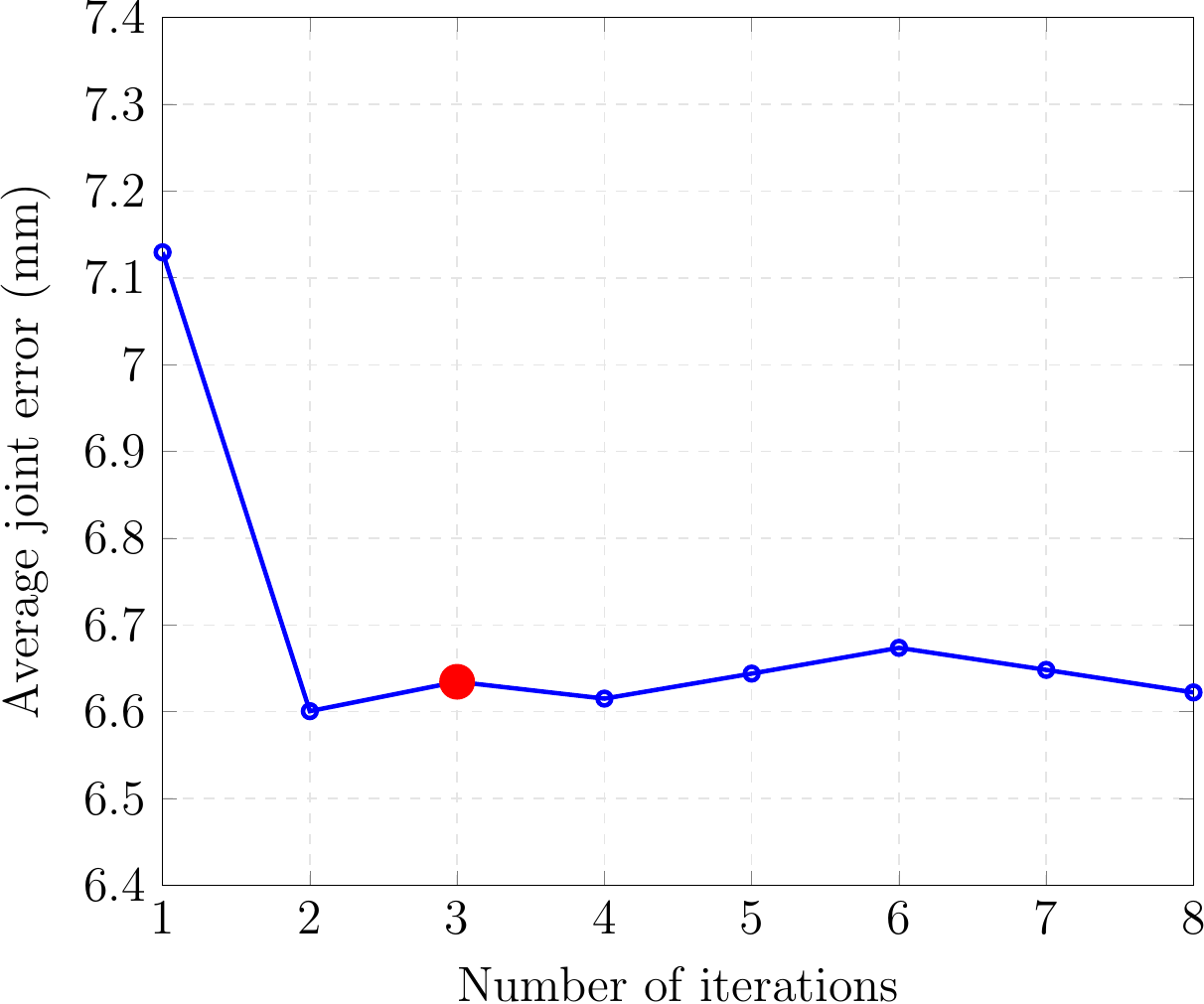}}
    }
	&
	\subfloat {
		\raisebox{-.5\height}{\includegraphics[width=0.22\textwidth]{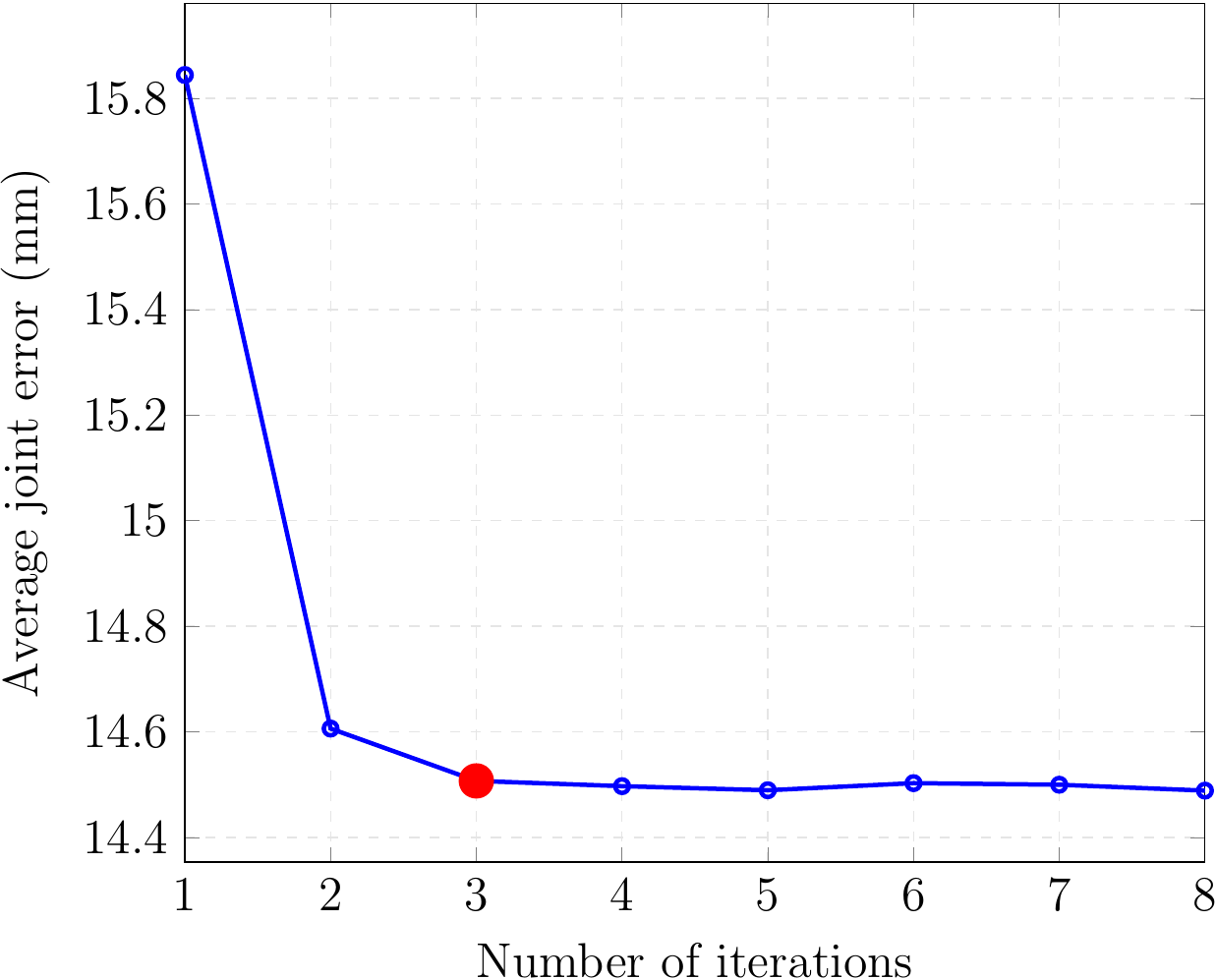}}
    }
	\\
	Number of trees for learning internal metric&
    \subfloat {
		\raisebox{-.5\height}{\includegraphics[width=0.22\textwidth]{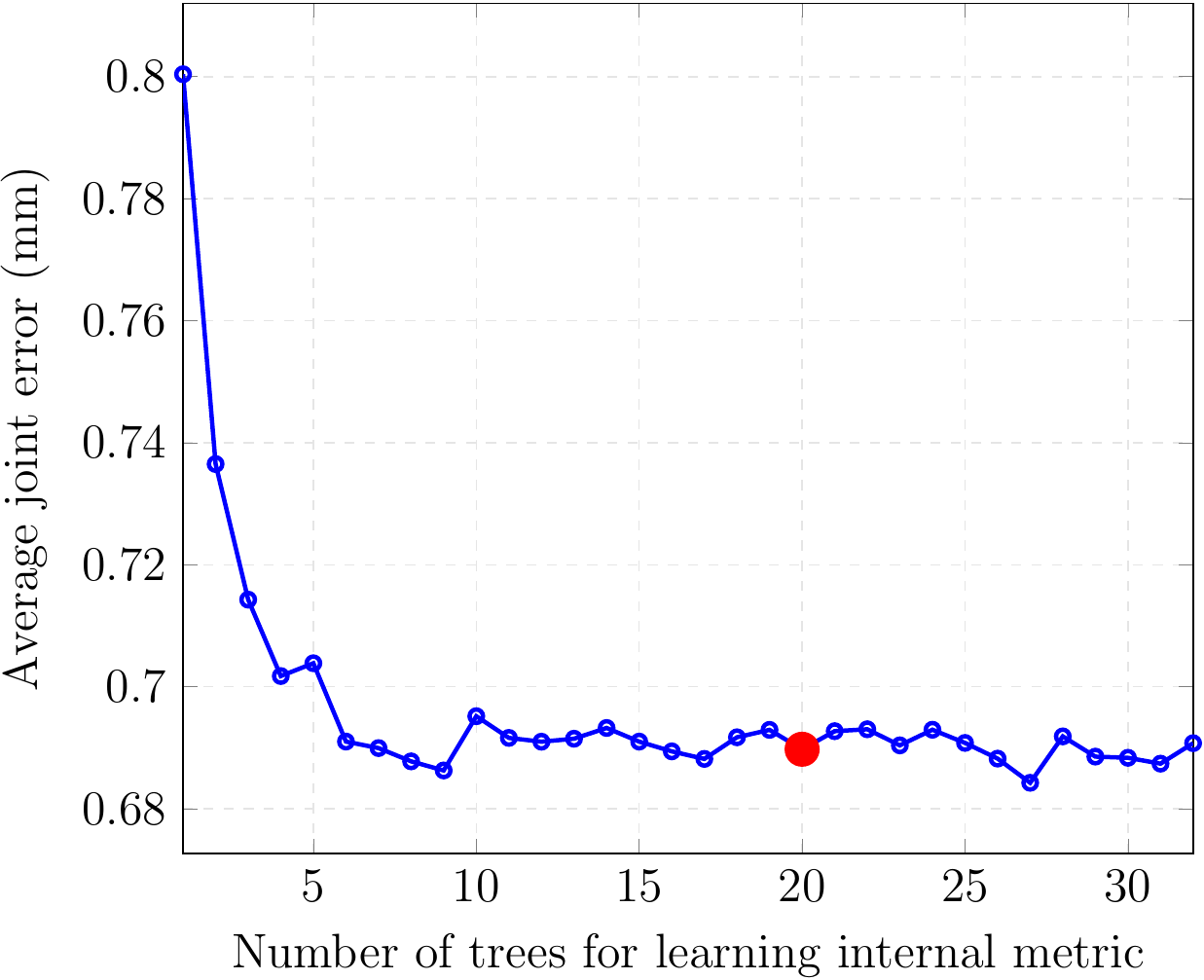}}
    }
	&
	\subfloat {
		\raisebox{-.5\height}{\includegraphics[width=0.22\textwidth]{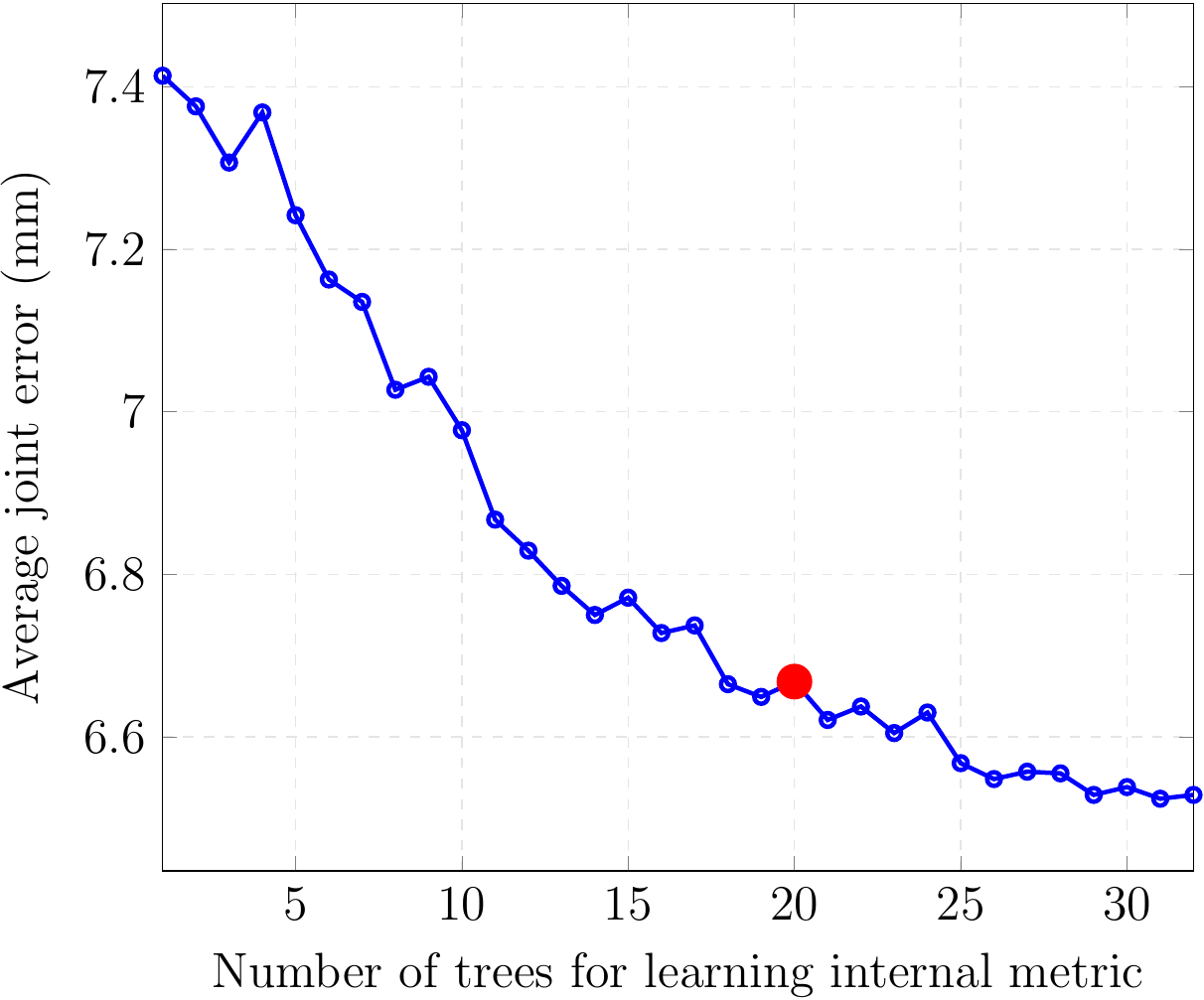}}
    }
	&
	\subfloat {
		\raisebox{-.5\height}{\includegraphics[width=0.22\textwidth]{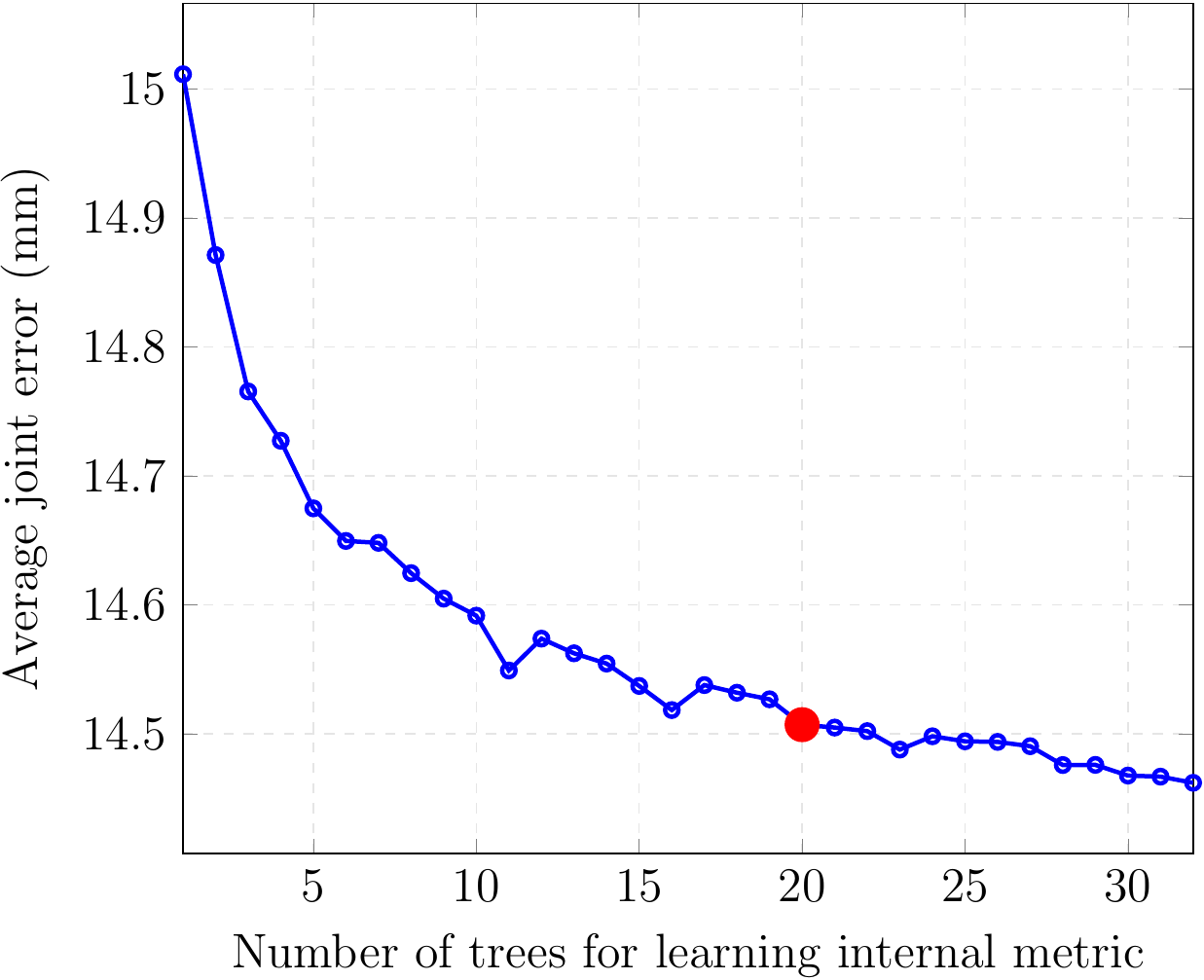}}
    }
    	\end{tabular}
    \caption{Sensitivity analysis of our \emph{Lie-X} approach w.r.t. internal parameters for pose estimation tasks: In each of the five rows, average joint error is plotted as a function of the respective internal parameter. It is further displayed in three columns for fish, mouse, and hand, respectively. In each of the panels, a \textcolor{red}{red} dot is placed to indicate the specific parameter value empirically employed in our approach.}
    \label{fig:internalParams}
\end{figure*}

\begin{figure*}[!t]
    \centering
    \subfloat[] {
	    \includegraphics[width=0.31\textwidth]{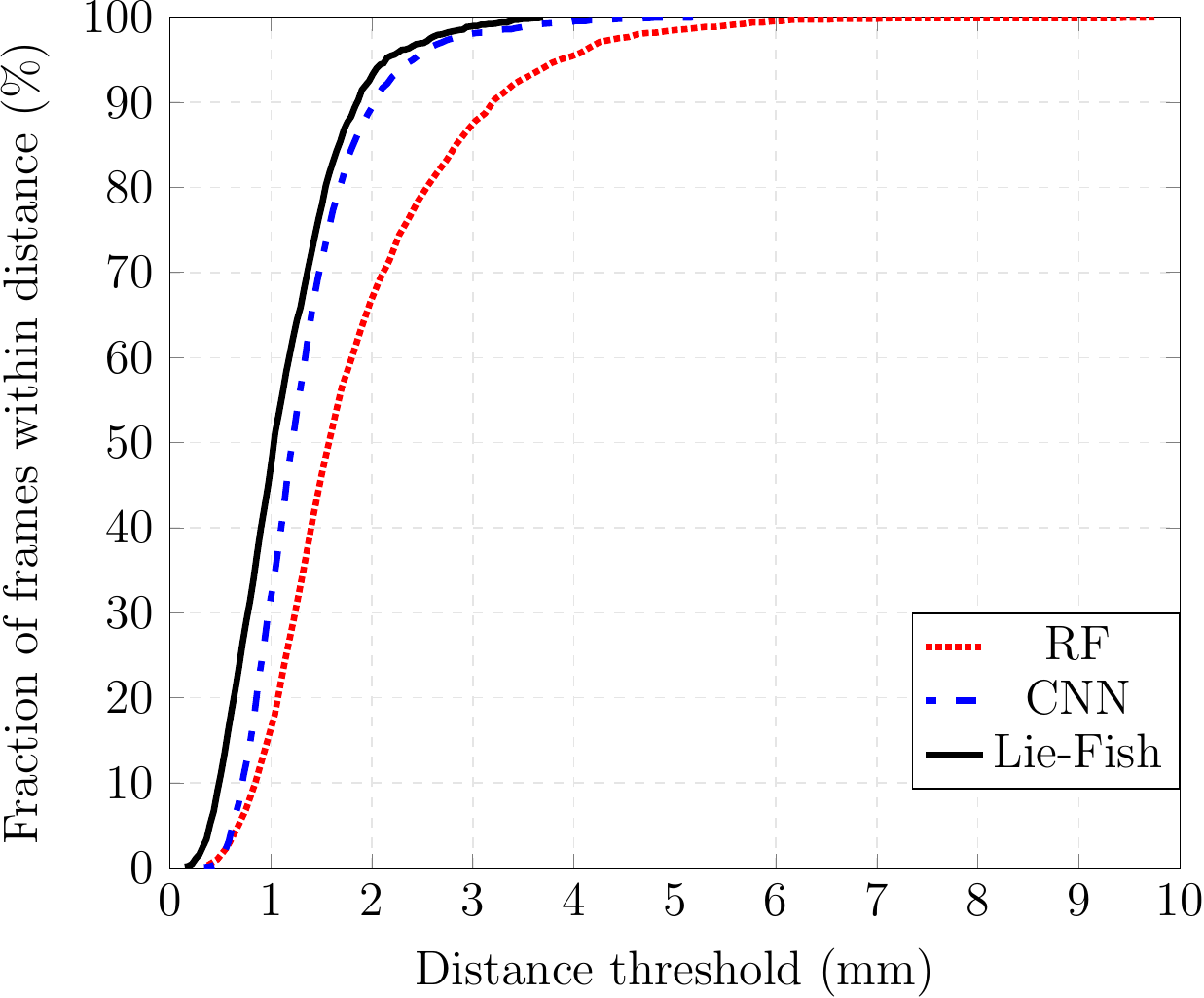}
    }
    \subfloat[] {
	    \includegraphics[width=0.31\textwidth]{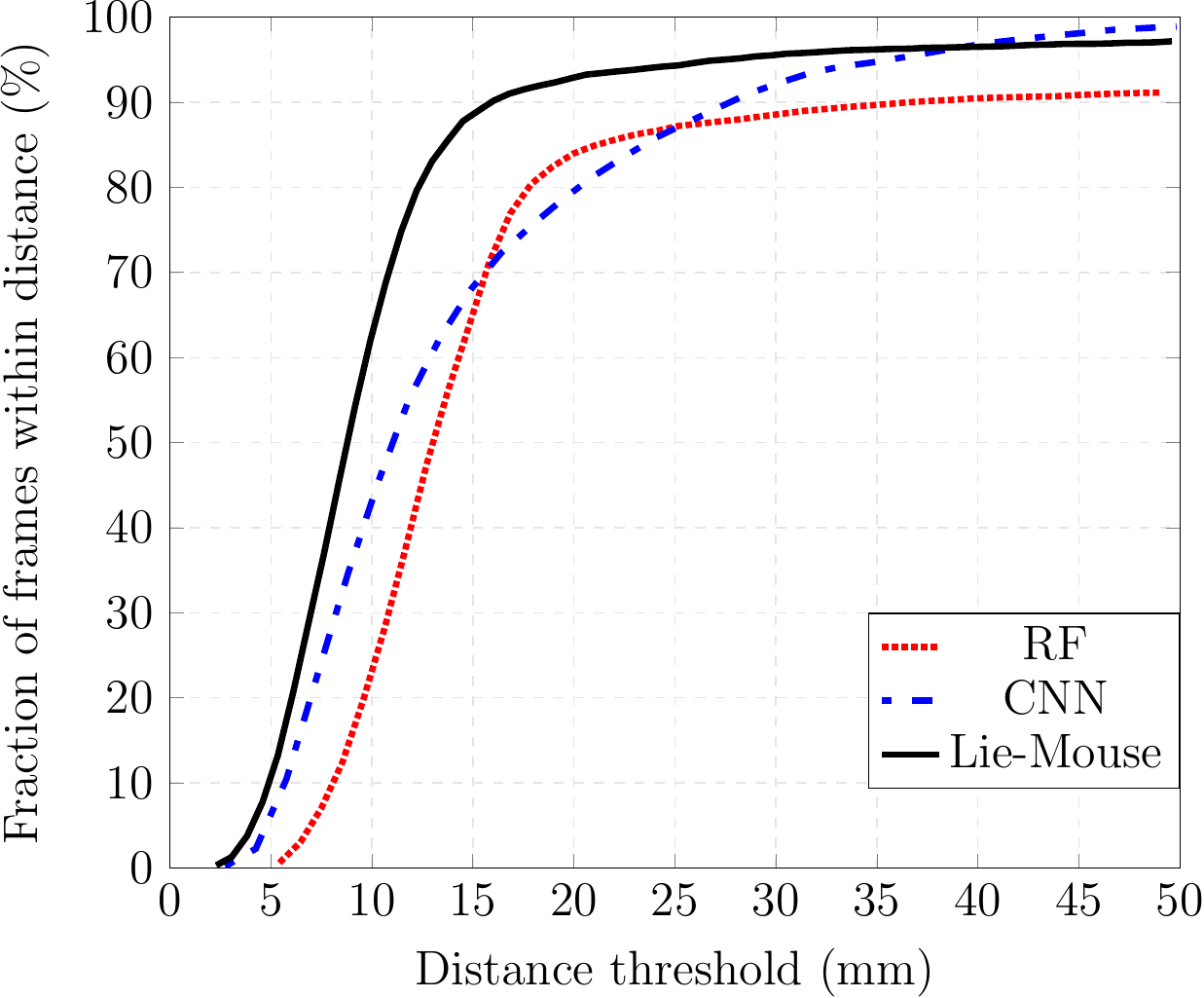}
    }
    \subfloat[] {
	    \includegraphics[width=0.31\textwidth]{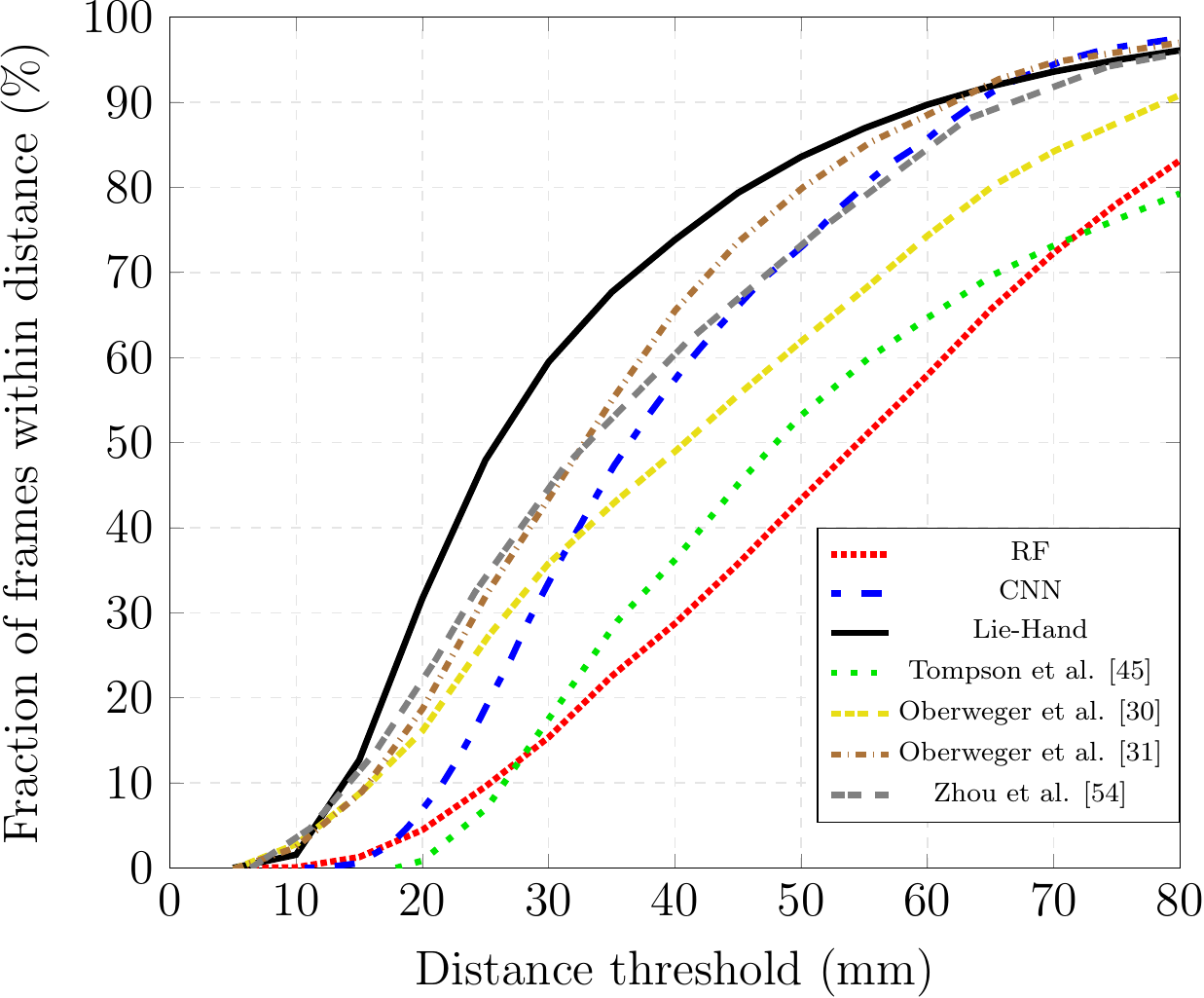}
    }
    \caption{Cumulative error distribution curves for pose estimation of (a) fish, (b) mouse and (c) hand, respectively. Horizontal axis displays the distance amount in mm of the estimated poses from ground-truths. Vertical axis presents the fraction of examples where their corresponding estimated poses possess average joint errors within the current distance range.}
    \label{fig:errDistribution}
\end{figure*}

\begin{figure*}[!t]
\centering
\includegraphics[width=0.8\textwidth]{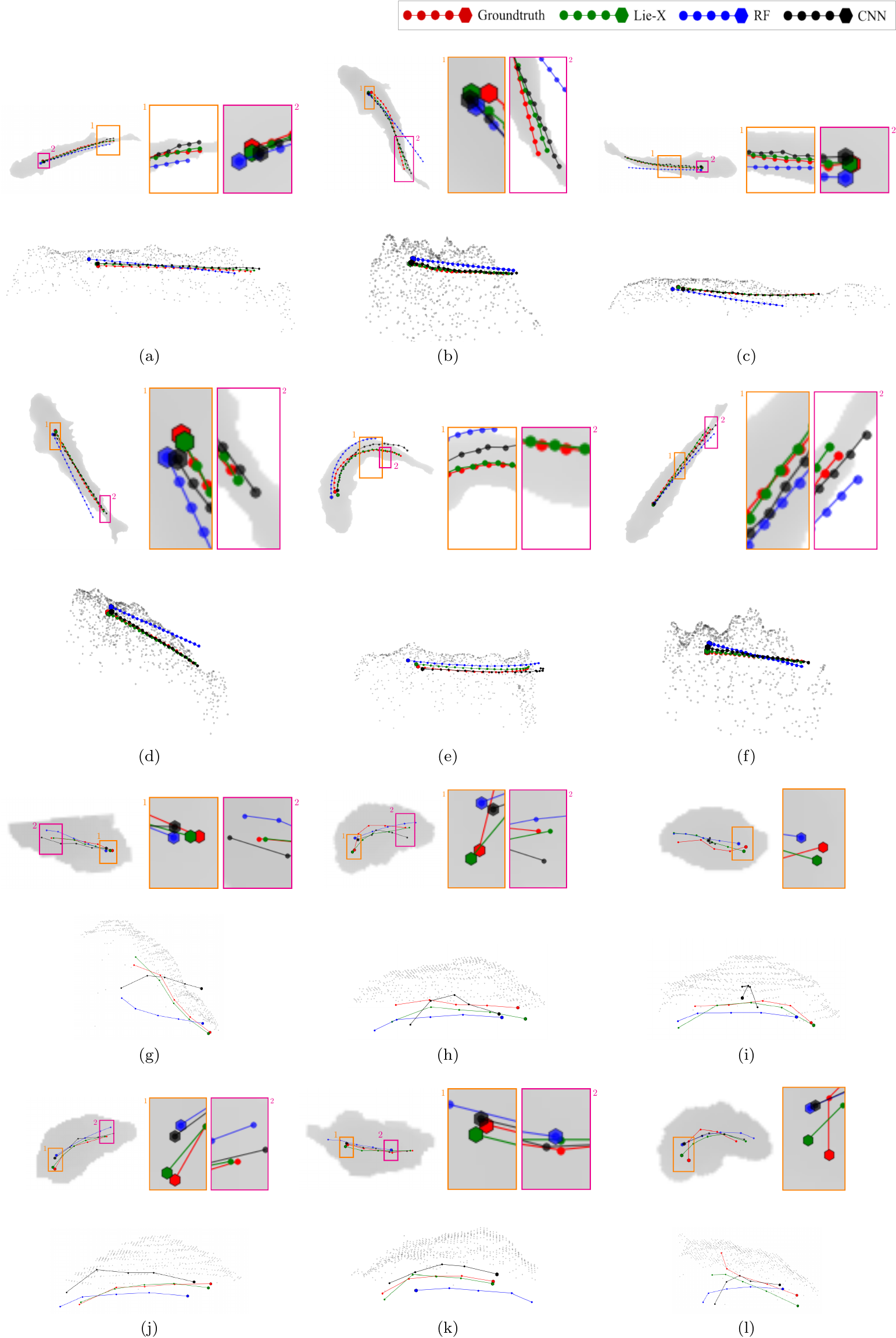}
	\caption{Visual comparison of fish and mouse examples. Here pose estimates of RF, CNN, as well as our \emph{Lie-X} approach are compared together with respective human-annotated ground-truths. Panels (a)--(f) present six fish examples, which is followed by panels (g)--(l) for six exemplar mouse results. In each of the twelve panels, top row displays the full top-view together with one or two zoom-in visual examinations. Meanwhile, the bottom row also provides a side-view. Best viewed in color.}
	\label{fig:poseEstimationVisualResults}
\end{figure*}

\begin{figure*}[!t]
\centering
\includegraphics[width=0.8\textwidth]{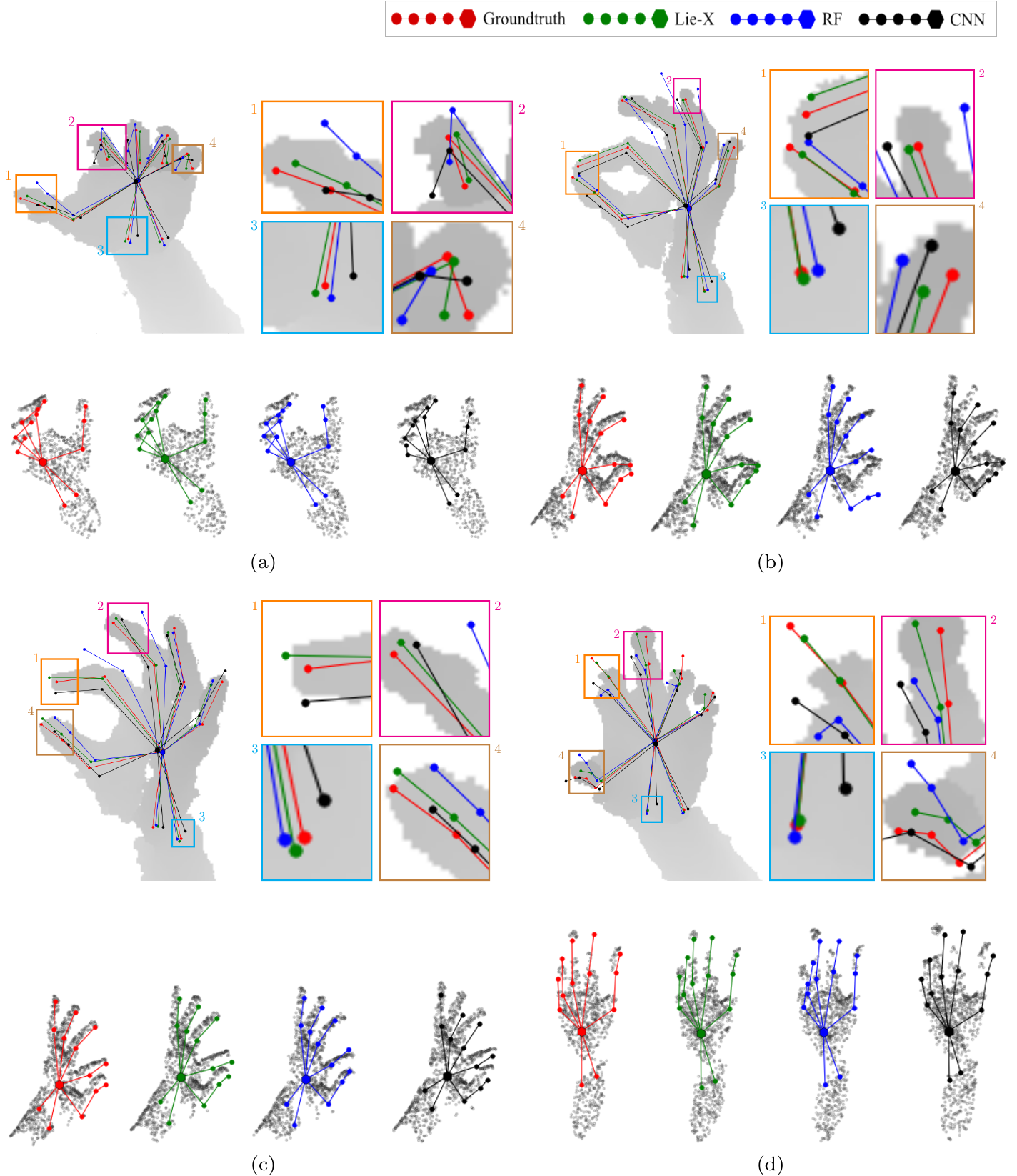}
	\caption{Visual comparison of hand examples. Here pose estimates of RF, CNN, as well as our \emph{Lie-X} approach are compared together with respective human-annotated ground-truths. In each of the four panels, top row displays the full top-view together with four zoom-in visual examinations. Meanwhile, the bottom row displays side-views of the respective methods. Best viewed in color.}
	\label{fig:poseEstimationVisualResultsHand}
\end{figure*}

\begin{figure*}[!t]
\centering
\includegraphics[width=0.9\textwidth]{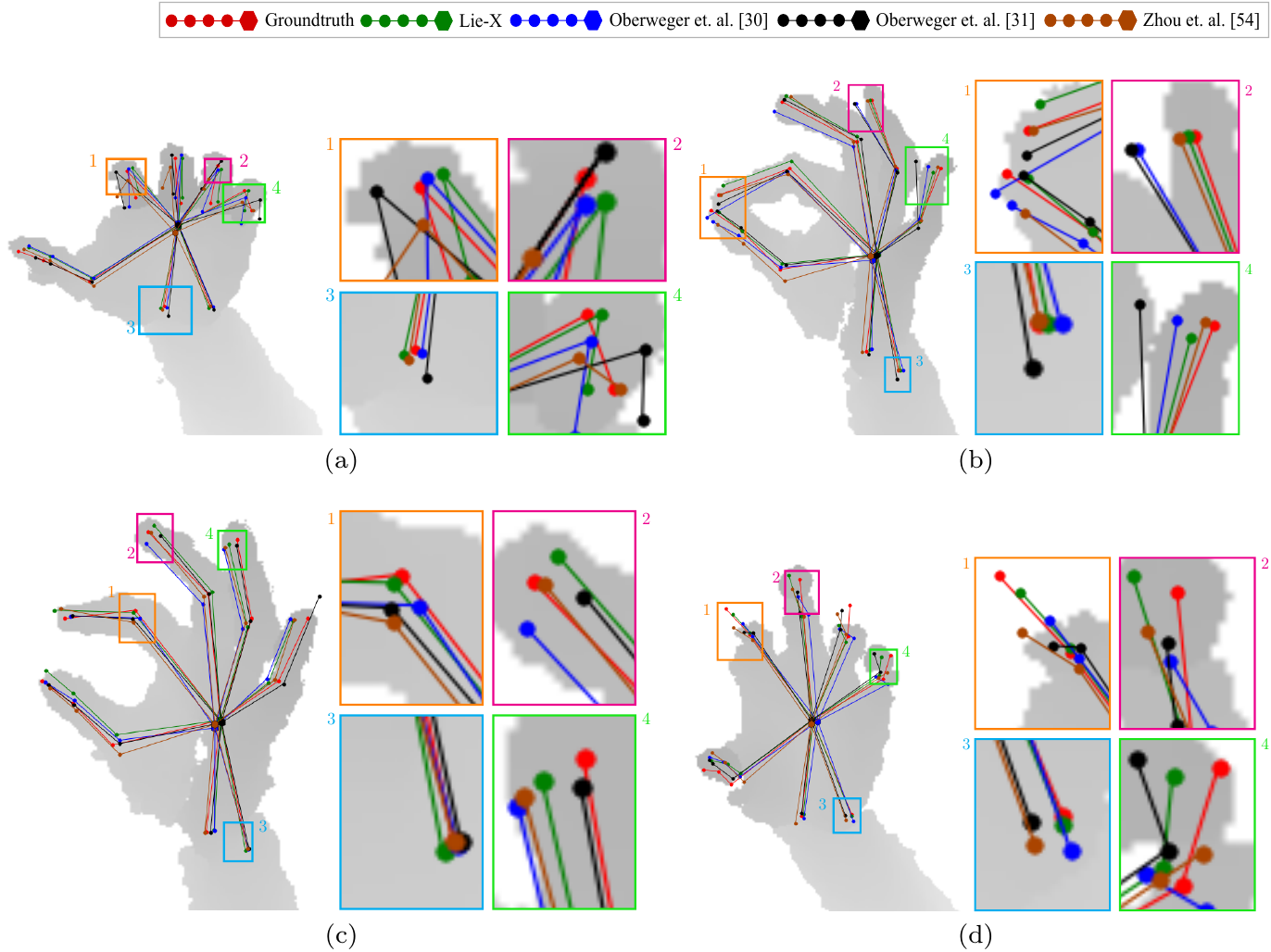}
	\caption{Visual comparison of \emph{Lie-X} as well as the state-of-the-art methods on the same four input hand images presented in Fig.~\ref{fig:poseEstimationVisualResultsHand}. In each of the panels, the corresponding example is presented with four zoom-in visual examinations. Best viewed in color.}
	\label{fig:poseEstimationVisualResultsHand2}
\end{figure*}

\begin{figure*}[!t]
\centering
\includegraphics[width=0.9\textwidth]{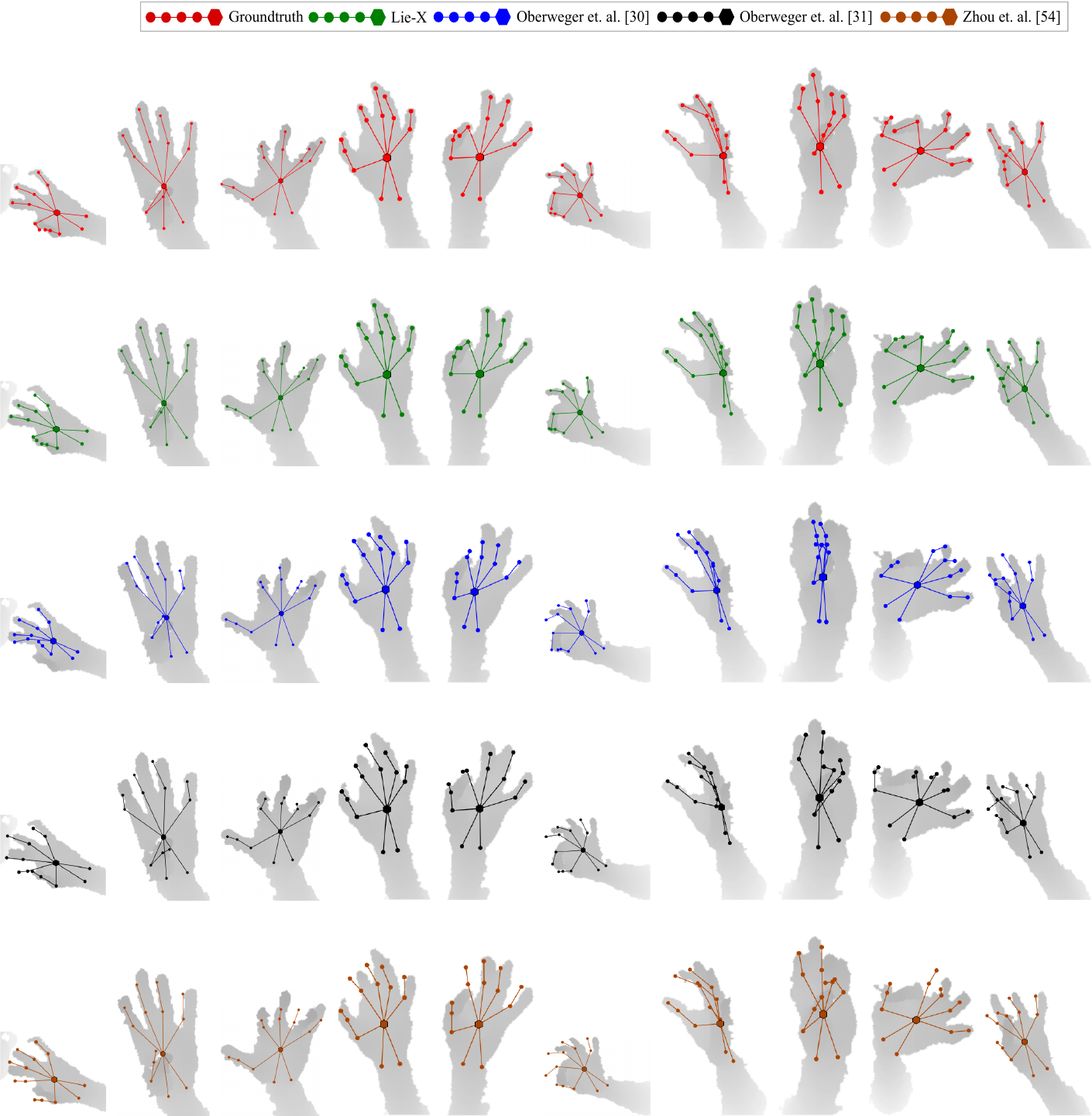}
	\caption{Visual comparison of \emph{Lie-X} results and the state-of-the-art methods on ten additional hand examples. Each column presents an example, while each row displays results from a particular competing method. Best viewed in color.}
	\label{fig:poseEstimationVisualResultsHand3}
\end{figure*}

\subsection{Datasets}
To examine the applicability of our approach on diverse articulated objects, we demonstrate in this paper its empirical implementation for fish, mouse, and human hand, respectively, where three distinct real-life datasets are employed.
In particular, here we introduce our home-grown 3D image datasets of zebrafish and lab mouse that are dedicated to the related problems of pose estimation, tracking, and action recognition.
The images have been captured and annotated by experts to provide the articulated skeleton information describing the pose of the subject. 
The popular NYU hand depth image dataset~\cite{TomEtAl:siggraph14} is also considered here. More details of the datasets are discussed next.
It is worth noting that different imaging modalities are utilized across the three datasets: light-field depth images are used for fish, while structured illumination depth cameras are employed for mouse and human hand objects. Regardlessly our approach is demonstrated to work well across these diverse image modalities.

%
%

\paragraph{Our Fish Dataset}
Depth images are acquired with a top-mount Raytrix R5 light-field camera at a frame rate of 50 FPS and a resolution of $1,024 \times 1,024$,
as displayed in Fig.~\ref{fig:capture_setup}(a). The depth images are obtained from the raw plenoptic images by utilizing Raytrix on-board SDK.
In total 7 different adult zebrafish of different genders and sizes are engaged in our study.
From the captured images, 2,972 images containing distinct poses are annotated.
The training dataset of $n_t=$95,104 images is thus formed by augmenting each fish object of these images with 31 additional transformations,
where each transformation comes with a random scaling within [0.9, 1.1] and with a random in-plane rotation within (-$\pi$, $\pi$).
The test dataset of pose estimation problem contains $n_{\mathrm{tst}}=$1,820 distinct fish images.

\begin{figure*}[!t]
    \centering
	\begin{tabular} {@{}c@{}c@{}cc}
    \subfloat {
		\includegraphics[width=0.2\textwidth]{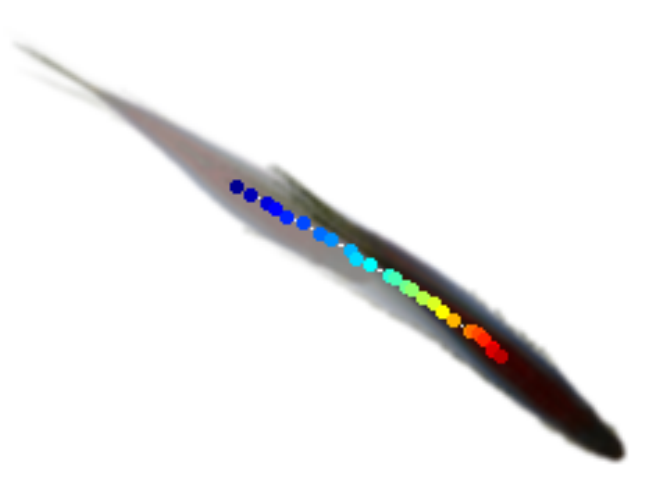}
    }
&    \subfloat {
		\includegraphics[width=0.2\textwidth]{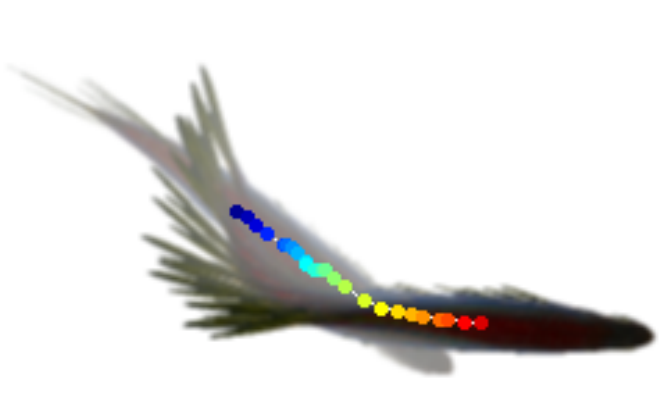}
    }
&    \subfloat{
		\includegraphics[width=0.2\textwidth]{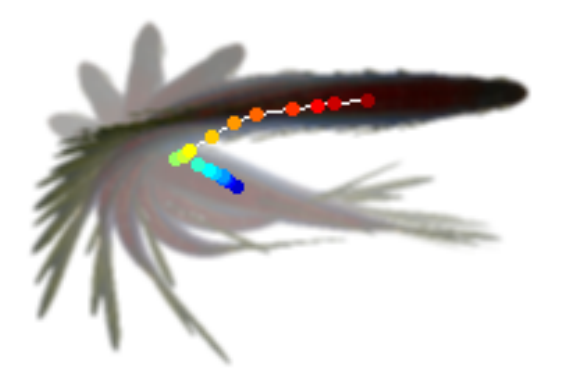}
    }
&	 \subfloat {
		\includegraphics[width=0.04\textwidth]{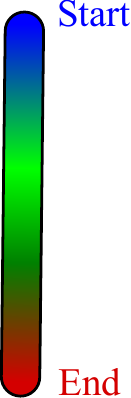}
    }
    \\
    (a) Scoot & (b) J-turn & (c) C-turn & \\
    \\
    \subfloat {
		\includegraphics[width=0.2\textwidth]{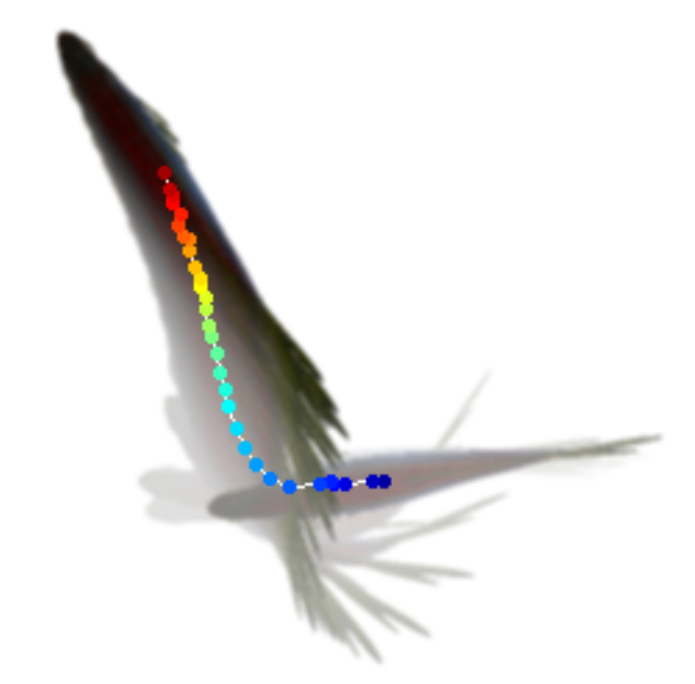}
    }
&    \subfloat {
		\includegraphics[width=0.17\textwidth]{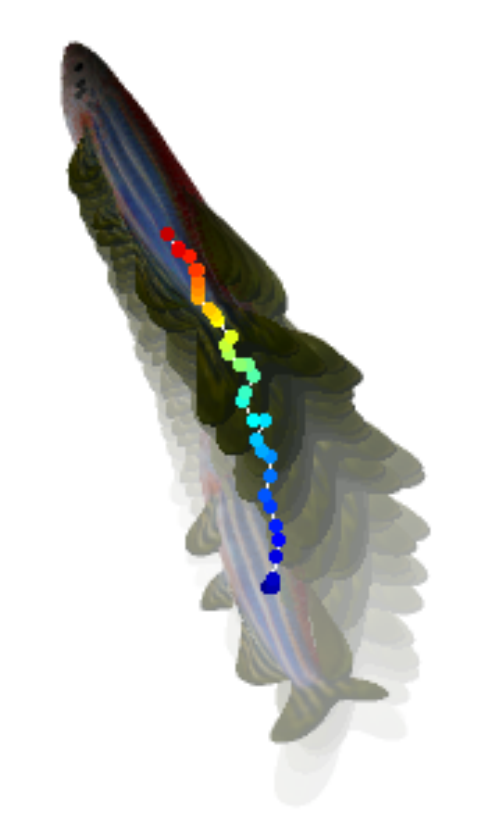}
    }
&    \subfloat {
		\includegraphics[width=0.2\textwidth]{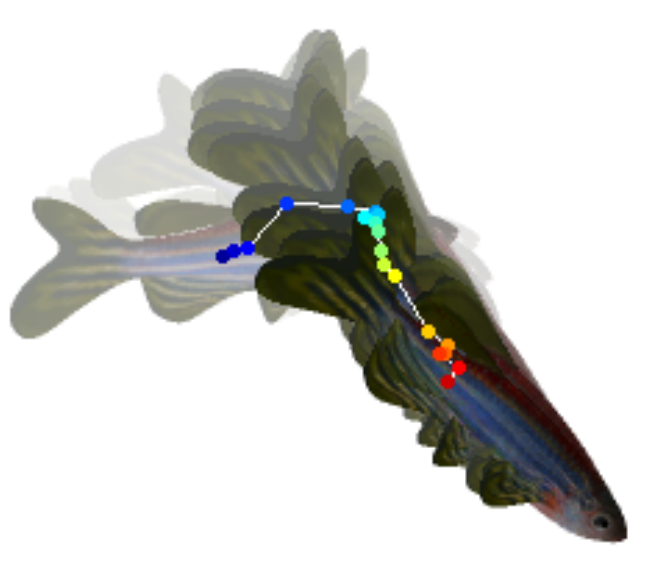}
    } &
    \\
	(d) R-turn & (e) Surface & (f) Dive &
	\\
        \subfloat {
		\includegraphics[width=0.2\textwidth]{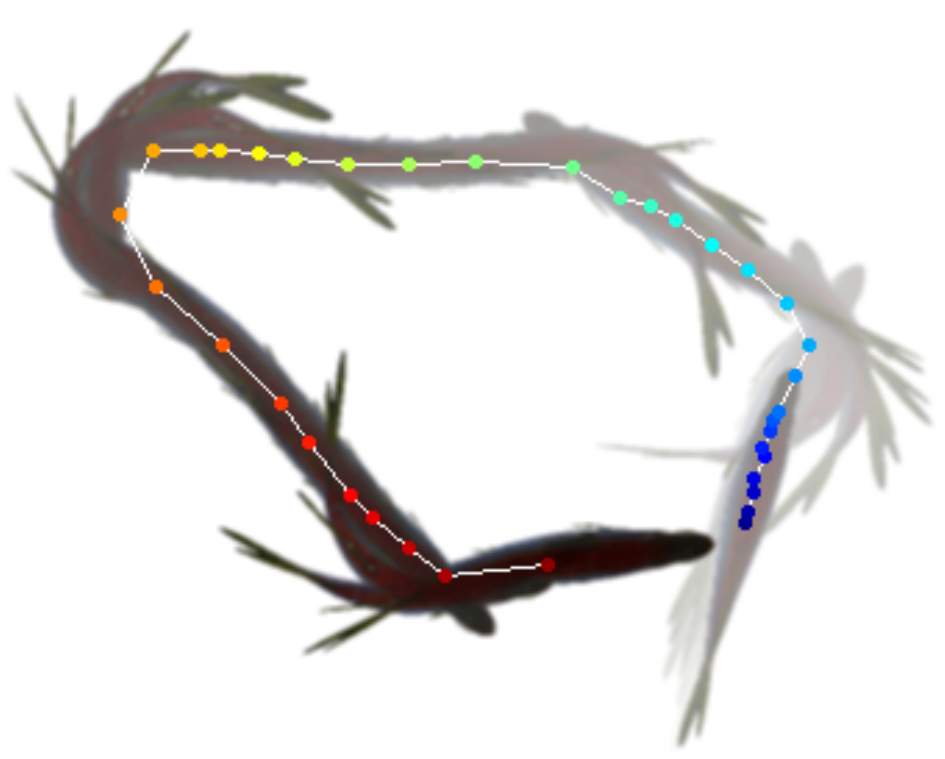}
    }
&    \subfloat{
		\includegraphics[width=0.22\textwidth]{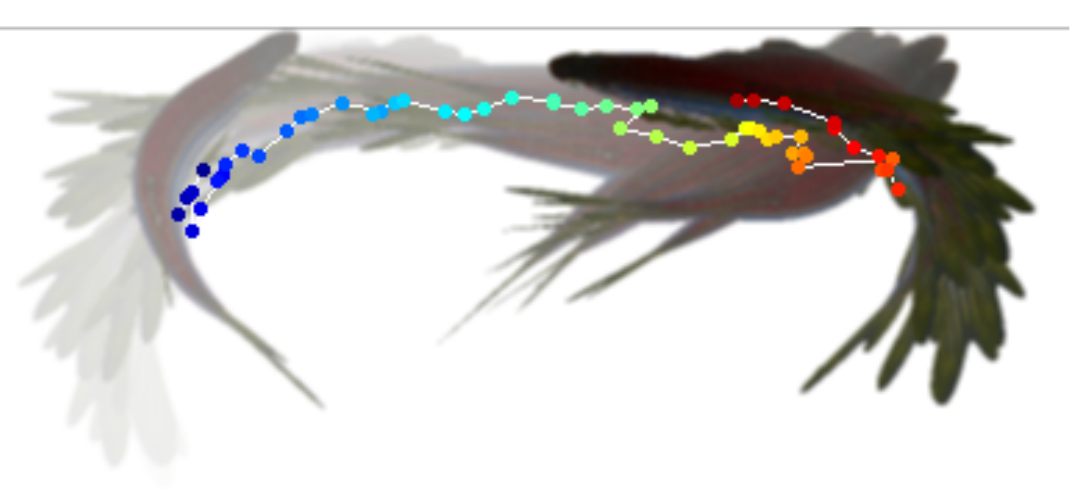}
    }
&    \subfloat{
		\includegraphics[width=0.2\textwidth]{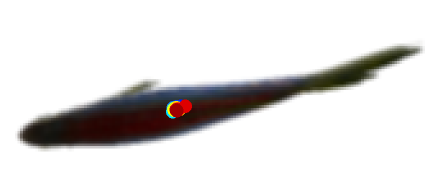}
    } &
    \\
	(g) Zigzag & (h) Thrash & (i) Freeze &
	\\
    	\end{tabular}
    \caption{Key frames from the nine distinct fish action categories considered in our experiments. The colored dots display the trajectory of the fish motions, where \textcolor{blue}{blue} and \textcolor{red}{red} mark the start and end of the action respectively. Note for a better illustration of the distinct fish action categories,
    (e) and (f) present a side view of the \emph{surface} and \emph{dive} actions, while a top view is adopted for the rest action types.}
    \label{fig:fishactions}
\end{figure*}

In addition to single-frame based pose annotations, we also record, annotate, and make available a \textit{fish action} dataset.
\cite{zfActDb:zf13} provides a comprehensive catalogue of zebrafish actions,
from which a subset of 9 action classes are considered in this paper, which is listed below as well as illustrated in Fig.\ref{fig:fishactions}:
\begin{description}
  \item[\textit{Scoot}:] Moves along a straight line.
  \item[\textit{J-turn}:] Fine reorientation during which the body slightly curves $(30^{\circ}-60^{\circ})$, with a characteristic bend at the tail.
  \item[\textit{C-turn}:] Fish body curves to form a C-shape en route to a near $180^{\circ}$ turn.
  \item[\textit{R-turn}:] Involves routine angular turn of greater than $60^{\circ}$.
  \item[\textit{Surface}:] Moves up towards the water surface.
  \item[\textit{Dive}:] Moves towards the tank bottom.
  \item[\textit{Zigzag}:] Contains erratic movements with multiple darts in various directions.
  \item[\textit{Thrash}:] Consists of forceful swimming against the side or bottom walls of the tank.
  \item[\textit{Freeze}:] Refers to complete cessation of movement.
\end{description}

Our fish action dataset contains 426 training sequences and 173 testing sequences, respectively, from 7 different fish over the aforementioned 9 categories.
The length of each fish action sequence varies from 7 frames to 135 frames.

\paragraph{Our Mouse Dataset}
Mouse depth images are collected using a top-mount Primesense Carmine depth camera at a frame rate of 30 FPS and with a resolution $640 \times 480$. Fig.~\ref{fig:capture_setup}(b) presents our dedicated imaging set-up.
Two different lab mouse are engaged in our study.
We select 3,253 images containing distinct poses and depth noise patterns, and augment them with additional transformations following the same protocol as above, which gives rise to
the training dataset here containing $n_t=$104,096 images.
The testing dataset of pose estimation problem contains $n_{\mathrm{tst}}=$4,125 distinct depth images.
For tracking problem, the test set consists of two sequences of length 511 and 300 frames, respectively.

\paragraph{The NYU Hand Dataset}
We also evaluate our approach on the benchark NYU hand depth image dataset\cite{TomEtAl:siggraph14}~\footnote{The NYU dataset is publicly available at \url{http://cims.nyu.edu/~tompson/NYU_Hand_Pose_Dataset.htm}.}. It contains $n_t=$72,757 depth images for training and $n_{\mathrm{tst}}$=8,252 frames for testing. All images are depth images captured by Microsoft Kinect using the structured illumination technique, which is the same as the Primesense camera used in our mouse dataset. Depth images in the training set are from a single user, while images in the test set are from two users. While a ground-truth hand label contains 36 annotated joints, only 14 of these joints are considered in many existing efforts using this dataset, such as~\cite{TomEtAl:siggraph14,ObeWohLep:cvww15,ObeWohLep:iccv15,zhoumodel:ijcai16}, which is followed during our experiments. This is also presented in Fig.~\ref{fig:nyu_hand_model}: Important hand joints are included in this subset of 14 joints, such as all the finger tips and the hand base.

\begin{table}
	\begin{center}
\scalebox{0.8}{
  		\begin{tabular}{ | l || c | c | }
    		\hline
	    Comparison methods on articulated objects &  fish & mouse \\ \hline \hline
    		RF & 1.28 & 12.24 \\ \hline
	    CNN & 0.79 & 9.17 \\ \hline
		Lie-X (w/o multiple initial poses) & 3.28 & 13.27 \\ \hline
	    Lie-X (w/o learned metric) & 1.71 & 9.82 \\ \hline
		Lie-X &  \textbf{0.68} & \textbf{6.64} \\ \hline
	\end{tabular}
}
\end{center}
	\caption{Quantitative evaluation of competing methods on pose estimation problem for fish and mouse respectively.
Performance is measured in terms of average joint error (mm).}
    \label{tab:poseEstimationResult}
\end{table}

\begin{table*}
	\begin{center}
  		\begin{tabular}{ | l || c | }
    		\hline
	    Comparison methods on hand pose estimation &  average joint error (mm) \\ \hline \hline
    		RF & 24.81  \\ \hline
	    CNN & 18.82 \\ \hline
		Tompson et. al. ~\cite{TomEtAl:siggraph14} & 21.00 \\ \hline
		Oberweger et. al. ~\cite{ObeWohLep:cvww15} & 20.00 \\ \hline
	    Oberweger et. al. ~\cite{ObeWohLep:iccv15} & 16.50 \\ \hline
		Zhou et. al. ~\cite{zhoumodel:ijcai16} & 16.90 \\ \hline
		Lie-X (w/o multiple initial poses) & 20.50 \\ \hline
		Lie-X (w/o learned metric) & 16.72 \\ \hline
		Lie-X &  \textbf{14.51} \\ \hline
	\end{tabular}
\end{center}
	\caption{Quantitative evaluation of competing methods on the benchmark NYU dataset~\cite{TomEtAl:siggraph14} for hand pose estimation task. Performance is in terms of average joint error (mm).}
    \label{tab:poseEstimationResultHand}
\end{table*}

\subsection{Pose Estimation of Fish, Mouse, and Human Hand}

In this subsection, we focus on the problem of pose estimation for articulated objects such as fish, mouse, and hand.
To make a fair comparison with existing methods, we specifically implement two non-trivial baseline methods, namely regression forest (RF), and convolutional neural network (CNN).
The RF method is a re-implementation of the classical regression method used by Microsoft Kinect~\cite{ShoEtAl:pami13},
with the only difference being that our RF implementation explicitly utilizes a skeletal model, instead of estimating joint locations without skeletal constraints as in~\cite{ShoEtAl:pami13}. Two separate regression forests, $F_1$ and $F_2$, are trained for this purpose. Here $F_1$ is used to estimate the 3D location and in-plane orientation of the subject, followed by $F_2$ which produces a set of 3D pose candidates. The number of trees trained are set to 7 and 12 for $F_1$ and $F_2$, respectively. In both cases, the maximum tree depth is fixed to $L$=20. The standard depth invariant two-point offset features of~\cite{ShoEtAl:pami13} are also used.
The CNN method is obtained as follows: The pre-trained AlexNet CNN model from ImageNet is engaged as the initial model.
To tailor the training data for our CNN, objects of interest from the training depth images are cropped according to their bounding boxes.
The depth values in each patch are rescaled to be in the range of 0 to 255.
Each object patch is replicated three times to form into a RGB image, which is then resized as an input instance of size 224 $\times$ 224.
This together with its corresponding annotation prepares a training example.
Then our CNN model is finally obtained by executing the MatConvNet package to train on these training examples for 50 epochs.

\begin{figure*}[!t]
    \centering
	    \includegraphics[width=0.99\textwidth]{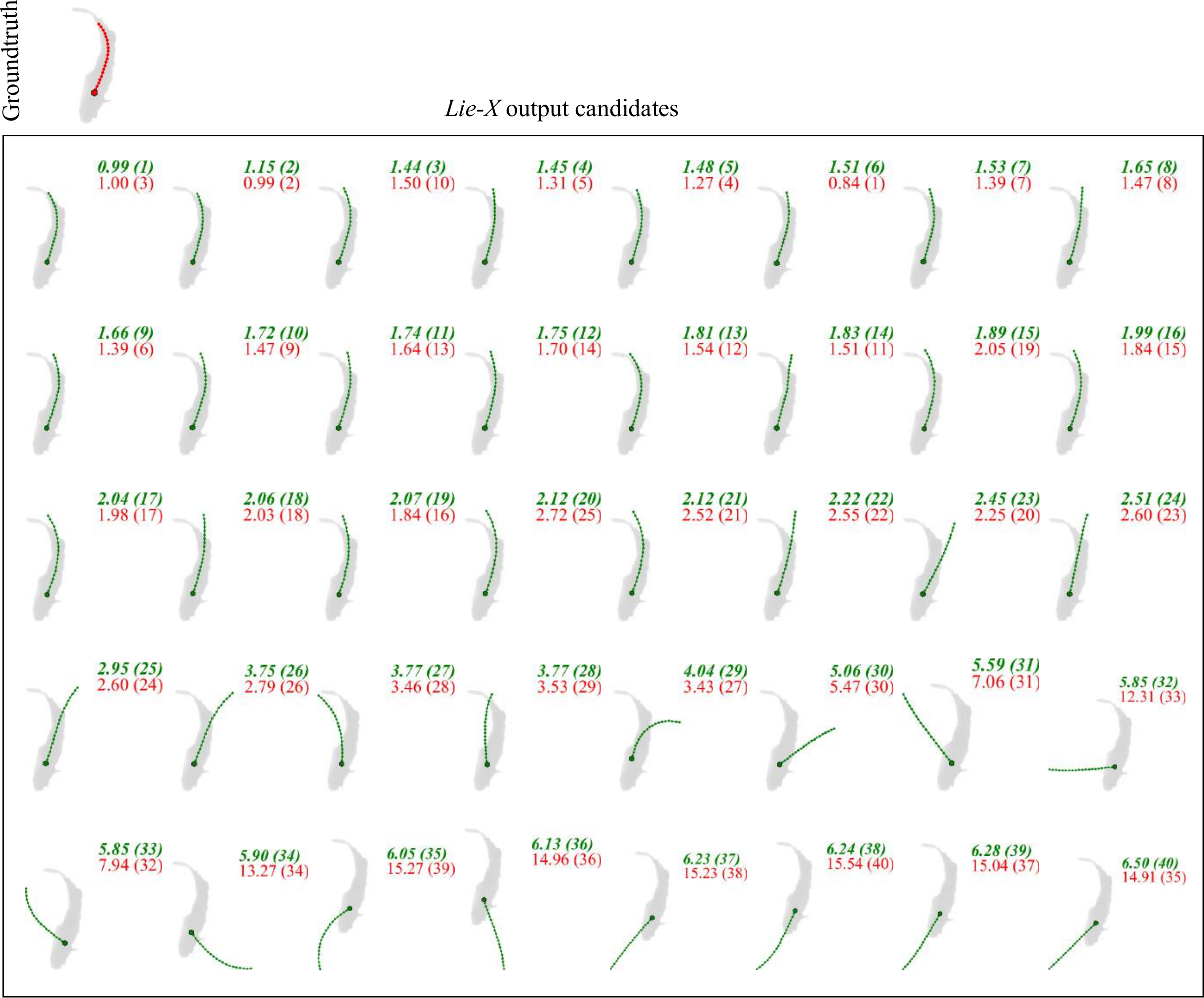}
    \caption{An example of fish pose estimation that visually illustrates the inner-working of the learned internal metric in our approach applied onto the $K_{\mathrm{t}}=40$ pose candidates of the input depth image. Here \textcolor{green}{green} colored numbers correspond to the scores (lower the better here) and ranking results obtained by applying the learned metric, \textcolor{red}{red} colored numbers denote the corresponding actual evaluation scores and ranking results by engaging the empirical evaluation metric of average joint error when we have access to the ground-truth annotations. See text for details.}
    \label{fig:fishCandidatesMetric}
\end{figure*}

\begin{figure*}[!t]
    \centering
	    \includegraphics[width=0.99\textwidth]{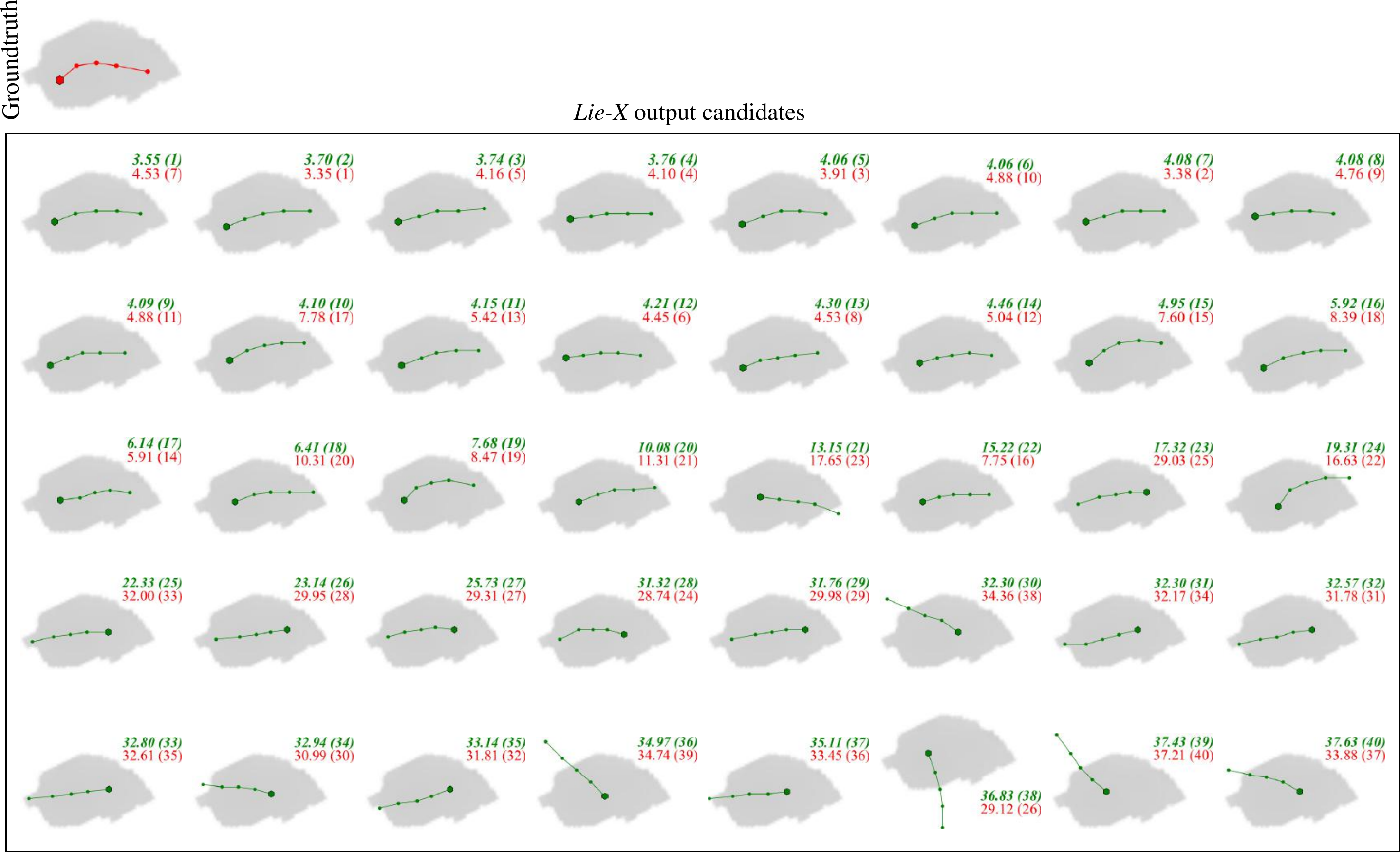}
    \caption{An example of mouse pose estimation that visually illustrates the inner-working of the learned internal metric in our approach applied onto the $K_{\mathrm{t}}=40$ pose candidates. Here \textcolor{green}{green} colored numbers refer to the scores (lower the better here) and ranking results obtained by applying the learned metric, \textcolor{red}{red} colored numbers are the corresponding actual evaluation scores and ranking results by engaging the empirical evaluation metric of average joint error when we have access to the ground-truth annotations. See text for details.}
    \label{fig:mouseCandidatesMetric}
\end{figure*}

\begin{figure*}
    \centering
	    \includegraphics[width=0.99\textwidth]{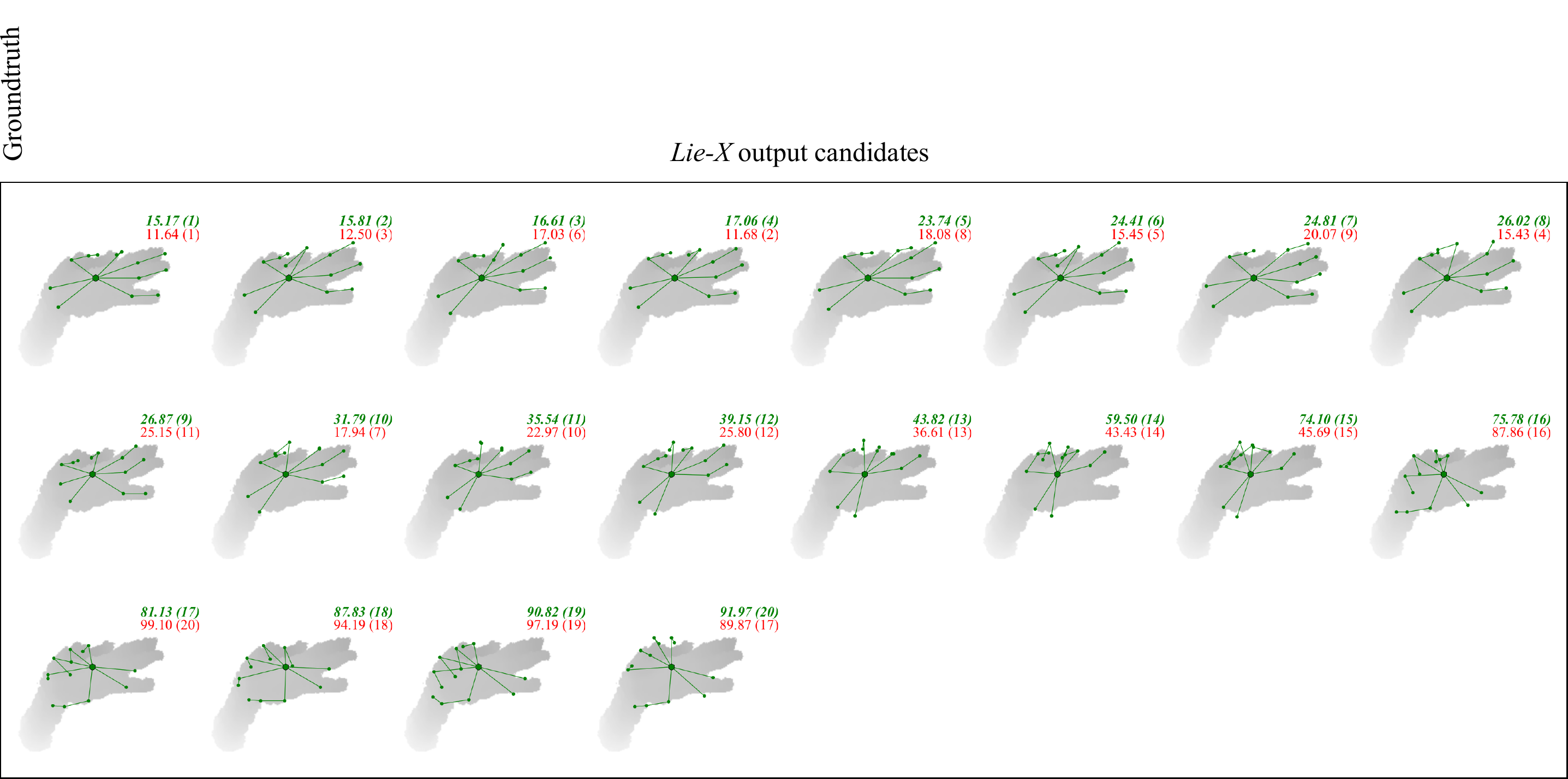}
    \caption{An example of hand pose estimation that visually illustrates the inner-working of the learned internal metric in our approach applied onto the $K_{\mathrm{t}}=20$ pose candidates. Here \textcolor{green}{green} colored numbers present the scores (lower the better here) and ranking results obtained by applying the learned metric, \textcolor{red}{red} colored numbers denote the corresponding actual evaluation scores and ranking results by engaging the empirical evaluation metric of average joint error when we have access to the ground-truth annotations. See text for details.}
    \label{fig:handCandidatesMetric}
\end{figure*}

\begin{figure}
    \centering
	    \includegraphics[width=0.45\textwidth]{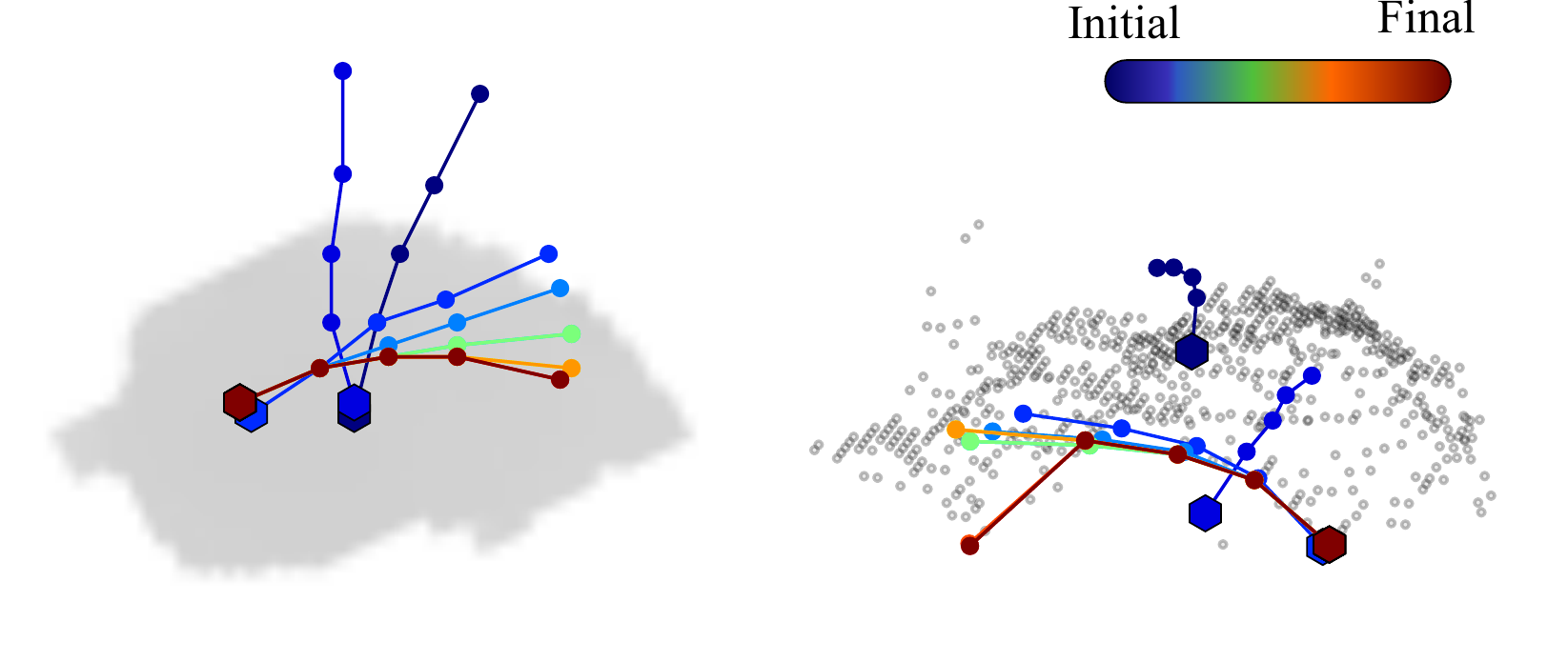}
    \caption{Illustrating the convergence process on the same mouse example presented in Fig.~\ref{fig:mouseCandidatesMetric}. It starts from an initial pose candidate to the final pose estimation result, which is the top-left one among the list of all 40 output candidates.}
    \label{fig:mouseConvergence}
\end{figure}

\paragraph{Sensitivity Analysis of the Internal Parameters}

As our approach contains a number of internal parameters, it is of interest to systematically investigate the influence of these parameters w.r.t. the final performance of our system.
Here we consider five influential parameters, which are the number of initial poses $K_t$, the number of rounds $C$, the number and depth of trees in our first type of regressors (i.e. the local regressors $\left\{ \mathcal{R}_{\mathrm{j}}^{(c)} \right\}$), as well as the number of trees used in our learned metric component $\mathcal{R}_{\mathrm{m}}$.
Fig.~\ref{fig:internalParams} displays the performance of \emph{Lie-X} with respect to each of these five parameters row-by-row. Meanwhile each of the three columns presents the respective results for fish, mouse, and hand. Each of the fifteen panels in this five by three matrix is obtained by varying one parameter of interest while keeping the other parameters fixed to default values. In general, our system behaves in a rather stable manner w.r.t. the change of internal parameters over a wide range of values. Moreover, in each of the panels, a red colored dot is placed to indicate the specific parameter value empirically employed by our approach in this paper. It is worth noting that the choice of these internal parameter values represents a compromise between performance and efficiency.

\begin{table*}
	\begin{center}
  		\begin{tabular}{ | l || c | c |}
    		\hline
	    Comparison methods &  uses GPU & frames per second (FPS) \\ \hline \hline
		Tompson et. al. ~\cite{TomEtAl:siggraph14} & \checkmark & 30 \\ \hline
		Oberweger et. al. ~\cite{ObeWohLep:cvww15} & \checkmark & \textbf{5,000} \\ \hline
	    Oberweger et. al. ~\cite{ObeWohLep:iccv15} & \checkmark & 400 \\ \hline
		Zhou et. al. ~\cite{zhoumodel:ijcai16} & \checkmark & 125 \\ \hline
		Lie-X &  $\times$ & 123 \\ \hline
	\end{tabular}
\end{center}
	\caption{Runtime speed comparison with state-of-the-art methods for hand pose pose estimation task. Note our \emph{Lie-X} results are obtained using \emph{CPU only}, while the rest methods all utilize GPUs.}
    \label{tab:efficiencyHand}
\end{table*}

\begin{figure*}[!t]
    \centering

    \begin{tabular} {@{}c@{}c@{}c@{}c@{}c@{}c@{}c@{}c@{}c}
	& & & & & & & \multicolumn{2}{c}{    	
    		\subfloat {
			\includegraphics[width=0.2\textwidth]{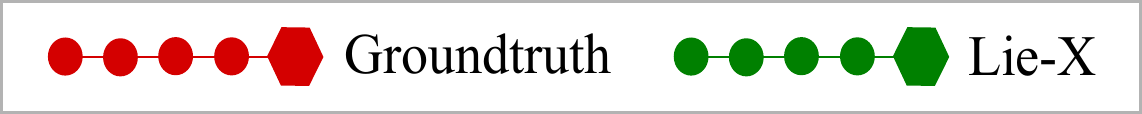}
    		}
    }
    \\
    \subfloat {
		\includegraphics[width=0.1\textwidth]{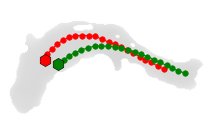}
    }
&   \subfloat {
		\includegraphics[width=0.1\textwidth]{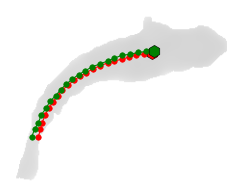}
    }
&    \subfloat {
		\includegraphics[width=0.1\textwidth]{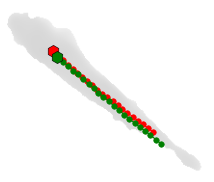}
    }
&	\subfloat {
		\includegraphics[width=0.1\textwidth]{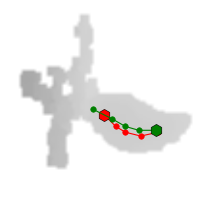}
    }
&   \subfloat {
		\includegraphics[width=0.1\textwidth]{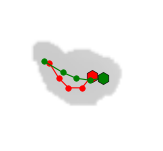}
    }
&    \subfloat {
		\includegraphics[width=0.1\textwidth]{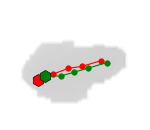}
    }
&	\subfloat {
		\includegraphics[width=0.1\textwidth]{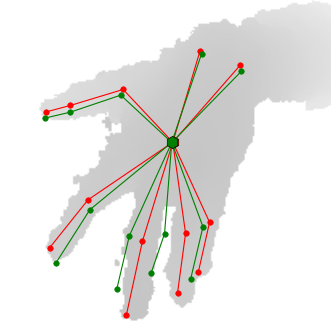}
    }
&   \subfloat {
		\includegraphics[width=0.1\textwidth]{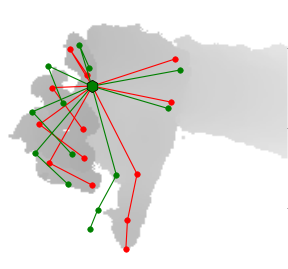}
    }
&    \subfloat {
		\includegraphics[width=0.1\textwidth]{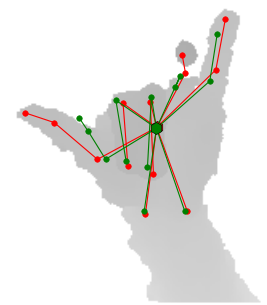}
    }
    \\
    \subfloat {
		\includegraphics[width=0.1\textwidth]{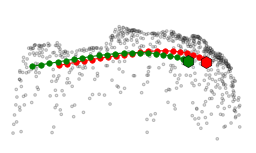}
    }
&   \subfloat {
		\includegraphics[width=0.1\textwidth]{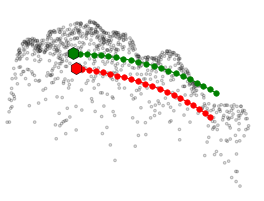}
    }
&    \subfloat {
		\includegraphics[width=0.1\textwidth]{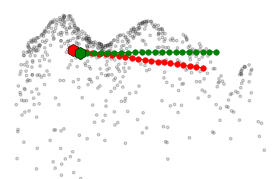}
    }
&	\subfloat {
		\includegraphics[width=0.1\textwidth]{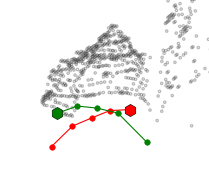}
    }
&   \subfloat {
		\includegraphics[width=0.1\textwidth]{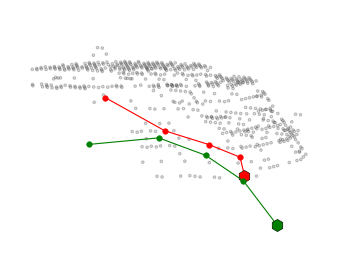}
    }
&    \subfloat {
		\includegraphics[width=0.1\textwidth]{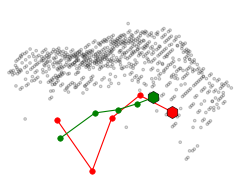}
    }
&	\subfloat {
		\includegraphics[width=0.1\textwidth]{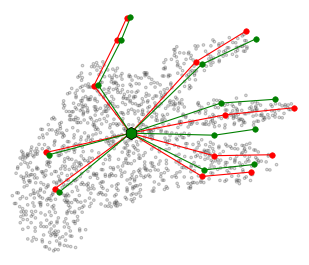}
    }
&   \subfloat {
		\includegraphics[width=0.1\textwidth]{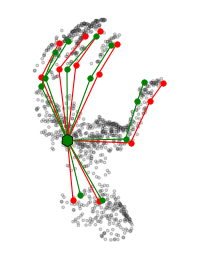}
    }
&    \subfloat {
		\includegraphics[width=0.1\textwidth]{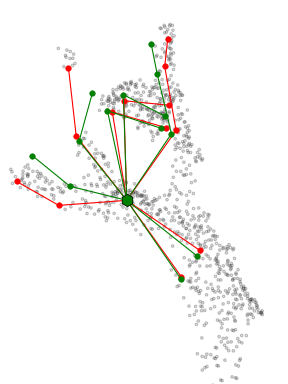}
    }
    \\
    \multicolumn{3}{c}{(a) Fish} & \multicolumn{3}{c}{(b) Mouse} & \multicolumn{3}{c}{(c) Hand}
    \\
    	\end{tabular}
    \caption{Visual examples of common pose estimation errors made by our \textit{Lie-X} approach. These errors include orientation flips, displacement along the z-direction and sub-optimal shape fits. Each of the columns presents an exemplar depth image of fish, mouse, and hand, respectively, while the first and second rows display its top and side views.}
    \label{fig:failureCases}
\end{figure*}

\paragraph{Comparison with Baselines and the State-of-the-art Methods}
To evaluate the performance of the proposed approach, a series of experiments are conducted on the aforementioned datasets for fish, mouse, and hand pose estimation tasks.
Table~\ref{tab:poseEstimationResult} presents a comparison of \emph{Lie-X} w.r.t. the two non-trivial baseline methods (i.e. RF and CNN) on fish and mouse pose estimation tasks. Overall, our approach clearly outperforms the others by a significant margin, while CNN achieves better results over RF.
Moreover, the error distributions of these comparison methods are also presented in Fig.~\ref{fig:errDistribution}(a-b),
where our approach clearly outperforms the baselines most of the time.
The superior performance of our \emph{Lie-X} approach is also demonstrated in Fig.~\ref{fig:poseEstimationVisualResults},
which provides visual comparisons of pose estimation results for six fish and six mouse examples, respectively, over the three competing methods. From these visual examples, it is observed that the estimated poses from our \emph{Lie-X} approach tend to be more faithfully aligned with the ground-truth when compared against the two baseline methods.

Our approach is also validated on the NYU hand depth benchmark, as is displayed in Table~\ref{tab:poseEstimationResultHand}. Overall, our CNN baseline result is on par with the standard deep learning results of e.g.~\cite{ObeWohLep:cvww15} that also utilizes a AlexNet-like CNN. This helps to establish that our baselines are consistent in terms of performance with what has been reported in the literature, which are also used as pose estimation baselines on fish and mouse objects.
Moreover, the results of the state-of-the-art methods are also directly compared here, including Tompson et. al.~\cite{TomEtAl:siggraph14}, Oberweger et. al.~\cite{ObeWohLep:iccv15}, and Zhou et. al.~\cite{zhoumodel:ijcai16}. 
It is worth pointing out that the test error rate of our approach is 14.51~mm in terms of average joint error. This is by far the best result on hand pose estimation task to our knowledge, which improves over the best state-of-the-art result of 16.50~mm of~\cite{ObeWohLep:iccv15} by almost 2~mm.
More detailed quantitative information is revealed through the error distributions of comparison methods in Fig.~\ref{fig:errDistribution}(c), where our approach clearly outperforms the baselines and the state-of-the-art methods.
Similarly, visual comparison results are provided in Fig.~\ref{fig:poseEstimationVisualResultsHand}, Fig.~\ref{fig:poseEstimationVisualResultsHand2}, and Fig.~\ref{fig:poseEstimationVisualResultsHand}, where our approach is again shown to clearly outperform other methods. More specifically, Fig.~\ref{fig:poseEstimationVisualResultsHand} and Fig.~\ref{fig:poseEstimationVisualResultsHand2} present the visual results of all competing methods on the same four exemplar hand images. Due to the access limit, we are only able to present the side-view results on our approach and the baseline methods of RF and CNN. Fig.~\ref{fig:poseEstimationVisualResultsHand} provides additional visual results comparing our approach to state-of-the-arts on ten more hand images.

To reveal the inner working of our approach, in Fig.~\ref{fig:fishCandidatesMetric}, Fig.~\ref{fig:mouseCandidatesMetric}, and Fig.~\ref{fig:handCandidatesMetric}, a visual example is respectively provided for pose estimation of fish, mouse, and hand. It is evident that a diverse set of pose candidates are obtained that covers distinct pose location, orientation, and sizes. This is made possible due to the adoption of multiple initial poses. Moreover, prediction scores and associated orders from our learned metric module in general closely resembles that of the empirical evaluation metric. In addition, Fig.~\ref{fig:mouseConvergence} presents several intermediate pose estimation results from different joints and rounds on an exemplar mouse image, when executing the pose estimation pipeline as illustrated in Fig.~\ref{fig:framework}. It is observed that each step of the iterative process usually helps in converging toward the final pose estimation.

\begin{figure*}[!t]
    \centering
	    \includegraphics[width=0.99\textwidth]{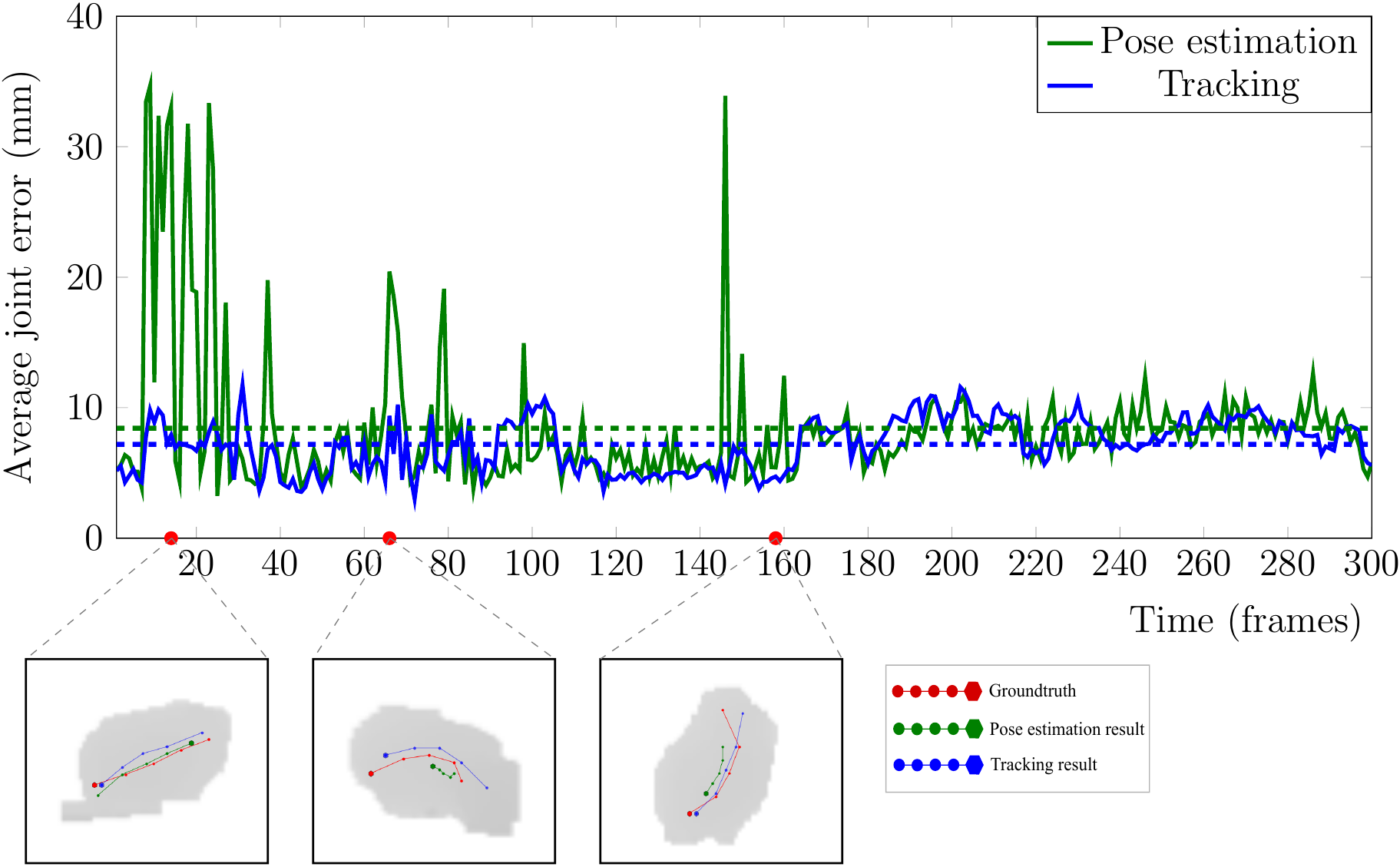}
    \caption{Pose estimation vs. tracking: An comparison of the average joint error on frames of a mouse test sequence when employing our pose estimation (Alg.~\ref{alg:PE_test}) vs. tracking (Alg.~\ref{alg:3Dtracking}) modules. The horizontal dotted lines in \textcolor{green}{green} and \textcolor{blue}{blue} colors are the respective mean errors of pose estimation and tracking results. Visual comparisons at various time frames are presented in the bottom row.}
    \label{fig:mouse_tracking_avg_err}
\end{figure*}

\begin{figure*}[!t]
    \centering
    \subfloat[] {
	    \includegraphics[width=0.5\textwidth]{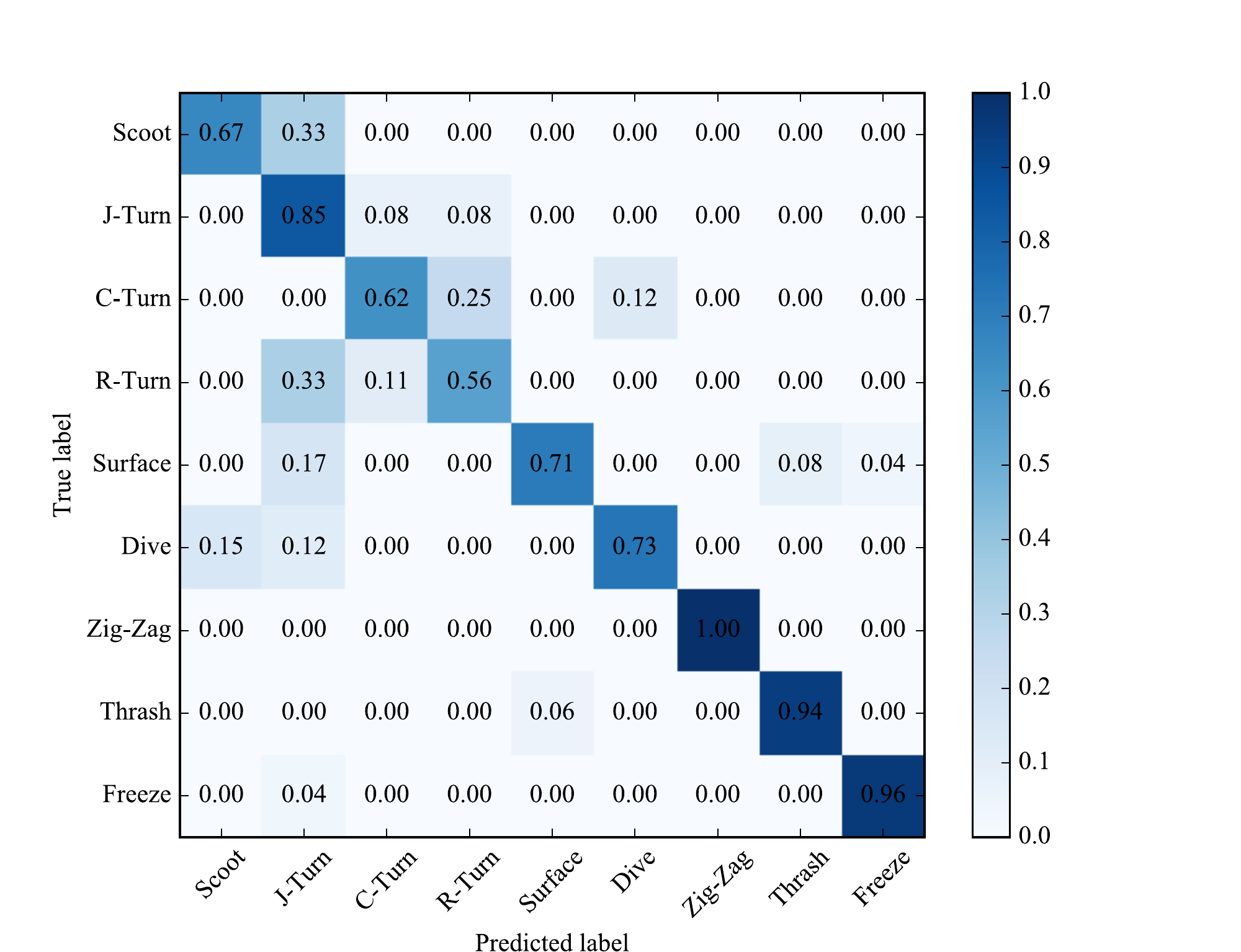}
    }
    \subfloat[] {
	    \includegraphics[width=0.5\textwidth]{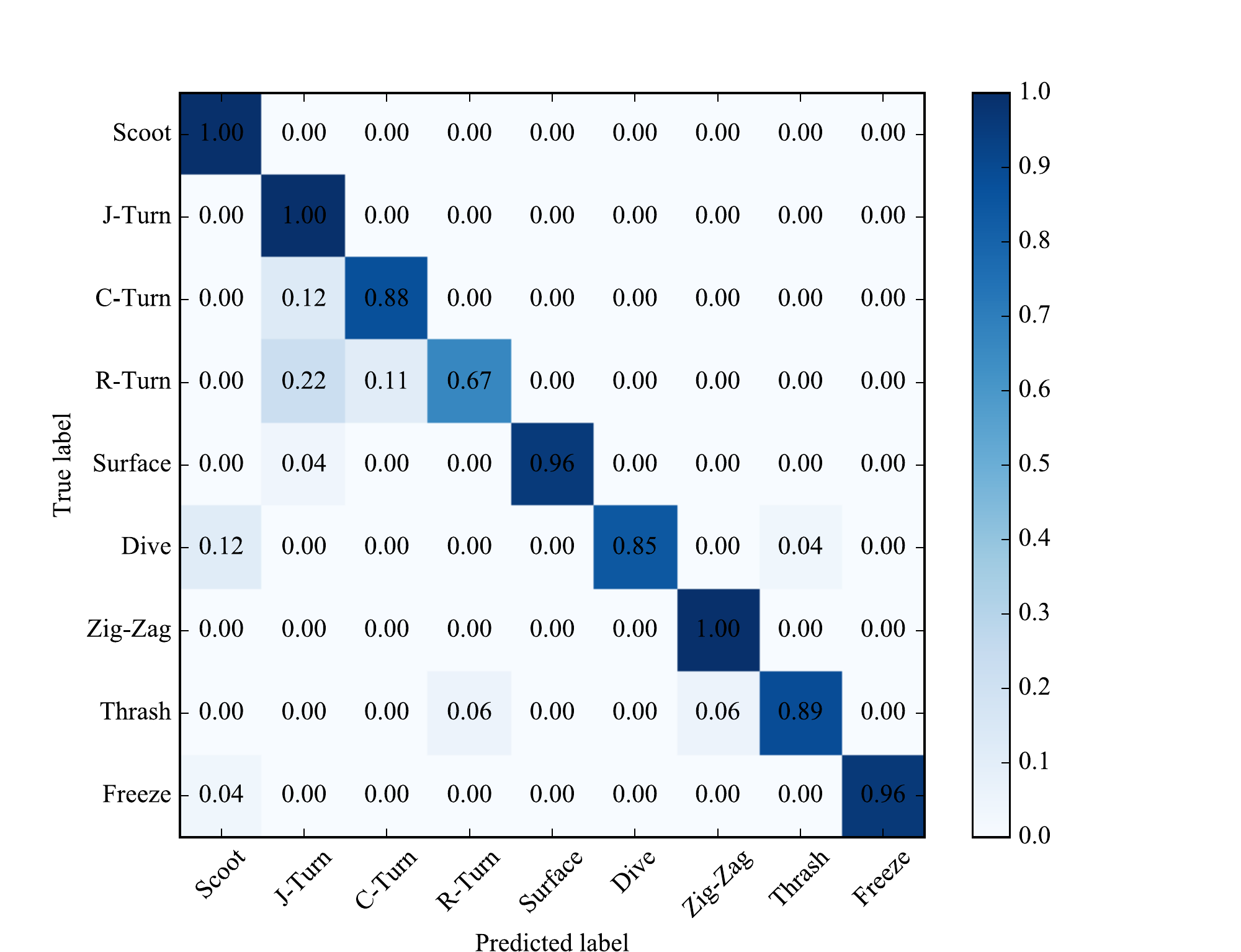}
    }
    \caption{Action recognition confusion matrices on our fish action dataset. (a) is for joint position based features, while (b) is for tangent vector based features. Their overall performance in terms of average classification accuracy for (a) and (b) is 79.19\% and 91.33\%, respectively.}
    \label{fig:actRecConfusionMtx}
\end{figure*}

\paragraph{With vs. Without Multiple Initial Poses}
As presented in Table~\ref{tab:poseEstimationResult} for fish and mouse objects and Table~\ref{tab:poseEstimationResultHand} for hand objects, empirically we observe that the presence of multiple initial poses always improves the pose estimation performance. 
As presented in Figs.~\ref{fig:fishCandidatesMetric},~\ref{fig:mouseCandidatesMetric}, and~\ref{fig:handCandidatesMetric}, execution of our pose estimation process, starting from multiple distinct initial poses, results in unique pose estimates, each of which can be regarded as a locally optimal result.
This is due to the highly non-convex nature of our problem, a well-known fact for systems of rigid-bodies in general. These visual examples also demonstrate the importance of having multiple initial poses to avoid getting trapped into local optimal points that are far from the ground-truth point.

\paragraph{With vs. Without the Learned Metric}
To examine the usefulness of our learned internal metric, a special variant of our approach without this component is engaged here, which is also referred to as \emph{Lie-X} w/o learned metric. Provided with multiple output pose candidates, this variant differs from our full-fledged approach by averaging over them for each of the joints in the 3D Euclidean space, instead of scoring them with the learned metric to pick up the best estimate.
Empirical experiments such as those presented in Table~\ref{tab:poseEstimationResult} for fish and mouse objects and Table~\ref{tab:poseEstimationResultHand} for hand objects suggest a noticeable performance degradation when without the learned metric. Clearly the learned internal metric does facilitate in selecting from a global viewpoint the final estimate, which is obtained from the pool of locally optimal candidates using multiple initial poses.
It has also been demonstrates in Figs.~\ref{fig:fishCandidatesMetric},~\ref{fig:mouseCandidatesMetric}, and~\ref{fig:handCandidatesMetric} that in our context a max operation (i.e. with the learned metric) may well outperform an average operation (i.e. w/o learned metric).
In particular, visually our learned internal metric is capable of producing predicted error scores that are nicely aligned with the \emph{true} average joint error when we have access to the ground-truth.

\paragraph{Computational Efficiency}
All experiments discussed in this paper are performed on a desktop PC with an Intel i7-960 CPU 
and with 24Gb memory. 
At this moment, our CPU implementation achieves an average run-time speed of 83 FPS, 267 FPS, and 123 FPS for fish, mouse, and hand tasks, respectively.
Table~\ref{tab:efficiencyHand} summarizes the run-time speed comparisons with state-of-the-art hand pose estimators on the NYU hand dataset~\cite{TomEtAl:siggraph14}.
Our result of 123 FPS is obtained with only CPU access, nevertheless it is still comparable with most of these recent methods which are based on GPUs.
Note the empirical runtime speed of our approach could be further improved by exploiting the computing power of modern GPUs.
Meanwhile, an exceptionally high speed method is developed in Oberweger et. al.~\cite{ObeWohLep:cvww15}, which is made possible by the usage of very shallow neural nets.
This however comes with degraded performance as shown in Table~\ref{tab:poseEstimationResultHand}, with a significant average joint error increase of 5.41 mm when compared to our approach.

\paragraph{Common Pose Estimation Errors of Our Approach}
Although our \emph{Lie-X} approach performs relatively well in practical pose estimation settings, inevitably it does make mistakes in practical situations.
These common errors include the following ones: orientation flips, displacement along the z-direction and sub-optimal shape fitting.
A visual illustration of these common errors made by our approach is provided in Fig.~\ref{fig:failureCases}.
As can be observed, usually our Lie-Fish results are best aligned with the ground-truths. The mistakes of Lie-Mouse are more noticeable. Meanwhile the visual displacements of our Lie-Hand results from the ground-truths are most significant. This is to be expected, as the corresponding complexity levels of the three tasks varies from being relatively simple (i.e. kinematic chains) to complex (i.e. kinematic trees).

\subsection{Tracking of Mouse}
Our \emph{Lie-X} approach is also examined on the tracking task using the mouse tracking dataset described beforehand. 
Compared with our single frame based pose estimation of Alg.~\ref{alg:PE_test}, it is of interest to examine on how much we can gain from our tracking algorithm of Alg.~\ref{alg:3Dtracking}, when temporal information is available.
Empirically our mouse tracker is shown to produce an improved performance of 7.19~mm from the 8.42~mm results from our pose estimator on single frames. This can also be observed from the bottom row of Fig.~\ref{fig:mouse_tracking_avg_err} where visual comparisons are provides at several time frames.
By exploiting temporal information, the results of our tracker are shown to produce less dramatic mistakes comparing to that of pose estimation.
It is again evidenced quantitatively in Fig.~\ref{fig:mouse_tracking_avg_err}, which presents a frame-by-frame average joint error comparison of tracking vs. pose estimation on a test sequence.
Clearly there exists a number of very noisy predictions of pose estimation. In comparison our tracking results are in general much less noisy.
Overall, the tracking results outperforms post estimation with a noticeable gap of at least 1~mm. 
Note the tracking results in some frames are slightly inferior to that of the pose estimation counterpart, which we attribute to the utilization of the averaging operations in our tracker.

\subsection{Action Recognition of fish}

To demonstrate the application of our approach on action recognition tasks, in what follows we conduct experiments on the aforementioned fish action dataset. In addition to the proposed tangent vector (i.e. Lie algebras) based feature representation, as comparison we also consider a joint based feature representation.
Here the main body of the feature representation follows exactly as in the tangent vector representation, including e.g. the adoption of a temporal pyramid of $\{4,2,1\}$,
with the only change as follows: Instead of tangent vectors, the corresponding the 3D joint positions are employed. This finally leads to an 888-dim feature vector representation.

Fig.~\ref{fig:fishactions} displays our fish dataset that contains nine unique action categories. The standard evaluation metric of average classification accuracy are considered in this context.
Empirically the comparison method that utilizing joint position features achieves a performance of 79.19\%, which is significantly outperformed by our approach based on tangent vector features with the average accuracy of 91.33\%.
Fig.~\ref{fig:actRecConfusionMtx} provides further information of category-wise errors in the form of the confusion matrices. It is observed that the joint based method tends to confuse among the subset of actions of \emph{scoot}, \textit{J-turn}, \emph{c-turn}, and \emph{r-turn}, which are indeed more challenging to be separated due to their inherent similarities.
Nonetheless, the performance on this subset is dramatically improved in our approach with tangent vector based features. We hypothesize that by following the natural tangent vector representation, our approach gains the discriminative power to separate these otherwise troublesome action categories.

\section{Conclusion and Future Work}

A unified Lie group approach is proposed for the related key problems of pose estimation, tracking,
and action recognition of diverse articulated objects from depth images. Empirically our approach is evaluated on human hand, fish and mouse datasets with very competitive performance.
For future work, we plan to work with more diverse articulated objects such as human full body and wild animals, as well as their interactions with other articulated objects and background objects.

\section*{Acknowledgment}
The project is partially supported by A*STAR JCO grants 1431AFG120 and 15302FG149.
Mouse and fish images are acquired with the help of Zoe Bichler, James Stewart, Suresh Jesuthasan, and Adam Claridge-Chang.
Zilong Wang helps with the annotation of mouse data, while Wei Gao and Ashwin Nanjappa help with implementing the mouse baseline method.

{\small
\bibliographystyle{plain}
\bibliography{main}
}

\end{document}